\documentclass[twoside,11pt]{article}

\usepackage{jmlr2e}
\usepackage{amsfonts, url, amsmath, color, times, algorithm, algorithmic}
\usepackage{multirow}

\newcommand{\BEAS}{\begin{eqnarray*}}
\newcommand{\EEAS}{\end{eqnarray*}}
\newcommand{\BEA}{\begin{eqnarray}}
\newcommand{\EEA}{\end{eqnarray}}
\newcommand{\BEQ}{\begin{equation}}
\newcommand{\EEQ}{\end{equation}}
\newcommand{\BIT}{\begin{itemize}}
\newcommand{\EIT}{\end{itemize}}
\newcommand{\BNUM}{\begin{enumerate}}
\newcommand{\ENUM}{\end{enumerate}}
\newcommand{\BA}{\begin{array}}
\newcommand{\EA}{\end{array}}

\newcommand{\mysec}[1]{Section~\ref{sec:#1}}
\newcommand{\myeq}[1]{Eq.~(\ref{eq:#1})}
\newcommand{\myfig}[1]{Figure~\ref{fig:#1}}
\newcommand{\myappendix}[1]{Appendix~\ref{app:#1}}
\newcommand{\mylemma}[1]{Lemma~\ref{lem:#1}}

\newcommand{\DistanceToCaption}{\vspace*{-0.5cm}}
\newcommand{\UpperDistanceToFig}{\vspace*{0.5cm}}

\newcommand{\Prob}[1]{\mathbb{P}\left( #1 \right) }
\newcommand{\IntegerSet}[1]{\{1,\dots,#1\}}

\newcommand{\cB}{{\mathcal B}}
\newcommand{\cS}{{\mathcal S}}
\newcommand{\ty}{{\tilde y}}
\newcommand{\tj}{{\alpha}}
\newcommand{\tF}{\tilde{F}}
\newcommand{\tFlim}{\tilde{F}_\text{lim}}
\newcommand{\Jo}{ { \mathbf{ \overline{J} } } }
\newcommand{\constant}{ \alpha }

\newcommand{\R}[1]{\mathbb{R}^{#1}}
\newcommand{\RR}[2]{\mathbb{R}^{#1 \times #2}}

\newcommand{\Qb}{\mathbf{Q}}

\newcommand{\J}{\mathbf{J}}
\newcommand{\Jb}{\mathbf{J}}
\newcommand{\Jh}{\hat{J}}
\newcommand{\JCb}{\mathbf{J}^c}
\newcommand{\JCh}{\hat{J}^c}

\newcommand{\G}{\mathcal{G}}
\newcommand{\dG}{d^{\scriptscriptstyle G}}
\newcommand{\dH}{d^{\scriptscriptstyle H}}
\newcommand{\etaG}{\eta^{\scriptscriptstyle G}}

\newcommand{\Pattern}{\mathcal{P}}
\newcommand{\ZPattern}{\mathcal{Z}}

\newcommand{\wb}{\mbox{${\mathbf{w}}$}}
\newcommand{\w}{\mbox{${\mathbf{w}}$}}
\newcommand{\wh}{\mbox{$\hat{w}$}}

\def \E{{\mathbb E}}
\def \P{{\mathbb P}}

\newcommand{\NormUn}[1]{\left\|#1\right\|_1}
\newcommand{\NormDeux}[1]{\left\|#1\right\|_2}

\newcommand{\NormInf}[1]{\left\|#1\right\|_{\infty}}

\newcommand{\NormOmega}[2]{\sum_{G \in #1}\NormDeux{\dG \circ #2}}

\newcommand{\NormUnc}[1]{\|#1\|_1}
\newcommand{\NormDeuxc}[1]{\|#1\|_2}

\newcommand{\ERisk}[2]{\frac{1}{n}\sum_{i=1}^n \ell\big( #1, #2 \big) }

\ShortHeadings{Structured Variable Selection with Sparsity-Inducing Norms}{Jenatton, Audibert and Bach }
\firstpageno{1}

\begin{document}

\title{Structured Variable Selection with Sparsity-Inducing Norms}

\author{\name Rodolphe Jenatton  \email rodolphe.jenatton@inria.fr \\
	\addr INRIA - WILLOW Project-team, \\
	 Laboratoire d'Informatique de l'Ecole Normale Sup\'erieure (INRIA/ENS/CNRS UMR 8548),\\
         23, avenue d'Italie, 75214 Paris, France \AND
         \name Jean-Yves Audibert  \email audibert@certis.enpc.fr \\
         \addr Imagine (ENPC/CSTB), Universit\'e Paris-Est,\\
          Laboratoire d'Informatique de l'Ecole Normale Sup\'erieure (INRIA/ENS/CNRS UMR 8548), \\
          6 avenue Blaise Pascal, 77455 Marne-la-Vall\'ee, France \AND
         \name Francis Bach  \email francis.bach@inria.fr \\
         \addr INRIA - WILLOW Project-team, \\
	 Laboratoire d'Informatique de l'Ecole Normale Sup\'erieure (INRIA/ENS/CNRS UMR 8548),\\
         23, avenue d'Italie, 75214 Paris, France}

\editor{}

\maketitle

\begin{abstract}
We consider the empirical risk minimization problem for linear supervised learning, with regularization 
by structured sparsity-inducing norms. These are defined as sums of Euclidean norms on certain subsets of variables, extending the usual $\ell_1$-norm and the group $\ell_1$-norm by allowing the subsets to overlap. This leads to a specific set of allowed nonzero patterns for the solutions of such problems. We first explore the relationship between the groups defining the norm and the resul\-ting nonzero patterns, providing both forward and backward algorithms to go back and forth from groups to patterns. This allows the design of norms adapted to specific prior knowledge expressed in terms of nonzero patterns. We also present an efficient active set algorithm, and analyze the consistency of variable selection for least-squares linear regression in low and high-dimensional settings. 
\end{abstract}


\section{Introduction}

Sparse linear models have emerged as a powerful framework to deal with various supervised estimation tasks, 
in machine learning as well as in statistics and signal processing.
These models basically seek to predict an output by linearly combining only a small subset of the features describing the data.
To simultaneously address this variable selection and the linear model estimation, 
$\ell_1$-norm regularization has become a popular tool, 
that benefits both from efficient algorithms~\citep[see, e.g.,][and multiple references therein]{lars,ng-sparsecoding,yuan-comparisonL1} 
and well-developed theory for generalization properties and variable selection consistency~\citep{Zhaoyu,martin,bickel_lasso_dantzig,zhang2009some}.

When regularizing by the $\ell_1$-norm, sparsity is yielded by treating each variable individually, 
regardless of its position in the input feature vector, 
so that existing relationships and structures
between the variables (e.g., spatial, hierarchical or related to the physics of the problem at hand) are merely disregarded.
However, many practical situations could benefit from this type of prior knowledge, 
potentially both for interpretability purposes and for improved predictive performance.

For instance, in neuroimaging, one is interested in localizing areas in 
functional magnetic resonance imaging (fMRI) or
magnetoencephalography (MEG)
signals that are discriminative to distinguish between different brain states \citep[and references therein]{kowalski-gramfort09, xiang-boosting}.
More precisely, fMRI responses consist in voxels whose three-dimensional spatial arrangement respects the anatomy of the brain.
The discriminative voxels are thus expected to have a specific localized spatial organization \citep{xiang-boosting}, 
which is important for the subsequent identification task performed by neuroscientists.
In this case, regularizing by a plain $\ell_1$-norm to deal with the ill-conditionedness of the problem 
(typically only a few fMRI responses described by tens of thousands of voxels) 
would ignore this spatial configuration, with a potential loss in interpretability and performance.

Similarly, in face recognition, robustness to occlusions can be increased by considering as features,
sets of pixels that form small convex regions on the face images \citep{SparseStructuredPCA}. 
Again, a plain $\ell_1$-norm regularization fails to encode this specific spatial locality constraint \citep{SparseStructuredPCA}.
Still in computer vision, object and scene recognition generally seek to extract bounding boxes in either images \citep{cordelia_image} 
or videos \citep{cordelia_video}. 
These boxes concentrate the predictive power associated with the considered object/scene class, 
and have to be found by respecting the spatial arrangement of the pixels over the images. 
In videos, where series of frames are studied over time, the temporal coherence also has to be taken into account.
An unstructured sparsity-inducing penalty that would disregard this spatial and temporal information is therefore not adapted to select such boxes.

Another example of the need for higher-order prior knowledge comes from bioinformatics.
Indeed, for the diagnosis of tumors, the profiles of array-based comparative genomic hybridization (arrayCGH) can be used as inputs 
to feed a classifier \citep{Rapaport2008Classification}.
These profiles are characterized by plenty of variables, but only a few samples of such profiles are available, prompting the need for variable selection.
Because of the specific spatial organization of bacterial artificial chromosomes along the genome, 
the set of discriminative features is expected to have specific contiguous patterns. 
Using this prior knowledge on top of a standard sparsity-inducing method leads to improvement in classification accuracy \citep{Rapaport2008Classification}.  
In the context of multi-task regression, a genetic problem of interest is to find a mapping between 
a small subset of single nucleotide polymorphisms (SNP's) that have a phenotypic impact on a given family of genes \citep{kim2009tree}.
This target family of genes has its own structure, where some genes share common genetic characteristics, 
so that these genes can be embedded into a underlying hierarchy \citep{kim2009tree}. 
Exploiting directly this hierarchical information in the regularization term outperforms the unstructured approach with a standard $\ell_1$-norm.
Such hierarchical structures have been likewise useful in the context of wavelet regression \citep{cap} 
or kernel-based non linear variable selection \citep{hkl}.

These real world examples motivate the need for the design of sparsity-inducing regularization schemes, 
capable of encoding more sophisticated prior knowledge about the expected sparsity patterns. 

As mentioned above, the $\ell_1$-norm focuses only on \textit{cardinality} 
and cannot easily specify side information about the patterns of nonzero coefficients (``nonzero patterns'') induced in the solution, 
since they are all theoretically possible. 
Group $\ell_1$-norms~\citep{Yuan-Lin-GroupLasso,roth,huang2009benefit} consider a partition of all variables into a certain number of subsets and penalize the sum of the Euclidean norms of each one, leading to selection of groups rather than individual variables. Moreover, recent works have considered overlapping but nested groups in constrained situations such as trees and directed acyclic graphs~\citep{cap,hkl,kim2009tree}. 

In this paper, we consider all possible sets of groups and characterize exactly what type of prior knowledge can be encoded by considering sums of norms of overlapping groups of variables. 
Before describing how to go from groups to nonzero patterns (or equivalently zero patterns), we show that it is possible to ``reverse-engineer'' a given set of nonzero patterns, i.e., to build the unique minimal set of groups that will generate these patterns. This allows the automatic design of sparsity-inducing norms, adapted to target sparsity patterns.  
We give in \mysec{patterns} some interesting examples of such designs in specific geometric and structured configurations, 
which covers the type of prior knowledge available in the real world applications described previously.

As will be shown in \mysec{patterns}, for each set of groups, a notion of hull of a nonzero pattern may be naturally defined. 
For example, in the particular case of the two-dimensional planar grid considered in this paper, 
this hull is exactly the axis-aligned bounding box or the regular convex hull. We show that, in our framework, the allowed nonzero patterns are exactly those equal to their hull, and that the hull of the relevant variables is consistently estimated under certain conditions, both in low and high-dimensional settings. Moreover, we present in \mysec{optimization} an efficient active set algorithm that scales well to high dimensions.
Finally, we illustrate in \mysec{experiments} the behavior of our norms with synthetic examples on specific geometric settings, such as lines and two-dimensional grids.

\paragraph{Notation.} 
For $x \in \R{p}$ and $q \in [1,\infty)$, we denote by $\| x\|_q$ its $\ell_q$-norm defined as $(\sum_{j=1}^p |x_j|^q )^{1/q}$ and $\NormInf{x}=\max_{j\in\IntegerSet{p}}|x_j|$. 
Given  $w \in \R{p}$ and a subset $J$ of $\{1,\dots,p\}$ with cardinality $|J|$, $w_J$ denotes the vector in $\R{|J|}$ of elements of $w$ indexed by $J$. 
Similarly, for a matrix $M \in \RR{p}{m}$, $M_{IJ} \in \RR{|I|}{|J|}$  denotes the sub-matrix of $M$ reduced to the rows indexed by $I$ and the columns indexed by $J$.
For any finite set $A$ with cardinality $|A|$, we also define the
$|A|$-tuple $(y^a)_{a \in A} \in \RR{p}{|A|}$ as the collection of
$p$-dimensional vectors $y^a$ indexed by the elements of $A$.
Furthermore, for two vectors $x$ and $y$ in $\R{p}$, we denote by $x \circ
y = (x_1y_1,\dots,x_p y_p)^\top \in \R{p}$ the elementwise product of $x$
and $y$.

\section{Regularized Risk Minimization}\label{sec:regularized_risk_minimization}

We consider the problem of predicting a random variable $Y \in \mathcal{Y}$ from a (potentially non random) vector $X \in \R{p}$, where  $\mathcal{Y}$ is the set of responses, typically a subset of $\R{}$. We assume that we are given  $n$ observations $(x_i,y_i) \in \R{p} \times
\mathcal{Y}$, $ i =1,\dots,n$.
We define the \emph{empirical risk} of a loading vector $w \in \R{p}$
as $L(w) = \ERisk{y_i}{w^\top x_i}$,
where $\ell: \mathcal{Y}\times \R{} \mapsto \R{+}$ is a \emph{loss function}. We assume that $\ell$ is \emph{convex and continuously differentiable} with respect to the second parameter. Typical examples of loss functions  
are the square loss for least squares regression, i.e., $\ell(y,\hat{y}) = \frac{1}{2}(y-\hat{y})^2$ with $y \in \R{}$, and the logistic loss $\ell(y,\hat{y}) = \log(1+e^{-y\hat{y}})$ for logistic regression, with $y \in \{-1,1\}$.

We focus on a general family of sparsity-inducing norms that allow the penalization of subsets of variables grouped together. Let us denote by $\G$ a subset of the power set of $\IntegerSet{p}$ such that $ \bigcup_{G \in \G}\! G = \IntegerSet{p}$, i.e.,  a spanning set of subsets of $\IntegerSet{p}$.
Note that $\G$ does not necessarily define a partition of $\IntegerSet{p}$, and therefore, \emph{it is possible for elements of $\G$ to overlap}. We consider the norm $\Omega$ defined by 
\begin{equation}\label{eq:def_norm_omega}
\Omega(w) = \sum_{G\in\G} \bigg( \sum_{j\in G} (\dG_j)^2 |w_j|^2 \bigg)^{\frac{1}{2}} = \NormOmega{\G}{w},
\end{equation} 
where $(\dG)_{G\in\G}$ is a $|\G|$-tuple of $p$-dimensional vectors such that $\dG_j > 0$ if $j \in G$ and $\dG_j = 0$ otherwise. 
A same variable $w_j$ belonging to two different groups $G_1, G_2 \in \G$ is allowed to be weighted differently in $G_1$ and $G_2$ (by respectively $d_j^{\scriptscriptstyle G_1}$ and $d_j^{\scriptscriptstyle G_2}$). 
We do not study the more general setting where each $\dG$ would be a (non-diagonal) positive-definite matrix, which we defer to future work.
Note that a larger family of penalties with similar properties may be obtained 
by replacing the $\ell_2$-norm in \myeq{def_norm_omega} by other $\ell_q$-norm, $q > 1$ \citep{cap}.
Moreover, non-convex alternatives to \myeq{def_norm_omega} with quasi-norms in place of norms may also be interesting, in order to yield sparsity more aggressively \citep[see, e.g.,][]{SparseStructuredPCA}.

This general formulation has several important sub-cases that we present below, the goal of this paper being to go beyond these, and to consider norms capable to incorporate richer prior knowledge.

\begin{itemize}

\item \textbf{$\ell_2$-norm}: $\G$ is composed of one element, the full set  $\IntegerSet{p}$.

\item \textbf{$\ell_1$-norm}: 
$\G$ is the set of all singletons, leading to the Lasso \citep{Tibshirani96regressionshrinkage} for the square loss.

\item \textbf{$\ell_2$-norm and $\ell_1$-norm}:
$\G$ is the set of all singletons and the full set $\IntegerSet{p}$, leading (up to the squaring of the $\ell_2$-norm) to the Elastic net \citep{zou2005rav} for the square loss.

\item \textbf{Group $\ell_1$-norm}: $\G$ is any partition of $\IntegerSet{p}$, leading to the group-Lasso for the square loss~\citep{Yuan-Lin-GroupLasso}.

\item \textbf{Hierarchical norms}: when the set $\IntegerSet{p}$ is embedded into a tree~\citep{cap} or more generally into a directed acyclic graph~\citep{hkl}, then a set of $p$ groups, each of them composed of descendants of a given variable, is considered.
\end{itemize}
We study the
following regularized problem:
\begin{equation}\label{eq:minF}
 \min_{w \in \R{p}} \  \ERisk{y_i}{w ^\top x_i}+\mu\Omega(w),
\end{equation}
where $\mu\!\geq\!0$ is a regularization parameter. Note that a non-regularized constant term could be included in this formulation, but it is left out for simplicity.
We denote by $\hat{w}$ any solution of \myeq{minF}. Regularizing by linear combinations of (non-squared) $\ell_2$-norms is known to induce sparsity in $\hat{w}$~\citep{cap}; our grouping leads to specific patterns that we describe in the next section.

\section{Groups and Sparsity Patterns}\label{sec:patterns}

We now study the relationship between the norm $\Omega$ defined in \myeq{def_norm_omega} and the nonzero patterns the estimated vector $\wh$ is allowed to have.  We first characterize the set of nonzero patterns, then we provide forward and backward procedures to go back and forth from groups to patterns.

\subsection{Stable Patterns Generated by $\G$}

\begin{figure}
\begin{center}
\includegraphics[scale=.55]{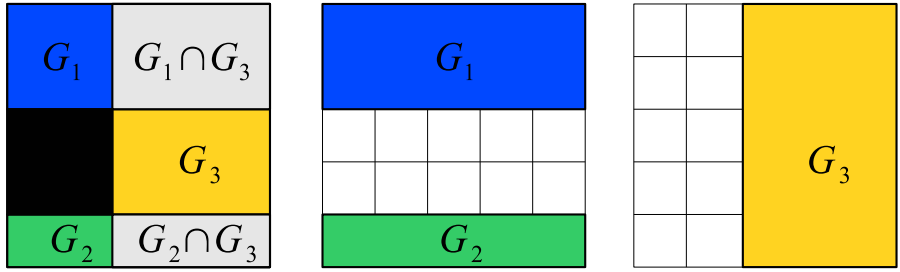}
\end{center}

\DistanceToCaption

\caption{Groups and induced nonzero pattern: three sparsity-inducing groups (middle and right, denoted by $\{G_1,G_2,G_3\}$) with the associated nonzero pattern which is the complement of the union of groups, i.e., $(G_1 \cup G_2 \cup G_3)^c$ (left, in black).}
\label{fig:groups_example}
 \end{figure}

The regularization term $\Omega(w)=\NormOmega{\G}{w}$ is a mixed $(\ell_1, \ell_2$)-norm~\citep{cap}. 
At the group level, it behaves like an $\ell_1$-norm  and therefore, $\Omega$ induces group sparsity.
In other words, each $\dG \circ w$, and equivalently
each $w_G$ (since the support of $\dG$ is exactly $G$), is encouraged to go to zero.
On the other hand, within the groups $G\in\G$, the $\ell_2$-norm does not promote sparsity. 
Intuitively, for a certain subset of groups $\G'\! \subseteq\! \G$, the vectors $w_G$ associated with the groups $G\! \in\! \G'$ will be exactly equal to zero, 
leading to a set of zeros which is the union of these groups, $\bigcup_{G\in\G'}\!G$. 
Thus, the set of allowed zero patterns should be the \emph{union-closure} of $\G$, i.e. (see \myfig{groups_example} for an example):
\BEQ
\label{eq:Z}
\ZPattern=\bigg\lbrace \bigcup_{G\in\G'}G;\  \G' \subseteq \G \bigg\rbrace.
\EEQ
The situation is however slightly more subtle as some zeros can be created by chance 
(just as regularizing by the $\ell_2$-norm may lead, though it is unlikely, to some zeros). 
Nevertheless, Theorem~\ref{thm:stability} shows that, under mild conditions, 
the previous intuition about the set of zero patterns is correct. 
Note that instead of considering the set of zero patterns $\ZPattern$, it is also convenient to manipulate nonzero patterns, and we define
 \BEQ
 \Pattern=\bigg\lbrace \bigcap_{G\in\G'}G^c;\ \G' \subseteq \G \bigg\rbrace=\big\lbrace Z^c;\ Z \in \ZPattern \big\rbrace. 
 \EEQ
We can equivalently use $\Pattern$ or $\ZPattern$ by taking the complement of each element of these sets.

The following two results characterize the solutions of the problem~(\ref{eq:minF}).
We first gives sufficient conditions under which this problem has a unique solution.
We then formally prove the aforementioned intuition about the zero patterns of the solutions of (\ref{eq:minF}),
namely they should belong to $\ZPattern$. 
In the following two results (see proofs in \myappendix{uniqueness_theorem} and \myappendix{stability_theorem}), 
we assume that $\ell:(y,y')\mapsto \ell(y,y')$ is nonnegative, twice
continuously differentiable with positive second derivative with respect to the
second variable and non-vanishing mixed derivative, i.e., for any $y,y'$ in
$\R{}$, $\frac{\partial^2 \ell}{\partial {y'}^2}(y,y')>0$ and $\frac{\partial^2
\ell}{\partial y\partial {y'}}(y,y')\neq 0$.

\begin{proposition} \label{thm:uniqueness}
Let $Q$ denote the Gram matrix $\frac{1}{n} \sum_{i=1}^n x_i x_i^\top$. We consider
the optimization problem in \myeq{minF} with $\mu > 0$.
If $Q$ is invertible or if $\{1,\dots,p\}$ belongs to $\G$, then
the problem in \myeq{minF} admits a unique solution.
\end{proposition}

Note that the invertibility of the matrix $Q$ requires $p\le n$. For
high-dimensional settings, the uniqueness of the solution will hold when $\{1,\dots,p\}$ belongs to $\G$,
or as further discussed at the end of the proof, as soon as for any $j,k\in\{1,\dots,p\}$, 
there exists a group $G\in\G$ which contains both $j$ and $k$.
Adding the group $\{1,\dots,p\}$ to $\G$ will in general not modify $\Pattern$ (and $\ZPattern$), 
but it will cause $\G$ to lose its minimality (in a sense introduced in the next subsection).
Furthermore, adding the full group $\{1,\dots,p\}$ has to be put in parallel with the equivalent (up to the squaring) $\ell_2$-norm term in the elastic-net penalty \citep{zou2005rav}, whose effect is to notably ensure strong convexity.
For more sophisticated uniqueness conditions that we have not explored here, 
we refer the readers to \citet[Theorem 1, 4 and 5]{osborne2000lasso}, \citet[Theorem 5]{rosset2004boosting} or
\citet[Theorem 3]{dossal2007} in the Lasso case,
and \citet{roth} for the group Lasso setting. 
We now turn to the result about the zero patterns of the solution of the problem in \myeq{minF}:

\begin{theorem}\label{thm:stability}
Assume that $Y=(y_1,\dots,y_n)^{\top}$ is a realization of an absolutely continuous probability distribution.
Let $k$ be the maximal number such that any $k$ rows of the matrix 
$(x_1,\dots,x_n) \!\in\! \RR{p}{n}$
are linearly independent.
For $\mu>0$, any solution of the problem in \myeq{minF} with at most $k-1$ nonzero coefficients has
a zero pattern in $\ZPattern=\left\lbrace \bigcup_{G\in\G'}G;\  \G' \subseteq \G \right\rbrace$ almost surely.
\end{theorem}

In other words, when $Y=(y_1,\dots,y_n)^{\top}$ is a realization of an absolutely continuous probability distribution,
the sparse solutions have a zero pattern in $\ZPattern=\left\lbrace \bigcup_{G\in\G'}G;\  \G' \subseteq \G \right\rbrace$ almost surely.
As a corollary of our two results, if the Gram matrix $Q$ is invertible, the problem in \myeq{minF} has a unique solution, 
whose zero pattern belongs to $\ZPattern$ almost surely.
Note that with the assumption made on $Y$, Theorem~\ref{thm:stability} is not directly applicable to the classification setting.
Based on these previous results, we can look at the following usual special cases from \mysec{regularized_risk_minimization} (we give more examples in \mysec{examples}):
 \BIT
 \item \textbf{$\ell_2$-norm}: the set of allowed nonzero patterns is composed of the empty set and the full set  $\IntegerSet{p}$.
\item \textbf{$\ell_1$-norm}: $\Pattern$ is the set of all possible subsets.
\item \textbf{$\ell_2$-norm and $\ell_1$-norm}: $\Pattern$ is also the set of all possible subsets.
\item \textbf{Group $\ell_1$-norm}: $\Pattern=\ZPattern$ is the set of all possible unions of the elements of the partition defining $\G$.
\item \textbf{Hierarchical norms}:  the set of patterns $\Pattern$ is then all sets $J$ for which all ancestors of elements in $J$ are included in $J$~\citep{hkl}.
 \EIT
Two natural questions now arise: (1)  starting from the groups $\G$, is there an efficient way to generate the set of nonzero patterns~$\Pattern$; (2) conversely, and more importantly, given $\Pattern $, how can the groups~$\G$---and hence the norm $\Omega(w)$---be designed?

\subsection{General Properties of $\G$, $\ZPattern$ and $\Pattern$}

We now study the different properties of the set of groups $\G$ and its corresponding sets of patterns $\ZPattern$ and $\Pattern$.

\paragraph{Closedness.}
The set of zero patterns $\ZPattern$ (respectively, the set of nonzero patterns~$\Pattern$) is closed under union (respectively, intersection), that is, for all $K \in \mathbb{N}$ and all $z_1,\dots,z_K \in \ZPattern,\ \bigcup_{k=1}^K z_k \in \ZPattern$ (respectively, $p_1,\dots,p_K \in \Pattern,\ \bigcap_{k=1}^K p_k \in \Pattern$). 
This implies that when ``reverse-engineering'' the set of nonzero patterns, we have to assume it is closed under intersection. Otherwise, the best we can do is to deal with its intersection-closure.
For instance, if we consider a sequence (see \myfig{sequence}), we cannot take $\Pattern$ to be the set of contiguous patterns with length two, 
since the intersection of such two patterns may result in a singleton (that does not belong to $\Pattern$).

\paragraph{Minimality.}
If a group in $\G$ is the union of other groups, it may be removed from $\G$ without changing the sets $\ZPattern$ or~$\Pattern$. This is the main argument behind the pruning backward algorithm in \mysec{backward}. Moreover, this leads to the notion of a \emph{minimal} set $\G$ of groups, which is such that
 for all $\G' \subseteq \ZPattern$ whose union-closure spans $\ZPattern$, we have $\G \subseteq \G'$. The existence and uniqueness of a minimal set is a consequence of classical results in set theory
\citep{doignon1998ks}. The elements of this minimal set are usually referred to as the \emph{atoms} of $\ZPattern$.

\begin{figure}
\begin{center}
\includegraphics[scale=.55]{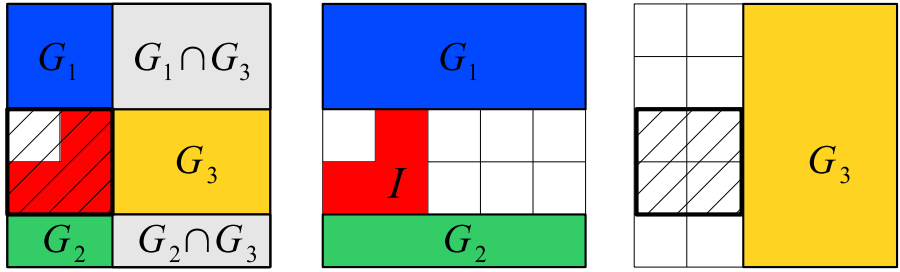}
\end{center}

\DistanceToCaption

\caption{$\G$-adapted hull: the pattern of variables $I$ (left and middle, in red) and its hull (left and right, hatched square) that is defined by the complement of the union of groups that do not intersect $I$, i.e., $(G_1 \cup G_2 \cup G_3)^c$.}
\label{fig:hull_definition}
 \end{figure}
 
Minimal sets of groups are attractive in our setting because they lead to a smaller number of groups and lower computational complexity---for example, for 2 dimensional-grids with rectangular patterns, we have a quadratic possible number of rectangles, i.e., $|\ZPattern|=O(p^2)$, that can be generated by a minimal set $\G$ whose size is $|\G|=O(\sqrt{p})$.

\paragraph{Hull.}
Given a set of groups $\G$, we can define for any subset $I \subseteq \IntegerSet{p}$ the \emph{$\G$-adapted hull}, or simply \emph{hull}, as:
$$
 \mathrm{Hull}(I) = \bigg\lbrace  \bigcup_{ G \in \G,\ G \cap I = \varnothing} \!\!\!\! G \bigg\rbrace ^c,
$$
which is the smallest set in $\Pattern$ containing $I$ (see \myfig{hull_definition}); we always have $I \subseteq \mathrm{Hull}(I)$ with equality if and only if 
$I \in \Pattern$. 
The hull has a clear geometrical interpretation for specific sets $\G$ of groups.
For instance, if the set $\G$ is formed by all vertical and horizontal half-spaces when the variables are organized in a 2 dimensional-grid (see \myfig{axis-aligned}), the hull of a subset $I\subset\{1,\dots,p\}$ is simply the axis-aligned bounding box of $I$. Similarly, when $\G$ is the set of all half-spaces with all possible orientations
(e.g., orientations $\pm\pi/4$ are shown in \myfig{convex}), the hull becomes the regular convex hull\footnote{We use the term \emph{convex} informally here. It can however be made precise with the notion of convex subgraphs \citep{chung1997sgt}.}.
Note that those interpretations of the hull are possible and valid only when we have geometrical information at hand about the set of variables. 

\begin{figure}
\begin{center}
\includegraphics[scale=.45]{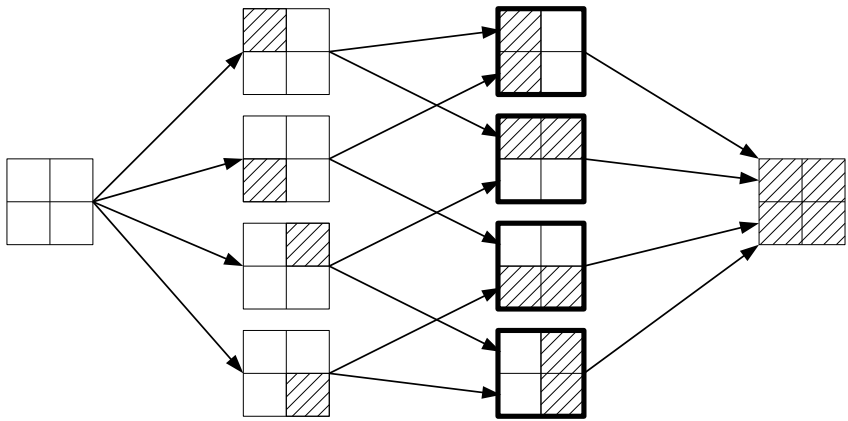}
\end{center}

\DistanceToCaption

\caption{  The DAG for the set $\ZPattern$ associated with the $2\! \times \!2$-grid. The members of $\ZPattern$ are the complement of the areas hatched in black. The elements of $\G$
 (i.e., the atoms of $\ZPattern$) are highlighted by bold edges. }
\label{fig:DAG}
 \end{figure}

\paragraph{Graphs of patterns.}
We consider the directed acyclic graph (DAG) stemming from the \emph{Hasse diagram}~\citep{cameron1994ctt} of the partially ordered set (poset) $(\G, \supset)$. By definition, the nodes of this graph are the elements $G$ of $\G$ and there is a directed edge from $G_1$ to $G_2$ if and only if $G_1 \supset G_2$ and there exists no $G \in \G$ such that $G_1 \supset G \supset G_2$~\citep{cameron1994ctt}. We can also build the corresponding DAG for the set of zero patterns $\ZPattern \supset \G$, which is a super-DAG of the DAG of groups (see \myfig{DAG} for examples). Note that we obtain also the isomorphic DAG for the nonzero patterns $\Pattern$, although it corresponds to the poset $(\Pattern, \subset)$: this DAG will be used in the active set algorithm presented in \mysec{optimization}.

Prior works with nested groups~\citep{cap,hkl,kim2009tree} have also used a similar DAG structure, 
with the slight difference that in these works, the corresponding hierarchy of variables is built from the prior knowledge about the problem at hand 
(e.g., the tree of wavelets in \citet{cap}, the decomposition of kernels in \citet{hkl} or the hierarchy of genes in \citet{kim2009tree}).
The DAG we introduce here on the set of groups naturally and always comes up, with no assumption on the variables themselves (for which no DAG is defined in general).

\subsection{From Patterns to Groups}
\label{sec:backward}
We now assume that we want to impose a priori knowledge on the sparsity structure of a solution $\hat{w}$ of our regularized problem in \myeq{minF}. 
This information can be exploited by restricting the patterns allowed by the norm $\Omega$. Namely, from an intersection-closed set of zero patterns $\ZPattern$, we can build back a minimal set of groups $\G$ by iteratively pruning away in the DAG corresponding to $\ZPattern$, all sets  which are unions of their parents. See Algorithm~\ref{alg:backward}.
This algorithm can be found under a different form in ~\citet{doignon1998ks}---we present it through a pruning algorithm on the DAG, which is natural in our context (the proof of the minimality of the procedure can be found in Appendix~\ref{app:backward}). The complexity of Algorithm~\ref{alg:backward} is 
$O( p | \ZPattern | ^2)$. 
The pruning may reduce significantly the number of groups necessary to generate the whole set of zero patterns, sometimes from exponential in $p$ to polynomial in $p$ (e.g., the $\ell_1$-norm). In \mysec{examples}, we give other examples of interest where $|\G|$ (and $|\Pattern|$) is also polynomial in~$p$.
 
 \begin{algorithm}
   \caption{  {Backward} procedure}
   \label{alg:backward}
\begin{algorithmic}
   \STATE {\bfseries Input:}  Intersection-closed family of nonzero patterns~$\Pattern$.
   \STATE {\bfseries Output:}  Set of groups $\G$.
   \STATE {\bfseries Initialization:} Compute $\ZPattern=\left\lbrace P^c;\ P \in \Pattern \right\rbrace$ and set $\G = \ZPattern$.\\
    Build the Hasse diagram for the poset $(\ZPattern, \supset)$.
    \FOR{$t=\min_{G\in\ZPattern}|G|$ {\bfseries to} $\max_{G\in\ZPattern}|G|$}
    \FOR{{each} node $G\in\ZPattern$ such that $|G|=t$}
     \IF{   
  $\left(\bigcup_{C \in \textrm{Children}(G)}C=G \right)$
  }
  \IF{  $\left(\textrm{Parents}(G) \neq \varnothing \right)$ }
  \STATE Connect   children of $G$ to   parents of $G$.
  \ENDIF
  \STATE Remove $G$ from $\G$.
  \ENDIF
  \ENDFOR 
   \ENDFOR
\end{algorithmic}
\end{algorithm}

\begin{algorithm}
   \caption{  {Forward} procedure}
   \label{alg:forward}
\begin{algorithmic}
   \STATE {\bfseries Input:} Set of groups $\G=\{G_1,\dots,G_M\}$.
   \STATE {\bfseries Output:} Collection of zero patterns $\ZPattern$ and nonzero patterns $\Pattern$.
   \STATE {\bfseries Initialization:} $\ZPattern=\{\varnothing\}$.
      \FOR{$m=1$ {\bfseries to} $M$}
   \STATE $C = \{\varnothing\}$
   \FOR{\textbf{each} $Z \in \ZPattern$}
    \IF{$\left(G_m \nsubseteq Z\right)$ \textbf{ and }
   $( \forall G \in \! \{G_1,\dots,G_{m-1}\},\ G \subseteq Z\cup G_m \Rightarrow G \subseteq Z  )$}
  \STATE $C \leftarrow C \cup \{ Z \cup G_m \}$ .
  \ENDIF
  \ENDFOR \\
  $\ZPattern \leftarrow \ZPattern \cup C$.
   \ENDFOR \\
   $\Pattern=\left\lbrace Z^c;\ Z \in \ZPattern \right\rbrace$.
\end{algorithmic}
\end{algorithm}

\subsection{From Groups to Patterns}
The \textit{forward} procedure presented in Algorithm~\ref{alg:forward}, taken from~\citet{doignon1998ks}, allows the construction of $\ZPattern$ from $\G$. It iteratively builds the collection of patterns by taking unions, and has
complexity  $O(p |\ZPattern| |\G|^2)$. The general scheme is straightforward. Namely, by considering increasingly larger sub-families of $\G$ and the collection of patterns already obtained, all possible unions are formed. However, some attention needs to be paid while checking we are not generating a pattern already encountered. Such a verification is performed by the \textit{if} condition within the inner loop of the algorithm. Indeed, we do not have to scan the whole collection of patterns already obtained (whose size can be exponential in $|\G|$), but we rather use the fact that $\G$ generates $\ZPattern$. Note that in general, it is not possible to upper bound the size of 
 $|\ZPattern|$ by a polynomial term in $p$, even when $\G$ is very small  (indeed,
 $|\ZPattern| = 2^p$ and $|\G|=p$ for the $\ell_1$-norm).

\subsection{Examples}\label{sec:examples}

We now present several examples of sets of groups $\G$, especially suited to encode geometric and temporal prior information.

\begin{figure}[!ht]
\begin{center}
\includegraphics[scale=.6]{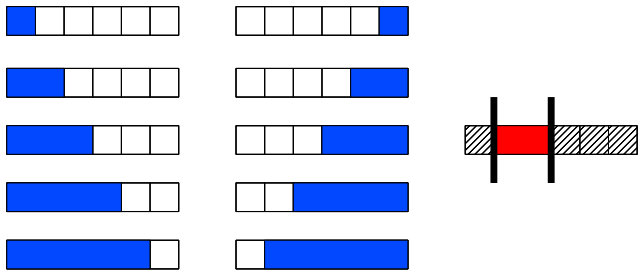}
\end{center}

\DistanceToCaption

\caption{ (Left) The set of blue groups to penalize in order to select contiguous patterns in a sequence.
(Right) In red, an example of such a nonzero pattern with its corresponding zero pattern (hatched area). } 
\label{fig:sequence}
 \end{figure}

\paragraph{Sequences.}
Given $p$ variables organized in a sequence, if we want only contiguous nonzero patterns, the backward algorithm will lead to the set of groups which are intervals $[1,k]_{k \in \{1,\dots,p-1\}}$ and $[k,p]_{k \in \{2,\dots,p\}}$, with both $|\ZPattern|=O(p^2)$ and $|\G|=O(p)$ (see Figure~\ref{fig:sequence}).
Imposing the contiguity of the nonzero patterns is for instance relevant for the diagnosis of tumors, based on the profiles of arrayCGH \citep{Rapaport2008Classification}.

\UpperDistanceToFig

\begin{figure}[!ht]
\begin{center}
\includegraphics[scale=.5]{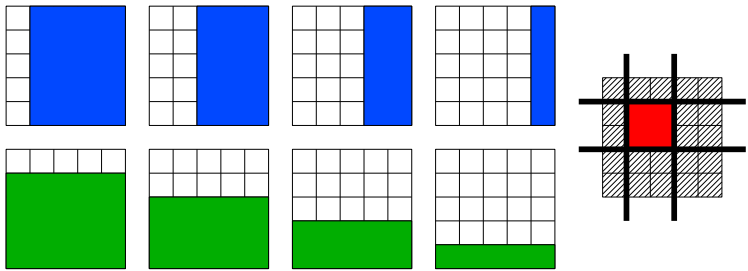}
\end{center}

\DistanceToCaption

\caption{ Vertical and horizontal groups: (Left) the set of blue and green groups with their (not displayed) complements to penalize in order to select
rectangles. (Right) In red, an example of nonzero pattern recovered in this setting, with its corresponding zero pattern (hatched area). } 
\label{fig:axis-aligned}
\end{figure}

\paragraph{Two-dimensional grids.}
In \mysec{experiments}, we notably consider for $\Pattern$ the set of all rectangles in two dimensions, leading by the previous algorithm to the set of axis-aligned half-spaces for~$\G$ (see \myfig{axis-aligned}),
with $|\ZPattern|=O(p^2)$ and $|\G|=O(\sqrt{p})$.
This type of structure is encountered in object or scene recognition, where the selected rectangle would correspond to a certain box inside an image, that concentrates the predictive power for a given class of object/scene \citep{cordelia_image}.

\begin{figure}[!ht]
\begin{center}
\includegraphics[scale=.5]{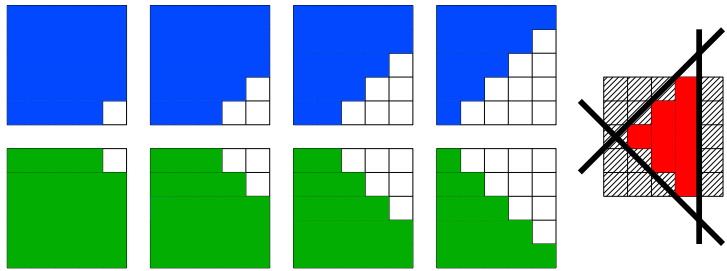}
\end{center}

\DistanceToCaption

\caption{   Groups with $\pm \pi/4$ orientations: (Left) the set of blue and green groups with their (not displayed) complements to penalize in order to select
diamond-shaped patterns. (Right) In red, an example of nonzero pattern recovered in this setting, with its corresponding zero pattern (hatched area). } 
\label{fig:convex}
 \end{figure}

Larger set of convex patterns can be obtained by adding in $\mathcal{G}$ half-planes with other orientations than vertical and horizontal.
For instance, if we use planes with angles that are multiples of $\pi/4$, the nonzero patterns of $\mathcal{P}$ can have polygonal shapes with up to 8 faces.
In this sense, if we keep on adding half-planes with finer orientations, the nonzero patterns of $\mathcal{P}$ can be described by polygonal shapes with an increasingly larger number of faces.
The standard notion of convexity defined in $\mathbb{R}^2$ would correspond to the situation where an infinite number of orientations is considered~\citep{soille}. See \myfig{convex}.
The number of groups is linear in $\sqrt{p}$ with constant growing linearly with the number of angles, while $|\ZPattern|$ grows more rapidly (typically non-polynomially in the number of angles). 
Imposing such convex-like regions turns out to be useful in computer vision. 
For instance, in face recognition, it enables the design of localized features that improve upon the robustness to occlusions \citep{SparseStructuredPCA}.

\UpperDistanceToFig

\paragraph{Extensions.}

The sets of groups presented above can be straightforwardly extended to more complicated topologies, such as three-dimensional spaces discretized in cubes or spherical volumes discretized in slices.
Similar properties hold for such settings. For instance, if all the axis-aligned half-spaces are considered for~$\G$ in a three-dimensional space, 
then $\Pattern$ is the set of all possible rectangular boxes with $|\Pattern|=O(p^2)$ and $|\G|=O(p^{1/3})$.
Such three-dimensional structures may be interesting to retrieve discriminative and local sets of voxels from fMRI/MEEG responses \citep{kowalski-gramfort09, xiang-boosting}.
Moreover, while the two-dimensional rectangular patterns described previously are adapted to find bounding boxes in static images \citep{cordelia_image}, 
scene recognition in videos requires to deal with a third temporal dimension \citep{cordelia_video}.
This may be achieved by designing appropriate sets of groups, embedded in the three-dimensional space obtained by tracking the frames over time.

\paragraph{Representation and computation of $\G$.}
The sets of groups described so far can actually be represented in a same form,
that lends itself well to the analysis of the next section.
When dealing with a discrete sequence of length $l$ (see \myfig{sequence}), we have
\begin{eqnarray*}
\G &=& \{ g_-^k;\; k \in \{1,\dots,l\!-\!1\}\} \cup \{ g_+^k;\; k \in \{2,\dots,l\} \},\\
   &=& \G_{\text{left}} \cup \G_{\text{right}}, 
\end{eqnarray*}
with $g_-^k = \{i;\; 1 \leq i \leq k\}$ and $g_+^k = \{i;\; l \geq i \geq k\}$.
In other words, the set of groups $\G$ can be rewritten as a partition\footnote{Note the subtlety: the sets $\G_\theta$ are disjoint, that is $\G_\theta \cap \G_{\theta'}=\varnothing$ for $\theta\neq\theta'$, but groups in $\G_\theta$ and $\G_{\theta'}$ can overlap.}
in two sets of nested groups, $\G_{\text{left}}$ and $\G_{\text{right}}$.

The same goes for a two-dimensional grid, with dimensions $h\!\times\!l$ (see \myfig{axis-aligned} and \myfig{convex}).
In this case, the nested groups we consider are defined based on the following groups of variables
$$
g^{k,\theta}=\{(i,j)\in\{1,\dots,l\}\times\{1,\dots,h\};\ \cos(\theta)i+\sin(\theta)j \leq k \},
$$
where $k \in \mathbb{Z}$ is taken in an appropriate range.

The nested groups we obtain in this way are therefore
parameterized by an angle\footnote{Due to the discrete nature of the underlying geometric structure of $\G$, angles $\theta$ that are not multiple of $\pi/4$ (i.e., such that $\tan(\theta)\notin\mathbb{Z}$) are dealt with by rounding operations.} $\theta$, $\theta \in (-\pi;\pi]$.
We refer to this angle as an \textit{orientation}, since it defines the normal vector 
$\tbinom{\cos(\theta)}{\sin(\theta)}$ to the line $\{(i,j)\in\R{2};\cos(\theta)i+\sin(\theta)j=k\}$.
In the example of the rectangular groups (see \myfig{axis-aligned}), we have four orientations, with $\theta\in\{0,\pi/2,-\pi/2,\pi\}$.
More generally, if we denote by $\Theta$ the set of the orientations, we have
$$
\G=\bigcup_{\theta \in \Theta}\!\G_\theta,
$$
where $\theta\in\Theta$ indexes the partition of $\G$ in sets $\G_\theta$ of nested groups of variables.
Although we have not detailed the case of $\R{3}$, we likewise end up with a similar partition of $\G$.

\section{Optimization and Active Set Algorithm}\label{sec:optimization}

For moderate values of $p$, one may obtain a solution for \myeq{minF} using generic toolboxes for second-order cone programming (SOCP) whose time complexity is equal to
$O(p^{3.5} + | \G|^{3.5})$ \citep{boyd}, which is not appropriate when $p$ or $|\G|$ are large. 
This time complexity corresponds to the computation of \myeq{minF} for a single value of the regularization parameter $\mu$.

We present in this section an \emph{active set algorithm} (Algorithm~\ref{alg:activeset}) that finds a solution for \myeq{minF} by considering increasingly larger active sets and checking global optimality at each step. 
When the rectangular groups are used, the total complexity of this method is in $O(s\max\{p^{1.75}\!,s^{3.5}\})$, where $s$ is the size of the active set at the end of the optimization. Here, the sparsity prior is exploited for computational advantages.
Our active set algorithm needs an underlying \textit{black-box} SOCP solver; 
in this paper, 
we consider both a first order approach (see \myappendix{eta_trick}) and a SOCP method\footnote{The C$++$/Matlab code used in the experiments may be downloaded from the authors website.} --- in our experiments, we use \texttt{SDPT3} \citep{SDPT3_bis, SDPT3}.
Our active set algorithm extends to general overlapping groups the work of \citet{hkl}, 
by further assuming that it is computationally possible to have a time complexity polynomial in the number of variables $p$.

We primarily focus here on finding an efficient active set algorithm; 
we defer to future work the design of specific SOCP solvers, e.g., based on proximal techniques \citep[see, e.g.,][and numerous references therein]{tseng_review},
adapted to such non-smooth sparsity-inducing penalties.
\subsection{Optimality Conditions: from Reduced Problems to Full Problems}

It is simpler to derive the algorithm for the following regularized optimization problem\footnote{It is also possible to derive the active set algorithm for the constrained formulation
$\min_{w \in \R{p}}  \ERisk{y_i}{w ^\top x_i} \quad \mbox{such that} \quad \Omega(w) \leq \lambda$. However, we empirically found it more difficult to select $\lambda$ in this latter formulation.}
which has the same solution set as the regularized problem of \myeq{minF} when $\mu$ and $\lambda$ are allowed to vary \citep[see Section 3.2]{borwein2006caa}:
 
\begin{equation} \label{eq:minFc}
 \min_{w \in \R{p}}  \ERisk{y_i}{w ^\top x_i} + \frac{\lambda}{2} \left[ \Omega(w) \right]^2.
\end{equation}

In active set methods, the set of nonzero variables, denoted by $J$, is built incrementally, and the problem is solved only for this reduced set of variables, adding the constraint $w_{J^c}=0$ to \myeq{minFc}. 
In the subsequent analysis, we will use arguments based on duality to monitor the optimality of our active set algorithm.
We denote by $L(w) =  \ERisk{y_i}{w ^\top x_i}$ the empirical risk (which is by assumption convex and continuously differentiable) and by $L^\ast$ its \emph{Fenchel-conjugate}, defined as \citep{boyd, borwein2006caa}:
$$
L^\ast(u)=\sup_{w\in\R{p}} \{w^\top u - L(w)\}.
$$
The restriction of $L$ to $\R{|J|}$ is denoted $L_J(w_J)=L(\tilde{w})$ for
$\tilde{w}_J = w_J $ and $\tilde{w}_{J^c}=0$, with Fenchel-conjugate $L_J^\ast$. 
Note that, as opposed to $L$, we do not have in general $L_J^\ast(\kappa_J) = L^\ast(\tilde{\kappa})$ for 
$\tilde{\kappa}_J = \kappa_J $ and $\tilde{\kappa}_{J^c}=0$.

For a potential active set $J \subset \IntegerSet{p}$ which belongs to the set of allowed nonzero patterns~$\Pattern$, we denote by $\G_J$ the set of active groups, i.e., the set of groups $G \in \G$ such that $G \cap J \neq \varnothing$.
We consider the reduced norm $\Omega_{J}$ defined on $\R{|J|}$ as
$$
\Omega_{J}(w_J)=\sum_{ G\in\G  } \NormDeux{\dG_J \circ w_J} = \sum_{ G\in\G_J  } \NormDeux{\dG_J \circ w_J},
$$
and its \emph{dual norm}
$
\Omega_J^*(\kappa_J) = \max_{\Omega_{J}(w_J)  \leq 1} w_J ^\top \kappa_J
$, also defined on $\R{|J|}$. 
The next proposition (see proof in \myappendix{dual_problem}) gives the optimization problem dual to the reduced problem (\myeq{reduced-primal} below):

\begin{proposition}[Dual Problems]\label{prop:dual_problem} Let $J \subseteq \IntegerSet{p}$. The following two problems
 \begin{equation}
 \label{eq:reduced-primal}
 \min_{w_J \in \R{|J|}}  L_J(w_J) + \frac{\lambda}{2} \left[ \Omega_J(w_J) \right]^2,
 \end{equation}
 \begin{equation}
\label{eq:reduced-dual}
 \max_{\kappa_{J} \in \R{|J|}} \  -L_J^*(-\kappa_{J})-\frac{1}{2\lambda} \left[ \Omega_{J}^*(\kappa_{J}) \right]^2,
 \end{equation} 
 are dual to each other and strong duality holds.
 The pair of primal-dual variables $\{w_J,\kappa_J\}$ is optimal if and only if we have 
 $$
 \begin{cases}
 \kappa_J           & = -\nabla L_J(w_J), \\
 w_J ^\top \kappa_J & = \frac{1}{\lambda} \left[ \Omega_J^*(\kappa_J) \right]^2 = \lambda \left[ \Omega_J(w_J) \right]^2.
 \end{cases}
 $$

\end{proposition}
As a brief reminder, the duality gap of a minimization problem is defined as the difference between the primal and dual objective functions,
evaluated for a feasible pair of primal/dual variables \citep[see Section 5.5]{boyd}.
This gap serves as a certificate of (sub)optimality: if it is equal to zero, then the optimum is reached, 
and provided that strong duality holds, the converse is true as well \citep[see Section 5.5]{boyd}.

The previous proposition enables us to derive the duality gap for the optimization problem \myeq{reduced-primal}, 
that is reduced to the active set of variables $J$.
In practice, this duality gap will always vanish (up to the precision of the underlying SOCP solver), 
since we will sequentially solve \myeq{reduced-primal} for increasingly larger active sets $J$.
We now study how, starting from the optimality of the problem in \myeq{reduced-primal}, we can control the optimality, 
or equivalently the duality gap, for the full problem \myeq{minFc}.
More precisely, the duality gap of the optimization problem \myeq{reduced-primal} is
\begin{eqnarray*}
&   & L_J(w_J)+L_J^*(-\kappa_{J})+\frac{\lambda}{2} \left[ \Omega_J(w_J) \right]^2  +\frac{1}{2\lambda} \left[ \Omega_{J}^*(\kappa_{J}) \right]^2  \\
& = & \left\lbrace L_J(w_J) +  L_J^*(-\kappa_{J}) + w_J ^\top \kappa_J\right\rbrace 
+ \left\lbrace \frac{\lambda}{2} \left[ \Omega_J(w_J) \right]^2  +\frac{1}{2\lambda} \left[ \Omega_{J}^*(\kappa_{J}) \right]^2 - w_J ^\top \kappa_J\right\rbrace  ,
\end{eqnarray*}
which is a sum of two nonnegative terms, the nonnegativity coming from the Fenchel-Young inequality \citep[Proposition 3.3.4 and Section 3.3.2 respectively]{borwein2006caa, boyd}.
We can think of this duality gap as the sum of two duality gaps, respectively relative to $L_J$ and $\Omega_J$.
Thus, if we have a primal candidate $w_J$ and we choose $\kappa_J=-\nabla L_J(w_J)$, the duality gap relative to $L_J$ vanishes and the total duality gap then reduces to 
$$
\frac{\lambda}{2} \left[ \Omega_J(w_J) \right]^2  +\frac{1}{2\lambda} \left[ \Omega_{J}^*(\kappa_{J}) \right]^2 - w_J ^\top \kappa_J.
$$

In order to check that the reduced solution $w_J$ is optimal for the full problem in \myeq{minFc}, we pad $w_J$ with zeros on ${J^c}$ to define $w$ and compute $\kappa = - \nabla L(w)$, which is such that $\kappa_J = - \nabla L_J(w_J)$.
For our given candidate pair of primal/dual variables $\{w,\kappa\}$, we then get a duality gap for the full problem in \myeq{minFc} equal to

\begin{eqnarray*}
&&  \frac{\lambda}{2} \left[ \Omega(w) \right]^2  +\frac{1}{2\lambda} \left[ \Omega^*(\kappa) \right]^2 - w ^\top \kappa  \\ 
&=& \frac{\lambda}{2} \left[ \Omega(w) \right]^2  +\frac{1}{2\lambda} \left[ \Omega^*(\kappa) \right]^2 - w_J ^\top \kappa_J \\
&=& \frac{\lambda}{2} \left[ \Omega(w) \right]^2  +\frac{1}{2\lambda} \left[ \Omega^*(\kappa) \right]^2  - \frac{\lambda}{2} \left[ \Omega_J(w_J) \right]^2  - \frac{1}{2\lambda} \left[ \Omega_{J}^*(\kappa_{J}) \right]^2 \\
&=& \frac{1}{2\lambda} \left( \left[ \Omega^*(\kappa) \right]^2 - \left[ \Omega_{J}^*(\kappa_{J}) \right]^2 \right) \\
&=& \frac{1}{2\lambda} \left( \left[ \Omega^*(\kappa) \right]^2 - \lambda w_J^\top\kappa_J \right).
\end{eqnarray*}
Computing this gap requires computing the dual norm which itself is as hard as the original problem, 
prompting the need for upper and lower bounds on $\Omega^*$ 
(see Propositions~\ref{prop:CN} and \ref{prop:CS} for more details).

\UpperDistanceToFig

\begin{figure}[!ht]
\begin{center}
\includegraphics[scale=.5]{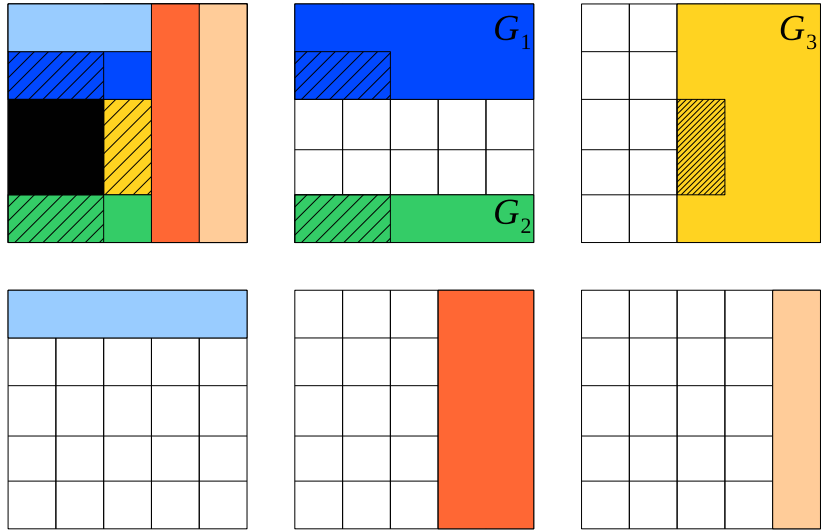}
\end{center}

\DistanceToCaption

\caption{ The active set (black) and the candidate patterns of variables, i.e. the variables in $K \backslash J$  (hatched in black) that can become active. The fringe groups are exactly the groups that have the hatched areas (i.e., here we have $\mathcal{F}_J = \bigcup_{K \in \Pi_\Pattern(J)} \G_K \backslash \G_J=\{G_1,G_2,G_3\}$).}
\label{fig:fringe_example_with_rect}
 \end{figure}

\begin{figure}[!ht]
\begin{center}
\includegraphics[scale=.55]{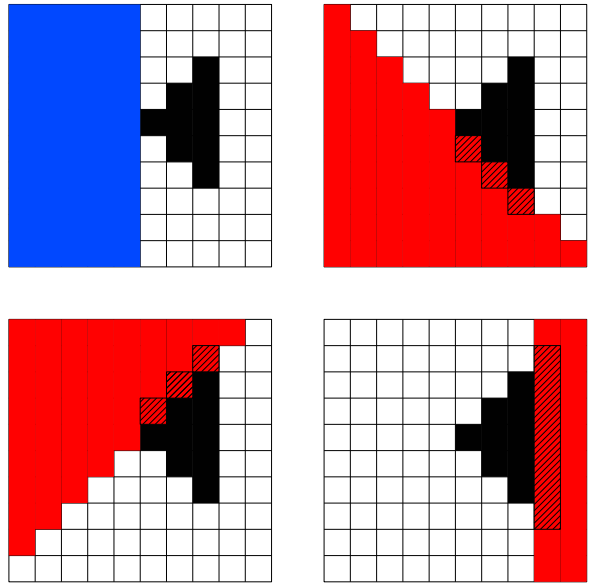}
\end{center}

\DistanceToCaption

\caption{ The active set (black) and the candidate patterns of variables, i.e. the variables in $K \backslash J$  (hatched in black) that can become active. The groups in red are those in $\bigcup_{K \in \Pi_\Pattern(J)} \G_K \backslash \G_J$, while the blue group is in $\mathcal{F}_J \backslash ( \bigcup_{K \in \Pi_\Pattern(J)} \G_K \backslash \G_J)$. The blue group does not intersect with any patterns in $\Pi_\Pattern(J)$.}
\label{fig:fringe_example_with_angle}
 \end{figure}

\subsection{Active set algorithm}

We can interpret the active set algorithm as a walk through the DAG of nonzero patterns allowed by the norm $\Omega$.  
The parents $\Pi_\Pattern(J)$ of $J$ in this DAG are exactly the patterns containing the variables that may enter the active set at the next iteration of Algorithm~\ref{alg:activeset}.
The groups that are exactly at the boundaries of the active set (referred to as the \emph{fringe groups}) are 
$
\mathcal{F}_J = \{G\in (\G_J)^c\ ;\ \nexists G' \in (\G_J)^c,\ G \subseteq G' \}
$, i.e., the groups that are not contained by any other inactive groups.

In simple settings, e.g., when $\G$ is the set of rectangular groups, 
the correspondence between groups and variables is straightforward since we have
$
\mathcal{F}_J=\bigcup_{K \in \Pi_\Pattern(J)} \G_K \backslash \G_J
$ (see \myfig{fringe_example_with_rect}).
However, in general, we just have the inclusion $(\bigcup_{K \in \Pi_\Pattern(J)} \G_K \backslash \G_J) \subseteq \mathcal{F}_J$ and some elements of $\mathcal{F}_J$ might not correspond to  
any patterns of variables in $\Pi_\Pattern(J)$ (see \myfig{fringe_example_with_angle}).

We now present the optimality conditions (see proofs in \myappendix{CN_CS}) that monitor the progress of Algorithm~\ref{alg:activeset}:
\begin{proposition}[Necessary condition]
\label{prop:CN}
 If $w$ is optimal for the full problem in \myeq{minFc}, then
\begin{equation}\label{eq:CNlowerbound} 
 \max_{K \in \Pi_\Pattern(J)}   \dfrac{\NormDeux{\nabla L(w)_{K \backslash J } } }{\sum_{ H \in \G_K \backslash \G_J } \NormInf{ \dH_{ \scriptscriptstyle  K \backslash J  } } }
                                \leq  \big\{ \!\! -\lambda w ^\top \nabla L(w) \big\}^{\frac{1}{2}}. 
\tag{$N$}
\end{equation}
\end{proposition}

\begin{proposition}[Sufficient condition]
\label{prop:CS}If
\begin{equation}\label{eq:CSupperbound}
\max_{ G \in \mathcal{F}_J }
                \left\{ 
                \sum_{k \in G}  \bigg\{ \dfrac{ \nabla L(w)_k}{ \sum_{  H \ni k,\ H \in (\G_J)^c  }  \dH_k }  \bigg\}^2  
                \right\}^{\frac{1}{2}}
\!
\leq  \big\{ \lambda ( 2\varepsilon -  w ^\top \nabla L(w) ) \big\}^{\frac{1}{2}},
\tag{$S_\varepsilon$}
\end{equation}
then $w$ is an approximate solution for \myeq{minFc} whose duality gap is less than $\varepsilon\geq0$.
\end{proposition}

Note that for the Lasso, the conditions $(N)$ and $(S_0)$ (i.e., the sufficient condition taken with $\varepsilon=0$) are both equivalent (up to the squaring of $\Omega$) to the condition $\| \nabla L(w)_{J^c} \|_\infty \leq  -  w ^\top \nabla L(w)$, which is the usual optimality condition~\citep{martin,Tibshirani96regressionshrinkage}.
Moreover, when they are not satisfied, our two conditions provide good heuristics for choosing which $K \in  \Pi_\Pattern(J)$ should enter the active set.

More precisely, since the necessary condition~(\ref{eq:CNlowerbound}) directly deals with the \emph{variables} (as opposed to groups) that can become active at the next step of Algorithm~\ref{alg:activeset}, it suffices to choose the pattern~$K \in \Pi_\Pattern(J)$ that violates most the condition.

The heuristics for the sufficient condition~(\ref{eq:CSupperbound}) implies to go from groups to variables. 
We simply consider the group $G \in \mathcal{F}_J$ that violates most the sufficient condition and then take all the patterns of variables $K \in  \Pi_\Pattern(J)$ such that $\ K \cap G \neq \varnothing$ to enter the active set.
If $G \cap (\bigcup_{K \in \Pi_\Pattern(J)} K) = \varnothing$, we look at all the groups $H \in \mathcal{F}_J$ such that $H \cap G \neq \varnothing$ and apply the scheme described before (see Algorithm~\ref{alg:heuristics_for_sufficient_condition}).

A direct consequence of this heuristics is that it is possible for the algorithm to \textit{jump over} the right active set and to consider instead a (slightly) larger active set as optimal. However if the active set is larger than the optimal set, then (it can be proved that) the sufficient condition $(S_0)$ is satisfied, and the reduced problem, which we solve exactly, will still output the correct nonzero pattern. 

Moreover, it is worthwhile to notice that in Algorithm~\ref{alg:activeset}, the active set may sometimes be increased only to make sure that the current solution is optimal (we only check a sufficient condition of optimality).

\begin{algorithm}
   \caption{ Active set algorithm}
   \label{alg:activeset}
\begin{algorithmic}
   \STATE {\bfseries Input:} Data $\{(x_i,y_i)$, $i=1,\dots,n\}$, regularization parameter $\lambda$, \\
  \hspace*{1cm} Duality gap precision $\varepsilon$, maximum number of variables $s$.
   \STATE {\bfseries Output:} Active set $J$, loading vector $\hat{w}$.
   \STATE {\bfseries Initialization:} $J =\{\varnothing\}$, $\hat{w} = 0$.
         \WHILE{ $\big($ $(N)$ is not satisfied $\big)$ \textbf{and} $\big($ $|J|\leq s$ $\big)$ }
   \STATE Replace $J$ by violating $K \in \Pi_\Pattern(J)$ in (\ref{eq:CNlowerbound}).
   \STATE Solve the reduced problem
     $
	\, \min_{w_J \in \R{|J|}}  L_J(w_J) + \frac{\lambda}{2} \left[ \Omega_J(w_J) \right]^2 \,
     $
     to get $\hat{w}$.
     \ENDWHILE
    \WHILE{ $\big($ $(S_\varepsilon)$ is not satisfied $\big)$ \textbf{and} $\big($ $|J|\leq s$ $\big)$ }
   \STATE Update $J$ according to Algorithm~\ref{alg:heuristics_for_sufficient_condition}.
 
      \STATE Solve the reduced problem
     $
        \, \min_{w_J \in \R{|J|}}  L_J(w_J) + \frac{\lambda}{2} \left[ \Omega_J(w_J) \right]^2 \,
     $ 
     to get $\hat{w}$.
     \ENDWHILE\end{algorithmic}
\end{algorithm}

\paragraph{Convergence of the active set algorithm.}

The procedure described in Algorithm~\ref{alg:activeset} can terminate in two different states. 
If the procedure stops because of the limit on the number of active variables $s$, 
the solution might be suboptimal. Note that, in any case, we have at our disposal a upperbound on the duality gap.

Otherwise, the procedure always converges to an optimal solution, either
(1) by validating both the necessary and sufficient conditions (see Propositions~\ref{prop:CN} and \ref{prop:CS}), 
ending up with fewer than $p$ active variables and a precision of (at least) $\varepsilon$,
or (2) by running until the $p$ variables become active, the precision of the solution being given by the underlying solver.

\begin{algorithm}
   \caption{ Heuristics for the sufficient condition~(\ref{eq:CSupperbound})}
   \label{alg:heuristics_for_sufficient_condition}
	\begin{algorithmic}
	\STATE Let $G \in \mathcal{F}_J$ be the group that violates (\ref{eq:CSupperbound}) most.
	\IF{ $( G \cap (\bigcup_{K \in \Pi_\Pattern(J)} K) \neq \varnothing )$ }
		\FOR{ $K \in \Pi_\Pattern(J)$ such that $K \cap G \neq \varnothing$}
		\STATE $J \leftarrow J \cup K $.
		\ENDFOR
	\ELSE
		\FOR{ $H \in \mathcal{F}_J $ such that $H \cap G \neq \varnothing$}
			\FOR{ $K \in \Pi_\Pattern(J)$ such that $K \cap H \neq \varnothing$}
			\STATE $J \leftarrow J \cup K $.
			\ENDFOR
		\ENDFOR
  	\ENDIF
	\end{algorithmic}
\end{algorithm}

\paragraph{Algorithmic complexity.}

We analyze in detail the time complexity of the active set algorithm when we consider sets of groups 
$\G$ such as those presented in the examples of \mysec{examples}.
We recall that we denote by $\Theta$ the set of orientations in $\G$ (for more details, see \mysec{examples}).

For such choices of $\G$, the fringe groups $\mathcal{F}_J$ reduces to the largest groups of each orientation 
and therefore $|\mathcal{F}_J| \leq |\Theta|$. 
We further assume that the groups in $\G_\theta$ are sorted by cardinality, so that computing $\mathcal{F}_J$ costs $O(|\Theta|)$. 

Given an active set $J$, both the necessary and sufficient conditions require to have access to the direct parents $\Pi_\Pattern(J)$ of $J$ in the DAG of nonzero patterns. In simple settings, e.g., when $\G$ is the set of rectangular groups, this operation can be performed in $O(1)$ (it just corresponds to scan the (up to) four patterns at the edges of the current rectangular hull). 

However, for more general orientations, computing $\Pi_\Pattern(J)$ requires to find the smallest nonzero patterns that we can generate from the groups in $\mathcal{F}_J$, reduced to the stripe of variables around the current hull.
This stripe of variables can be computed as
$\big[\bigcup_{G \in (\G_J)^c \backslash \mathcal{F}_J} G \big]^c \backslash J$,
so that getting $\Pi_\Pattern(J)$ costs $O( s 2^{|\Theta|} + p|\G| )$ in total.

Thus, if the number of active variables is upper bounded by $s \! \ll \! p$ (which is true if our target is actually sparse), the time complexity of Algorithm~\ref{alg:activeset} is the sum of:
\begin{itemize}
 \item the computation of the gradient, $O(s n p)$ for the square loss.
 \item if the underlying solver called upon by the active set algorithm is a standard SOCP solver, $O(s \max_{ J \in \Pattern, |J| \leq s } |\G_J|^{3.5} + s^{4.5})$
 (note that the term $s^{4.5}$ could be improved upon by using warm-restart strategies for the sequence of reduced problems).
 \item $t_1$ times the computation of $(N)$, that is 
$O( t_1( s 2^{|\Theta|} + p|\G| + s n_\theta^2  ) + p|\G|) = O(t_1 p|\G|)$.

During the initialization (i.e., $J=\varnothing$), we have $|\Pi_\Pattern(\varnothing)|=O(p)$ (since we can start with any singletons), and  $|\G_K \backslash \G_J|=|\G_K|=|\G|$, which leads to a complexity of $O(p|\G|)$ for the sum $\sum_{G \in \G_K \backslash \G_J }=\sum_{G \in \G_K}$. 
Note however that this sum does not depend on $J$ and can therefore be cached if we need to make several runs with the same set of groups~$\G$.

 \item $t_2$ times the computation of $(S_{\varepsilon})$, that is 
 $O( t_2( s 2^{|\Theta|} + p |\G| + |\Theta|^2 + |\Theta| p  + p|\G| ) ) = O( t_2 p |\G| )$,
 with $t_1+t_2 \leq s$.
\end{itemize}

We finally get complexity with a leading term in $O( s p|\G| + s \max_{ J \in \Pattern, |J| \leq s } |\G_J|^{3.5} + s^{4.5} )$,
which is much better than $O(p^{3.5}+|\G|^{3.5})$, without an active set method.
In the example of the two-dimensional grid (see \mysec{examples}), we have $|\G|=O(\sqrt{p})$ and $O(s\max\{p^{1.75}\!,s^{3.5}\})$ as total complexity.
The simulations of \mysec{experiments} confirm that the active set strategy is indeed useful when $s$ is much smaller than $p$.
Moreover, the two extreme cases where $s\approx p$ or $p \ll 1$ are also shown not to be advantageous for the active set strategy, since 
either it is cheaper to use the SOCP solver directly on the $p$ variables, or we uselessly pay the additional fixed-cost of the active set machinery
(such as computing the optimality conditions).
Note that we have derived here the \textit{theoretical} complexity of the active set algorithm when we use a SOCP method as underlying solver. 
With the first order method presented in \myappendix{eta_trick}, we would instead get a total complexity in $O( s p^{1.5} )$.

\subsection{Intersecting Nonzero Patterns} \label{sec:nonzero_pattern_intersection}
We have seen so far how overlapping groups can encore prior information about a desired set of (non)zero patterns. In practice, controlling these overlaps may be delicate and hinges on the choice of the weights $(\dG)_{G\in\G}$ (see the experiments in \mysec{experiments}).
In particular, the weights have to take into account that some variables belonging to several overlapping groups are penalized multiple times.

However, it is possible to keep the benefit of overlapping groups whilst limiting their side effects, by taking up the idea of support intersection \citep{bolasso, meinshausen2008stability}. 
First introduced to stabilize the set of variables recovered by the Lasso, we reuse this technique in a different context, 
based on the fact that $\ZPattern$ is closed under union.

If we deal with the same sets of groups as those considered in \mysec{examples}, it is natural to rewrite $\G$ as
$\bigcup_{\theta \in \Theta}\!\G_\theta$, where $\Theta$ is the set of the orientations of the groups in $\G$ (for more details, see \mysec{examples}).
Let us denote by $\wh$ and $\wh^{\theta}$ the solutions of \myeq{minFc},
where the regularization term $\Omega$ is respectively defined by the groups in $\G$ 
and by the groups\footnote{To be more precise, in order to regularize every variable, we add the full group $\IntegerSet{p}$ to $\G_\theta$, which does not modify $\Pattern$.} in $\G_\theta$.

The main point is that, since $\Pattern$ is closed under intersection,
the two procedures described below actually lead to the same set of allowed nonzero patterns:
\begin{itemize}
 \item[a)] Simply considering the nonzero pattern of $\wh$.
 \item[b)] Taking the \textit{intersection} of the nonzero patterns obtained for each $\wh^{\theta}$, $\theta$ in $\Theta$.
\end{itemize}
With the latter procedure, although the learning of several models $\wh^\theta$ is required (a number of times equals to the number of orientations considered, e.g., 2 for the sequence, 4 for the rectangular groups and more generally $|\Theta|$ times), 
each of those learnings involves a smaller number of groups (that is, just the ones belonging to $\G_\theta$). 
In addition, this procedure is a \emph{variable selection} technique that therefore needs a second step for estimating the loadings (restricted to the selected nonzero pattern). 
In the experiments, we follow \citet{bolasso} and we use an ordinary least squares (OLS). 
The simulations of \mysec{experiments} will show the benefits of this variable selection approach.

\section{Pattern Consistency}\label{sec:consistency}

In this section, we analyze the model consistency of the solution of \myeq{minF} for the square loss. Considering the set of nonzero patterns~$\Pattern$ derived in \mysec{patterns}, we can only hope to estimate the correct hull of the generating sparsity pattern, since Theorem~\ref{thm:stability} states that other patterns occur with zero probability. We derive necessary and sufficient conditions for model consistency in a low-dimensional setting, and then consider a high-dimensional result.

We consider the square loss and a fixed-design analysis (i.e., $x_1,\dots,x_n$ are fixed).
The extension of the following consistency results to other loss functions is beyond the scope of the paper \citep[see for instance][]{bach2009self}.
We assume that for all $i \in \IntegerSet{n}$, $y_i = \wb^\top x_i + \varepsilon_i$ where the vector $\varepsilon$ is an i.i.d.~vector with Gaussian distributions with mean zero and variance $\sigma^2 >0$, and $\wb \in \R{p}$ is the population sparse vector; we denote by $\Jb$ the $\G$-adapted hull of its nonzero pattern.
Note that estimating the $\G$-adapted hull of $\wb$ is equivalent to estimating the nonzero pattern of $\wb$ if and only if this nonzero pattern belongs to $\Pattern$.
This happens when our prior information has led us to consider an appropriate set of groups $\G$.
Conversely, if $\G$ is misspecified, recovering the hull of the nonzero pattern of $\wb$ may be irrelevant,  
which is for instance the case if $\wb=\binom{\wb_1}{0} \in \R{2}$ and $\G=\{\{1\},\{1,2\}\}$.
Finding the appropriate structure of $\G$ \textit{directly from the data} would therefore be interesting future work.

\subsection{Consistency Condition}
We begin with the low-dimensional setting where $n$ is tending to infinity with $p$ \emph{fixed}. In addition, we also assume that the design is \emph{fixed} and that the Gram matrix $Q = \frac{1}{n} \sum_{i=1}^n x_i x_i^\top$ is invertible with positive-definite (i.e., invertible) limit
\[
 \lim_{n \to \infty} Q = \Qb \succ 0.
\]
In this setting, the noise is the only source of randomness.
We denote by $\mathbf{r}_\Jb$ the vector defined as
$$ 
\forall j\in\Jb,\ \mathbf{r}_j= \wb_j \bigg( \sum_{  G \in \G_{\Jb}, G \ni j } (\dG_j)^2 \NormDeux{\dG \circ \wb}^{-1}  \bigg).
$$
In the Lasso and group Lasso setting, the vector $\mathbf{r}_\Jb$ is respectively the sign vector $\textrm{sign}(\wb_\Jb)$ and the vector defined by the blocks
$(\frac{\wb_G }{\NormDeux{\wb_G}})_{G \in \G_\Jb}$.

We define
$\Omega_\Jb^c (w_{\Jb^c})
=\sum_{ G \in (\G_\Jb)^c } \NormDeux{\dG_{\Jb^c} \circ w_{\Jb^c}} $ (which is the norm composed of inactive groups) with its dual norm $(\Omega_\Jb^c)^\ast$; note the difference with the norm reduced to $\JCb$, defined as $\Omega_{\Jb^c}(w_{\Jb^c})
 = \sum_{ G \in\G  } \NormDeux{ \dG_{\Jb^c} \circ w_{\Jb^c} }$.

The following Theorem gives the sufficient and necessary conditions under which the hull of the generating pattern is consistently estimated. Those conditions naturally extend the results of \citet{Zhaoyu} and \citet{grouplasso} for the Lasso and the group Lasso respectively (see proof in Appendix~\ref{app:lowdim_patternconsistency}).
\begin{theorem}[Consistency condition]\label{thm:lowdim_patternconsistency}
Assume 
$\mu \rightarrow 0,\ \mu \sqrt{n} \rightarrow \infty$ in \myeq{minF}.
If the hull is consistently estimated, then
$
(\Omega_\Jb^c)^\ast[\Qb_{\JCb\Jb} \Qb_{\Jb\Jb}^{-1} \mathbf{r}_\Jb ]  \leq 1$.
Conversely, if
$
(\Omega_\Jb^c)^\ast[ \Qb_{\JCb\Jb} \Qb_{\Jb\Jb}^{-1} \mathbf{r}_\Jb] < 1$,
then the hull is consistently estimated, i.e.,
$$
\Prob{\{j\in\IntegerSet{p}, \hat{w}_j \neq 0 \} = \Jb} \underset{ n\rightarrow +\infty}{\longrightarrow} 1.
$$

\end{theorem}

The two previous propositions bring into play the dual norm $(\Omega_\Jb^c)^\ast$ that we cannot compute in closed form, but requires to solve an optimization problem as complex as the initial problem in \myeq{minFc}. 
However, we can prove bounds similar to those obtained in Propositions~\ref{prop:CN} and \ref{prop:CS} for the necessary and sufficient conditions.

\paragraph{Comparison with the Lasso and group Lasso.}
For the $\ell_1$-norm, our two bounds lead to the usual consistency conditions for the Lasso, i.e., the quantity $\| \Qb_{\Jb^c \Jb} \Qb_{\Jb \Jb}^{-1} {\rm sign}  (\wb_\Jb) \|_\infty$ must be less or strictly less than one. 
Similarly, when $\G$ defines a partition of $\IntegerSet{p}$ and if all the weights equal one, our two bounds lead in turn to the consistency conditions for the group Lasso, i.e., the quantity $\max_{G \in (\G_\Jb)^c} \| \Qb_{G\; \textrm{Hull}(\Jb)} \Qb_{\textrm{Hull}(\Jb) \textrm{Hull}(\Jb)}^{-1} (\frac{\wb_G }{\NormDeux{\wb_G}})_{G \in \G_\Jb} \|_2$ must be less or strictly less than one.

\subsection{High-Dimensional Analysis}

We prove a high-dimensional variable consistency result (see proof in Appendix~\ref{app:highdim}) that extends the corresponding result
for the Lasso~\citep{Zhaoyu,martin}, by assuming that the consistency condition in
Theorem~\ref{thm:lowdim_patternconsistency} is satisfied.

\begin{theorem} \label{thm:highdim_patternconsistency}
Assume that $Q$ has unit diagonal, $\kappa = \lambda_{\min}(Q_{\Jb \Jb})>0$ and
$(\Omega_\Jb^c)^\ast[ Q_{\JCb\Jb} Q_{\Jb\Jb}^{-1} \mathbf{r}_\Jb] < 1  - \tau $, with $\tau >0$. If
 $
 \tau \mu \sqrt{n} \ge \sigma C_3(\G,\Jb),
 $
and
 $
 \mu |\Jb|^{1/2} \le C_4(\G,\Jb),
 $
then the probability of incorrect hull selection is upper bounded by:
 $$
 \exp\left( \! - \frac{ n \mu^2 \tau^2 C_1(\G,\Jb)}{2 \sigma^2} \right)
 + 2 | \J | \exp \left( -\frac{ n C_2(\G,\Jb) }{2 |\Jb| \sigma^2} \right),
 $$
where $C_1(\G,\Jb)$, $C_2(\G,\Jb)$, $C_3(\G,\Jb)$ and $C_4(\G,\Jb)$ are constants defined in Appendix~\ref{app:highdim},
which essentially depend on the groups, the smallest nonzero coefficient of $\wb$
and how close the support $\{j\in\Jb: \wb_j\neq 0\}$ of $\wb$ is to its hull $\Jb$, that is
the relevance of the prior information encoded by $\G$.
\end{theorem}

In the Lasso case, we have $C_1(\G,\Jb)=O(1)$, $C_2(\G,\Jb)=O(|\Jb|^{-2})$, $C_3(\G,\Jb)=O( (\log p)^{1/2} )$ and $C_4(\G,\Jb) = O(|\Jb|^{-1})$, leading to the usual scaling $n \approx \log p$ and $\mu \approx \sigma(\log p / n)^{1/2}$.

We can also give the scaling of these constants in simple settings where groups overlap. For instance, let us consider that the variables are organized in a sequence (see \myfig{sequence}).
Let us further assume that the weights $(\dG)_{G\in\G}$ satisfy the following two properties:
\begin{itemize}
 \item[a)] The weights take into account the overlaps, that is,
  $$
  \dG_j = \beta {\textstyle (  |\{H \in \G \, ; \, H \ni j,\ H \subset G \mbox{ and } H \neq G \}|  )},
  $$
  with $t \mapsto \beta(t) \in (0,1]$ a non-increasing function such that $\beta(0)=1$,
 \item[b)] The term 
 $$
 \max_{j\in\IntegerSet{p}} \sum_{G \ni j, G\in\G} \dG_j
 $$
 is upper bounded by a constant $\mathcal{K}$ independent of $p$.
\end{itemize}
Note that we consider such weights in the experiments (see \mysec{experiments}). 
Based on these assumptions, some algebra directly leads to
$$
 \NormUn{u} \leq \Omega(u) \leq 2\mathcal{K} \NormUn{u},\, \mbox{ for all } u \in \R{p}.
$$
We thus obtain a scaling similar to the Lasso (with, \emph{in addition}, a control of the allowed nonzero patterns).

With stronger assumptions on the possible positions of $\Jb$, we may have better scalings, but these are problem-dependent (a careful analysis of the group-dependent constants would still be needed in all cases).

\section{Experiments}\label{sec:experiments}

In this section, we carry out several experiments to illustrate the behavior of the sparsity-inducing norm $\Omega$.
We denote by \textit{Structured-lasso}, or simply \textit{Slasso}, the models regularized by the norm $\Omega$. 
In addition, the procedure (introduced in \mysec{nonzero_pattern_intersection}) that consists in intersecting the nonzero patterns obtained for different models of Slasso 
will be referred to as \textit{Intersected Structured-lasso}, or simply \textit{ISlasso}.

\begin{figure}[!ht]
 
 \vspace*{-0.2cm}
 
 \begin{center}
 \begin{tabular}{cc}
 
           &  \multirow{7}{*}{\includegraphics[scale=.45]{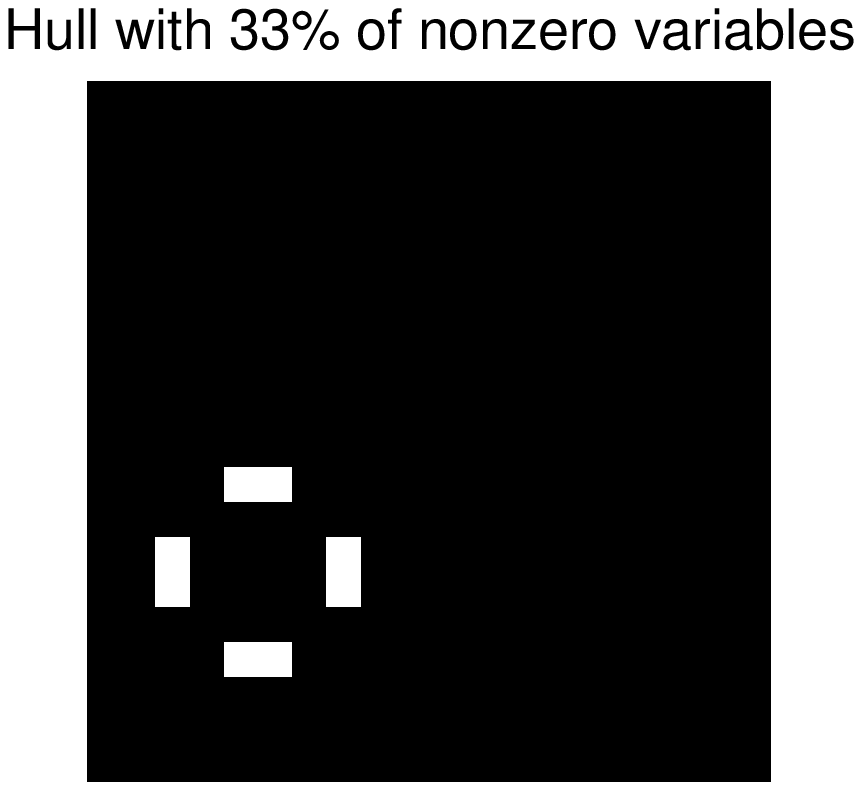}} \\
           & \\ 
           & \\ 
 	 \includegraphics[scale=.45]{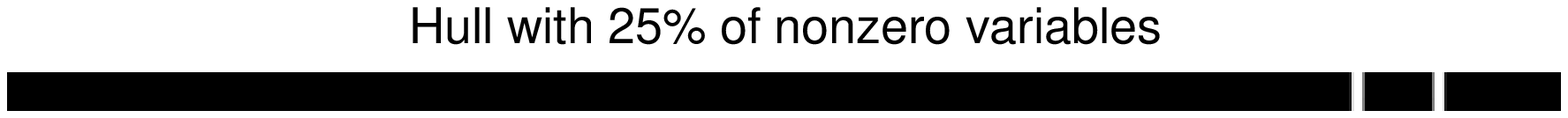}  &   \\
 	   & \\ 
 	   & \\
 	   & \\ 
 	 
 	 & \\
 	  	 
           &  \multirow{7}{*}{\includegraphics[scale=.45]{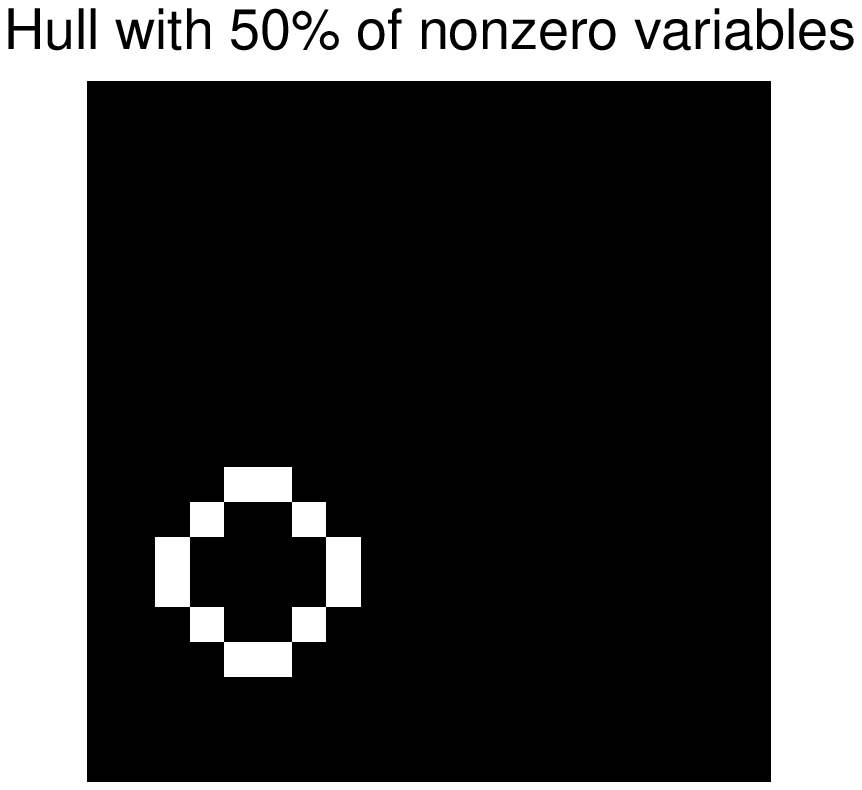}} \\
           & \\ 
           & \\ 
 	 \includegraphics[scale=.45]{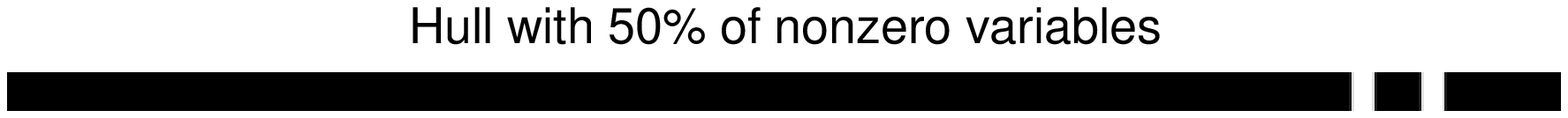}  &   \\
 	   & \\ 
 	   & \\
 	   & \\ 
 	 
 	 & \\
 	  	   
           &  \multirow{7}{*}{\includegraphics[scale=.45]{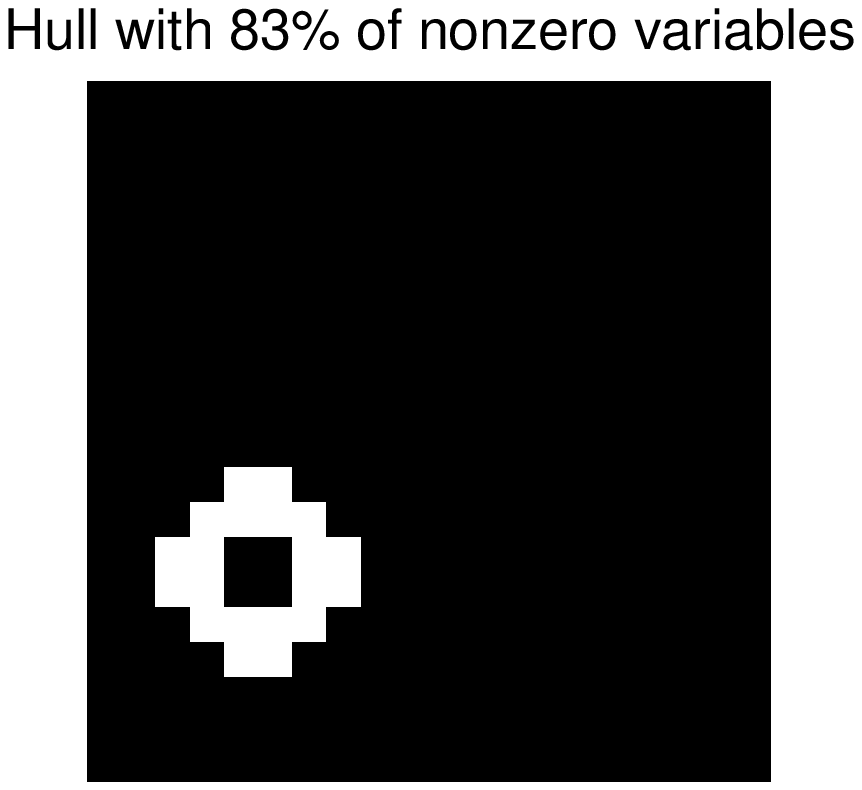}} \\
           & \\ 
           & \\ 
 	 \includegraphics[scale=.45]{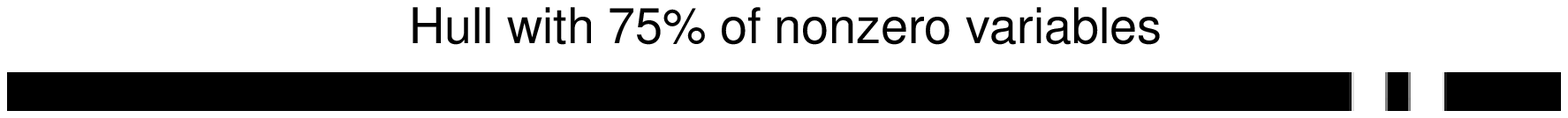}  &   \\
 	   & \\ 
 	   & \\
 	   & \\ 
 	 
 	 & \\
 	 
           &  \multirow{7}{*}{\includegraphics[scale=.45]{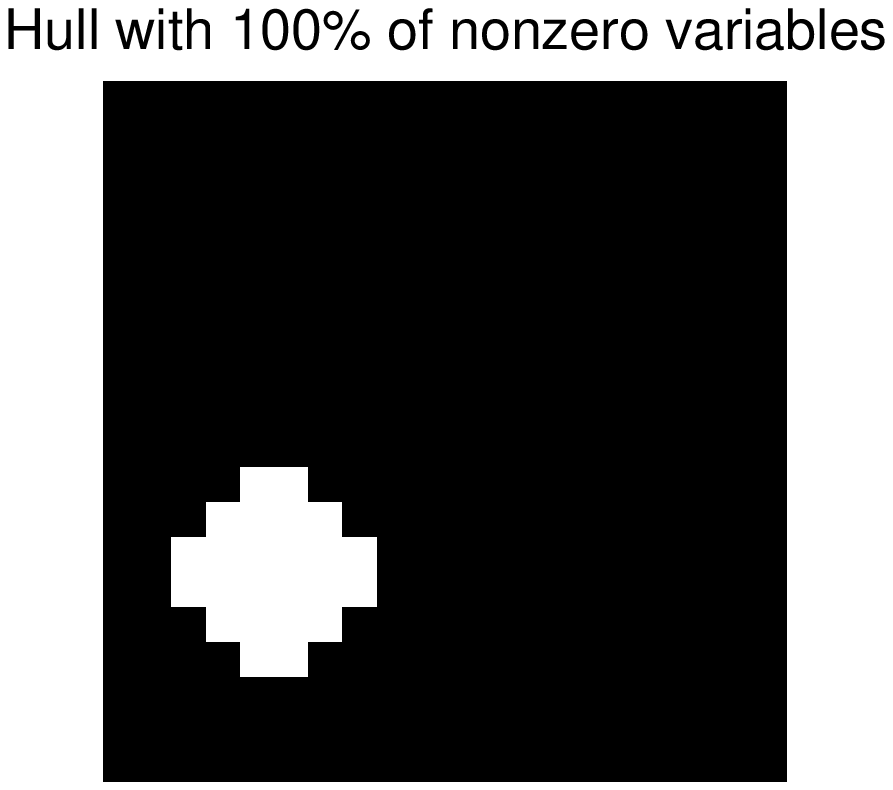}} \\
           & \\ 
           & \\ 
 	 \includegraphics[scale=.45]{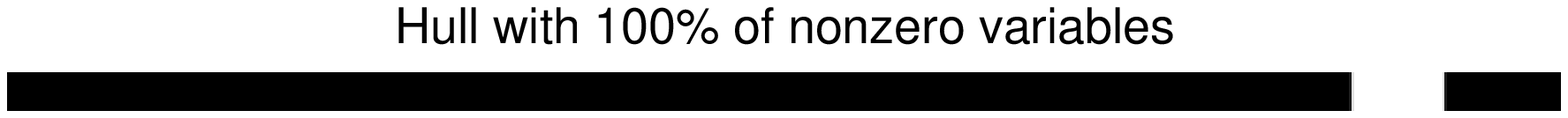}  &   \\
 	   & \\ 
 	   & \\
 	   & \\

 \end{tabular}
 \end{center}

\vspace*{-0.75cm}

 \caption{Examples of generating patterns (the zero variables are represented in black, while the nonzero ones are in white): 
 (Left column, in white) generating patterns that are used for the experiments on $400$-dimensional sequences; those patterns all form the same hull of 24 variables 
 (i.e., the contiguous sequence in the bottom left figure).
 (Right column, in white) generating patterns that we use for the $20\!\times\!20$-dimensional grid experiments; again, those patterns all form the same hull of 24 variables
 (i.e., the diamond-shaped convex in the bottom right figure).
 The positions of these generating patterns are randomly selected during the experiments.
 For the grid setting, the hull is defined based on the set of groups that are half-planes, with orientations that are multiple of $\pi / 4$
 (see \mysec{examples}).}
 \label{fig:generating_patterns}
\end{figure}

Throughout the experiments, we consider noisy linear models 
$Y=\wb^{\top}\!X+\varepsilon$,
where 
$\wb \in \R{p}$ is the generating loading vector and 
$\varepsilon$ is a centered Gaussian noise with its variance set to satisfy 
$\NormDeux{\wb^{\top}\!X}=3\NormDeux{\varepsilon}$. This consequently leads to a fixed signal-to-noise ratio.
We assume that the vector $\wb$ is sparse, i.e., it has only a few nonzero components, that is, $|\Jb| \ll p$. 
We further assume that these nonzero components are either organized on a sequence or on a two-dimensional grid (see \myfig{generating_patterns}).
Moreover, we consider sets of groups $\G$ such as those presented in \mysec{examples}. 
We also consider different choices for the weights $(\dG)_{G\in\G}$ that we denote by \textbf{(W1)}, \textbf{(W2)} and \textbf{(W3)} 
(we will keep this notation in the following experiments):
\begin{enumerate}
 \item[\textbf{(W1)}:] uniform weights, $\dG_j = 1,$
 \item[\textbf{(W2)}:] weights depending on the size of the groups, $\dG_j = |G|^{-2},$
 \item[\textbf{(W3)}:] weights that take into account overlapping groups, 
  $
  \dG_j = \rho^{\, |\{H \in \G \, ; \, H \ni j,\ H \subset G \mbox{ and } H \neq G \}|},
  $ for some $\rho \in (0,1)$.
\end{enumerate}
For each orientation in $\G$, the third type of weights \textbf{(W3)} aims at reducing the unbalance caused by the overlapping groups.
Specifically, given a group $G\in\G$ and a variable $j\in G$, the corresponding weight $\dG_j$ is all the more small as the variable $j$ 
already belongs to other groups with the same orientation. 

Unless otherwise specified, we use the third type of weights \textbf{(W3)} with $\rho=0.5$.   
In the following experiments, the loadings $w_\Jb$, as well as the design matrices, are generated from a standard Gaussian distribution with identity covariance matrix.
The positions of $\Jb$ are also random and are uniformly drawn.


\paragraph{Prediction error and relevance of the structured prior.} 
We show in this experiment that the prior information we put through the norm $\Omega$ improves upon the predictive power. 
We are looking at two situations where we can express a structural prior through~$\Omega$, 
namely 
(1) the selection of a contiguous pattern on a sequence and 
(2) the selection of a convex pattern on a grid (see \myfig{generating_patterns}).

\begin{figure}[!ht]
\begin{center}

	\begin{tabular}{cc}
	\includegraphics[scale=.41]{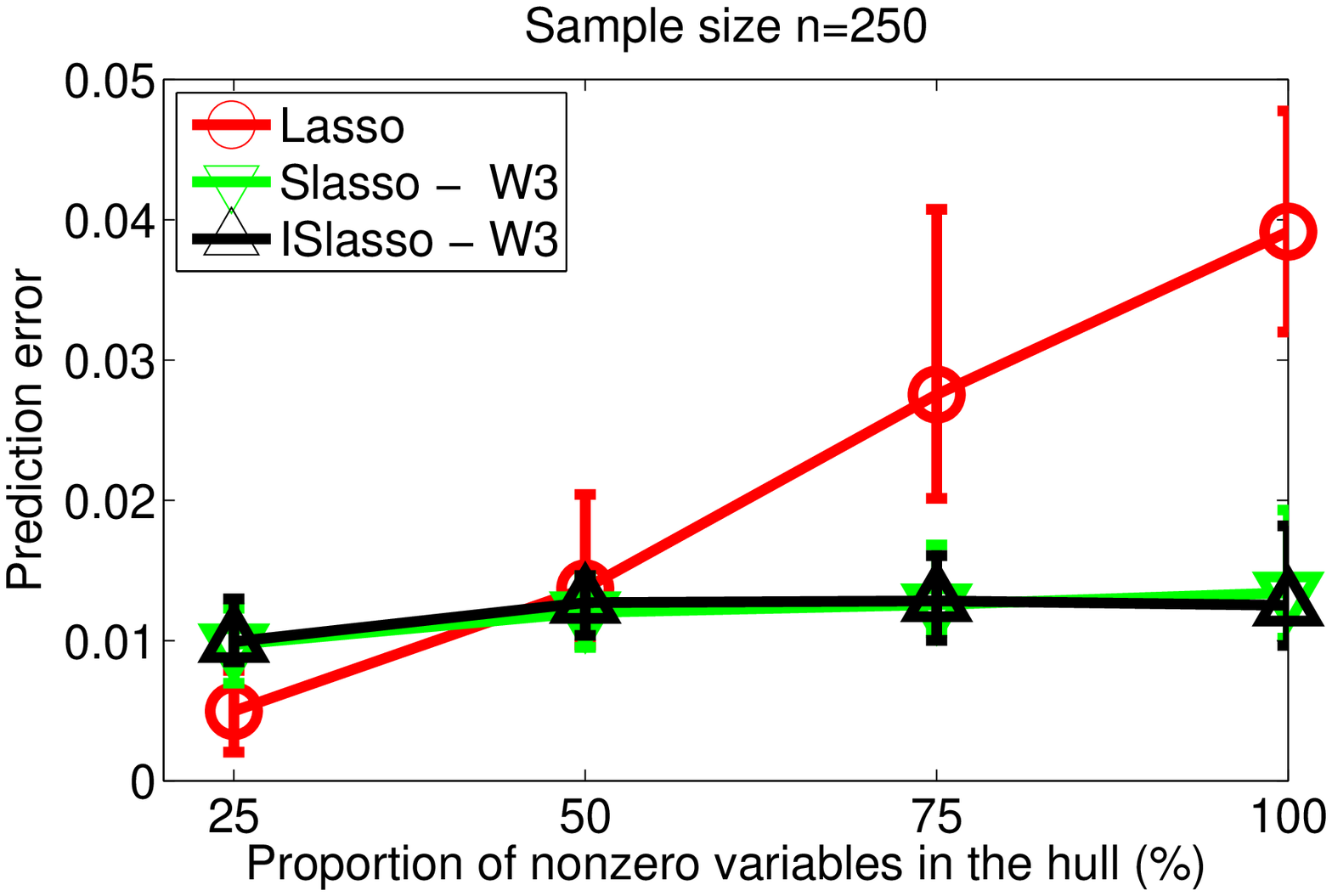}  & \includegraphics[scale=.41]{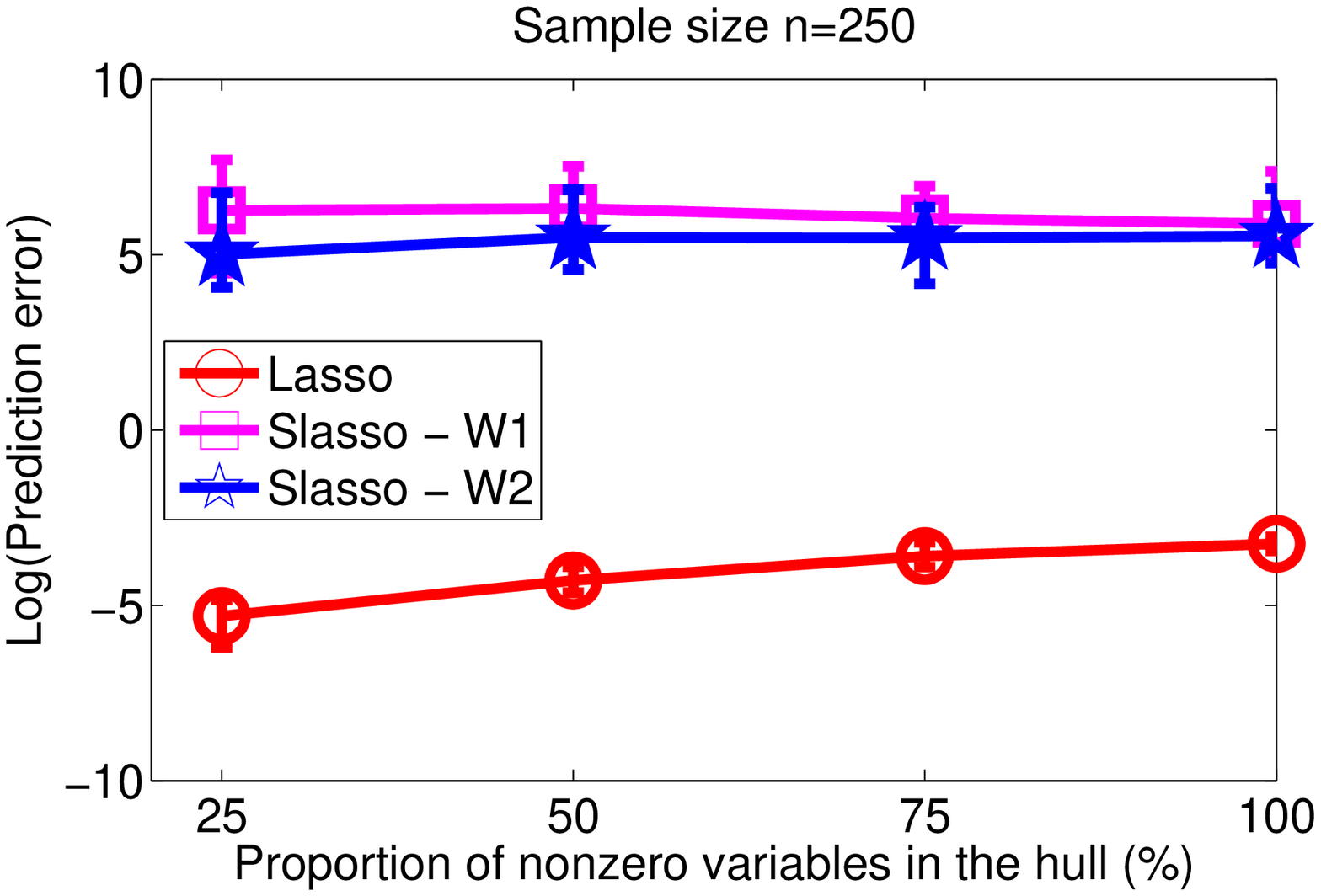} \\
	& \\
	\includegraphics[scale=.41]{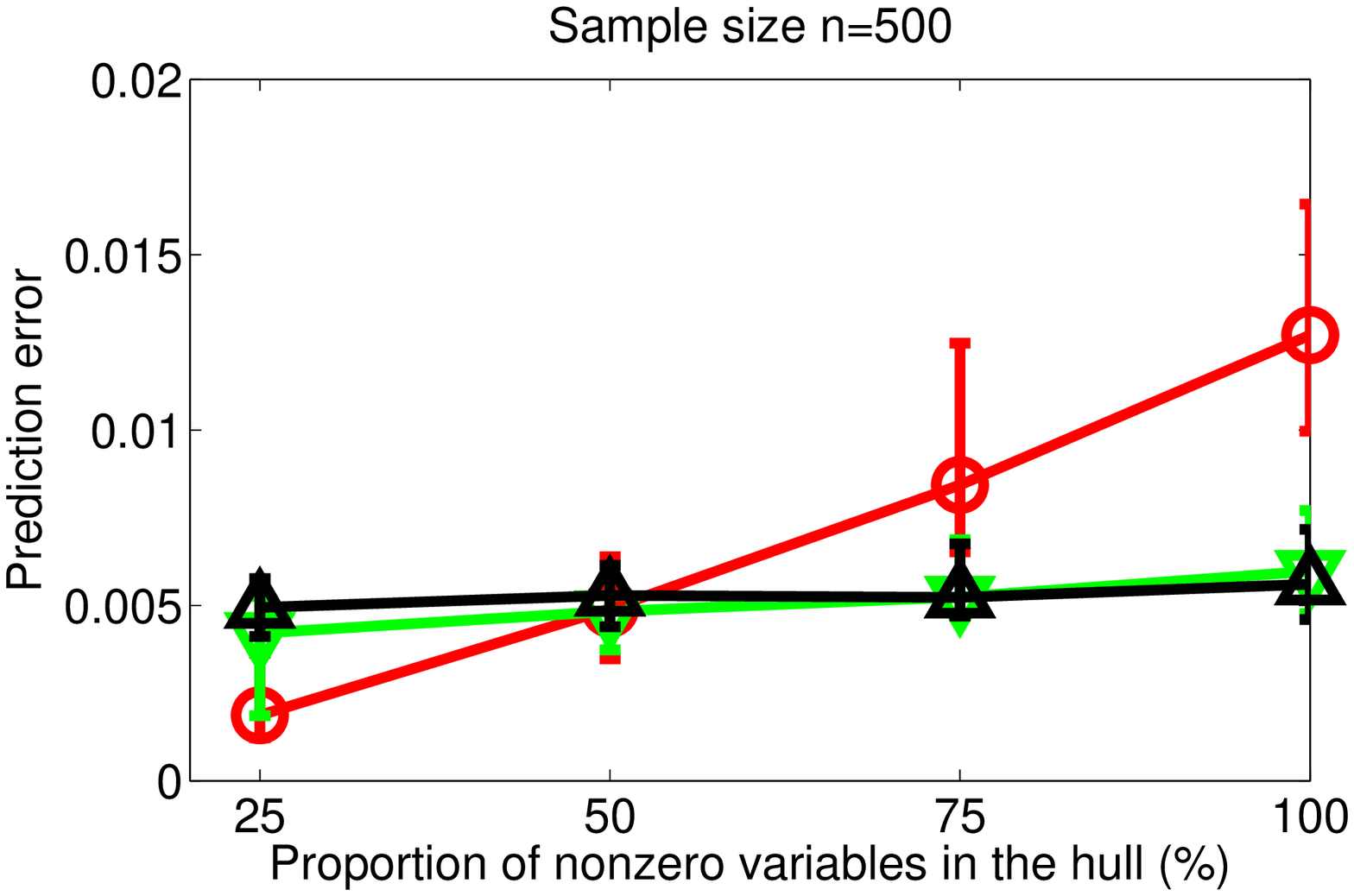}  & \includegraphics[scale=.41]{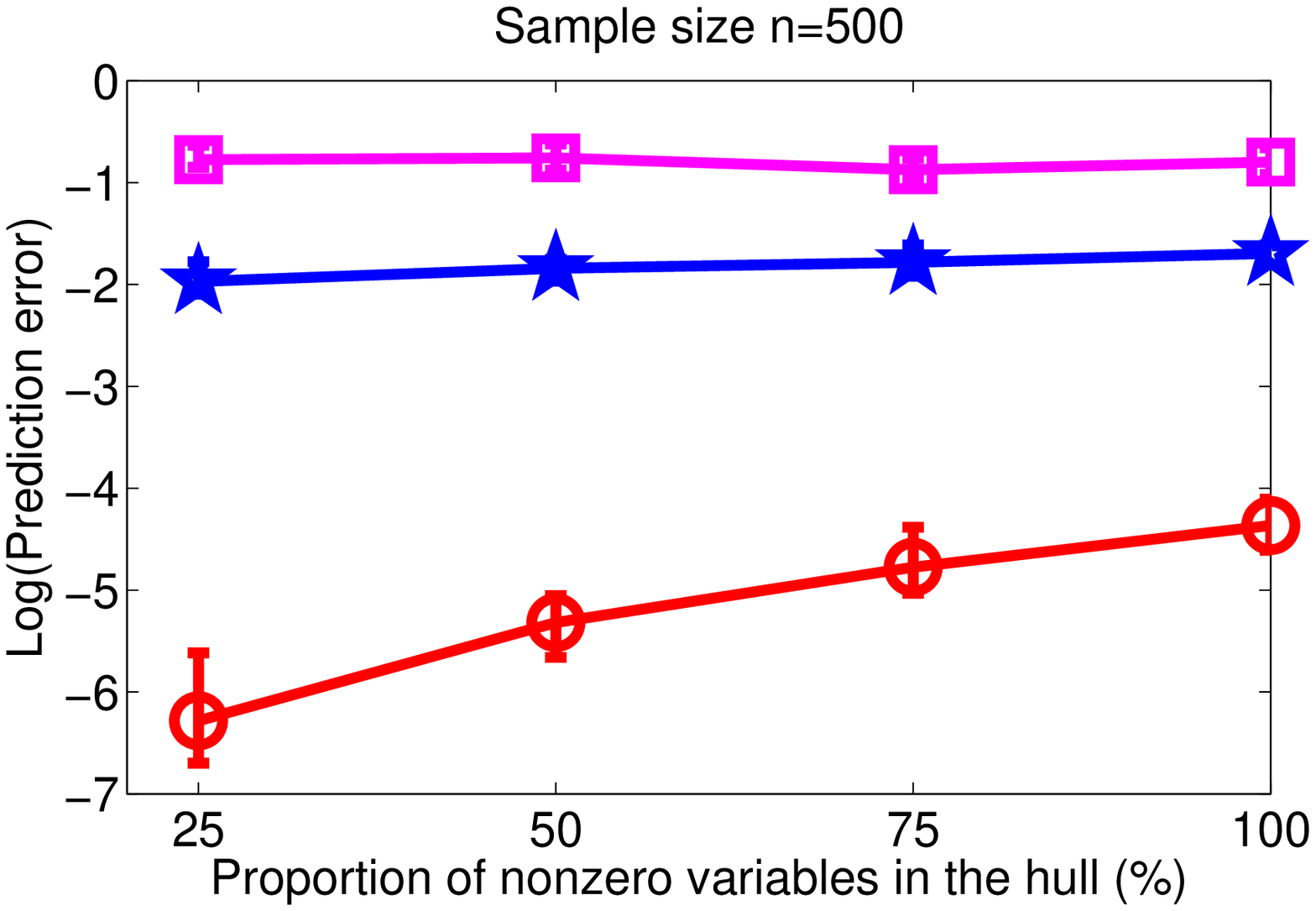} \\
	& \\
	\includegraphics[scale=.41]{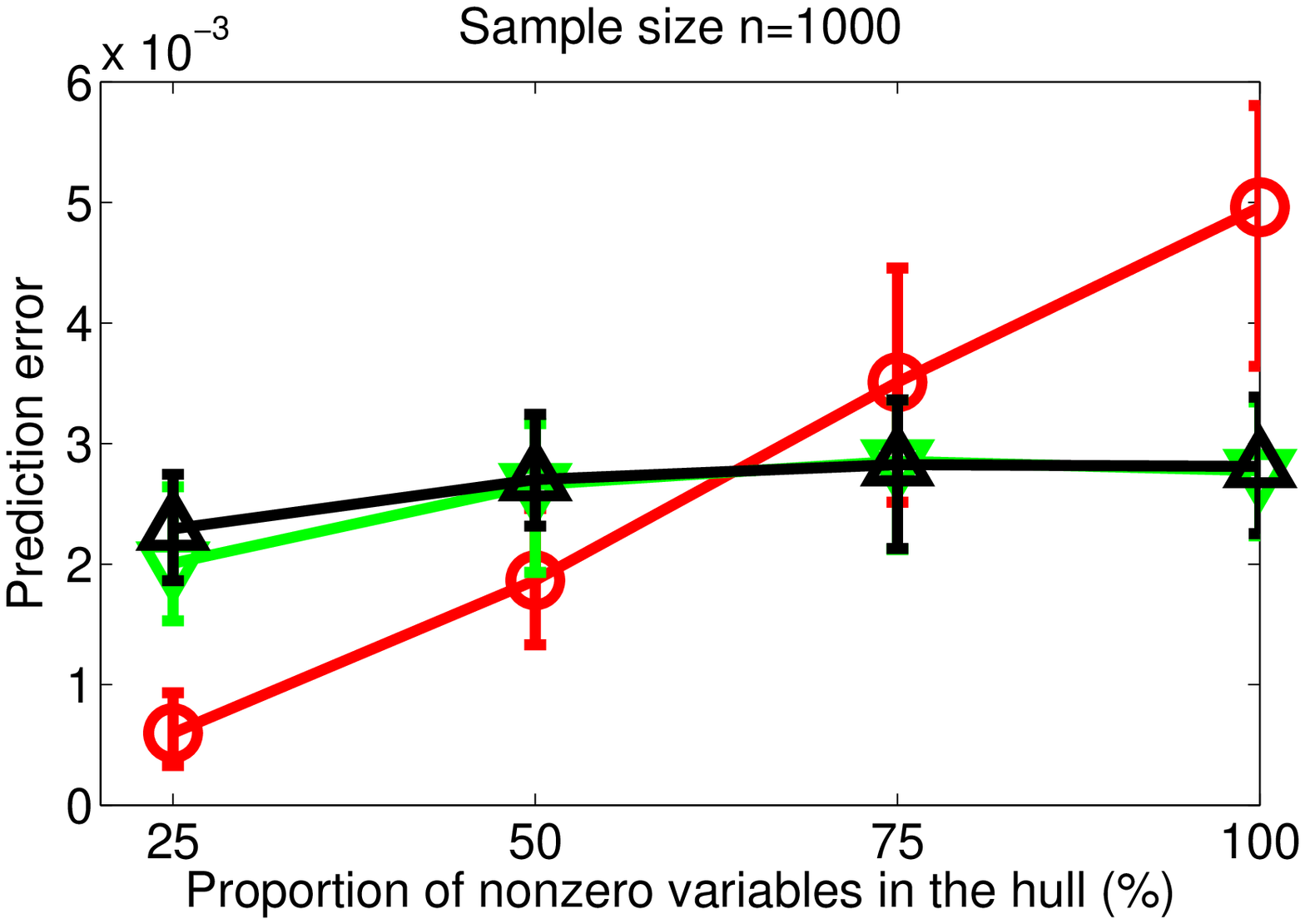} & \includegraphics[scale=.41]{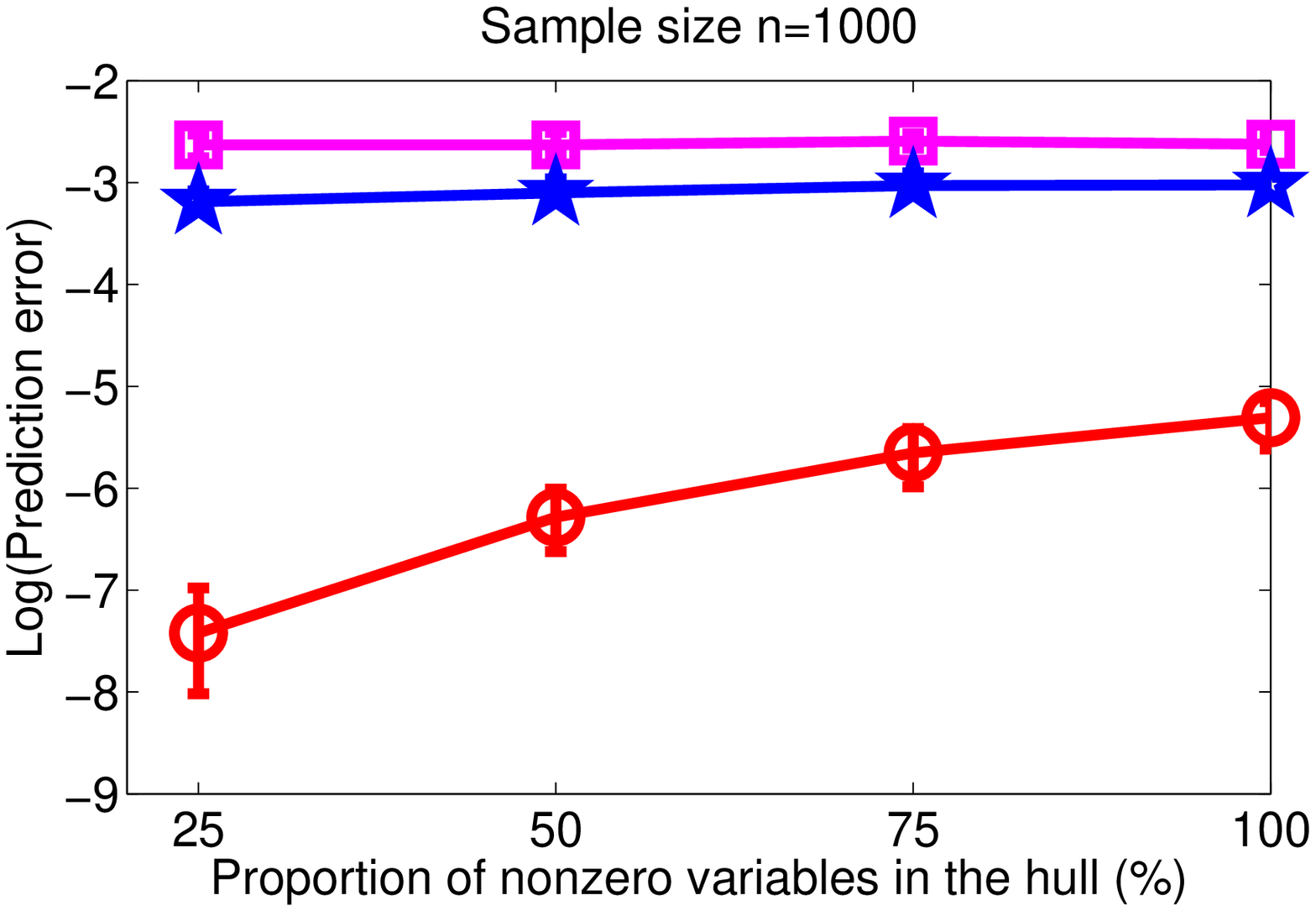}
	\end{tabular}

\end{center}

\DistanceToCaption

\caption{Experiments on the sequence: for the sample sizes $n \in \{250, 500, 1000\}$,
 we plot the prediction error versus the proportion of nonzero variables in the hull of the generating pattern.
 We display the results on two different columns since the models obtain very heterogeneous performances (on the right column, the error is in $\log$ scale).
 The points, the lower and upper error bars on the curves respectively represent the median, the first and third quartile, based on 50 random settings $\{\Jb,\wb,X,\varepsilon\}$.
}
\label{fig:sequence_prediction_error}
\end{figure}

\begin{figure}[!ht]
\begin{center}

\vspace*{-0.5cm}

	\begin{tabular}{cc}
	\includegraphics[scale=.41]{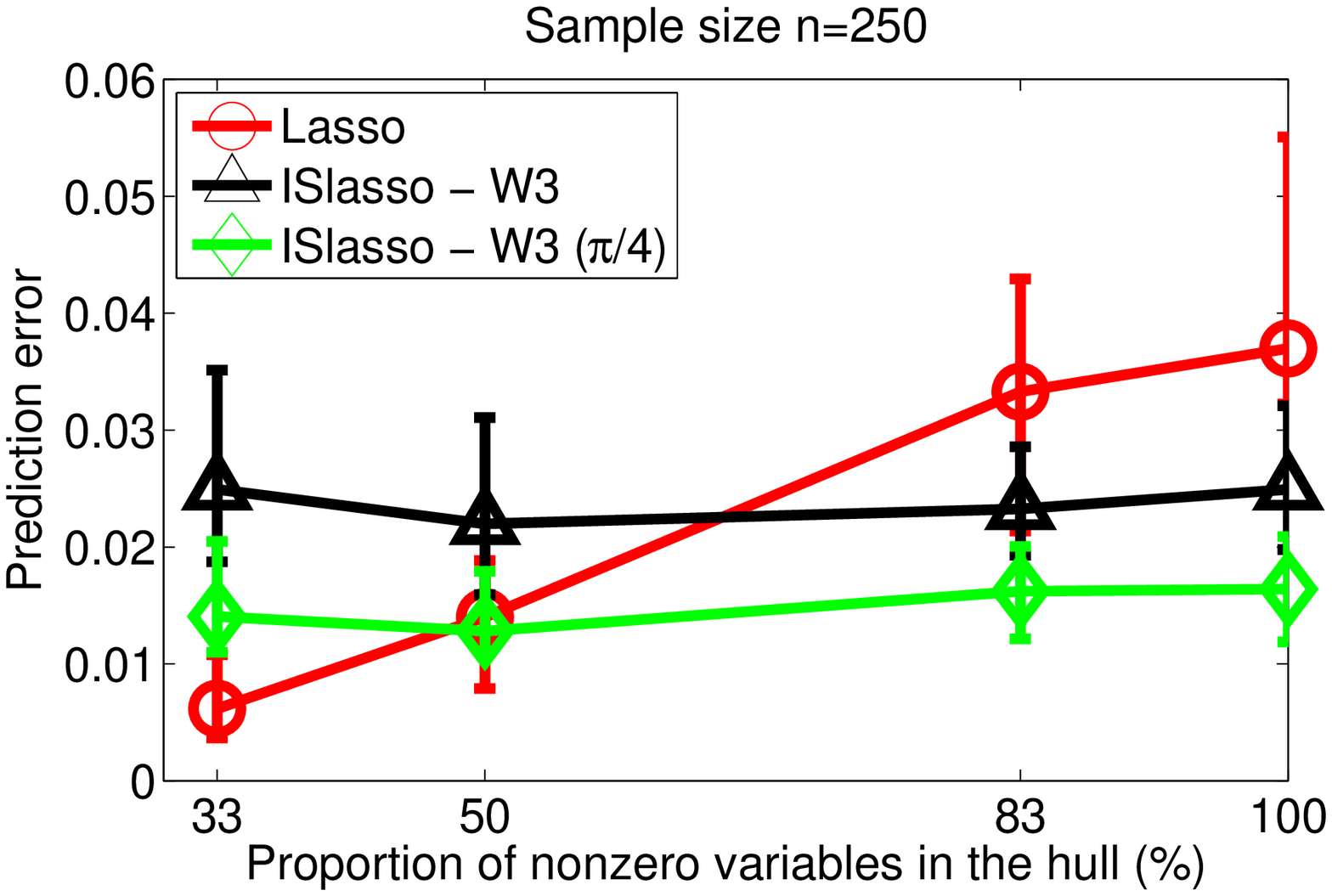} & \includegraphics[scale=.41]{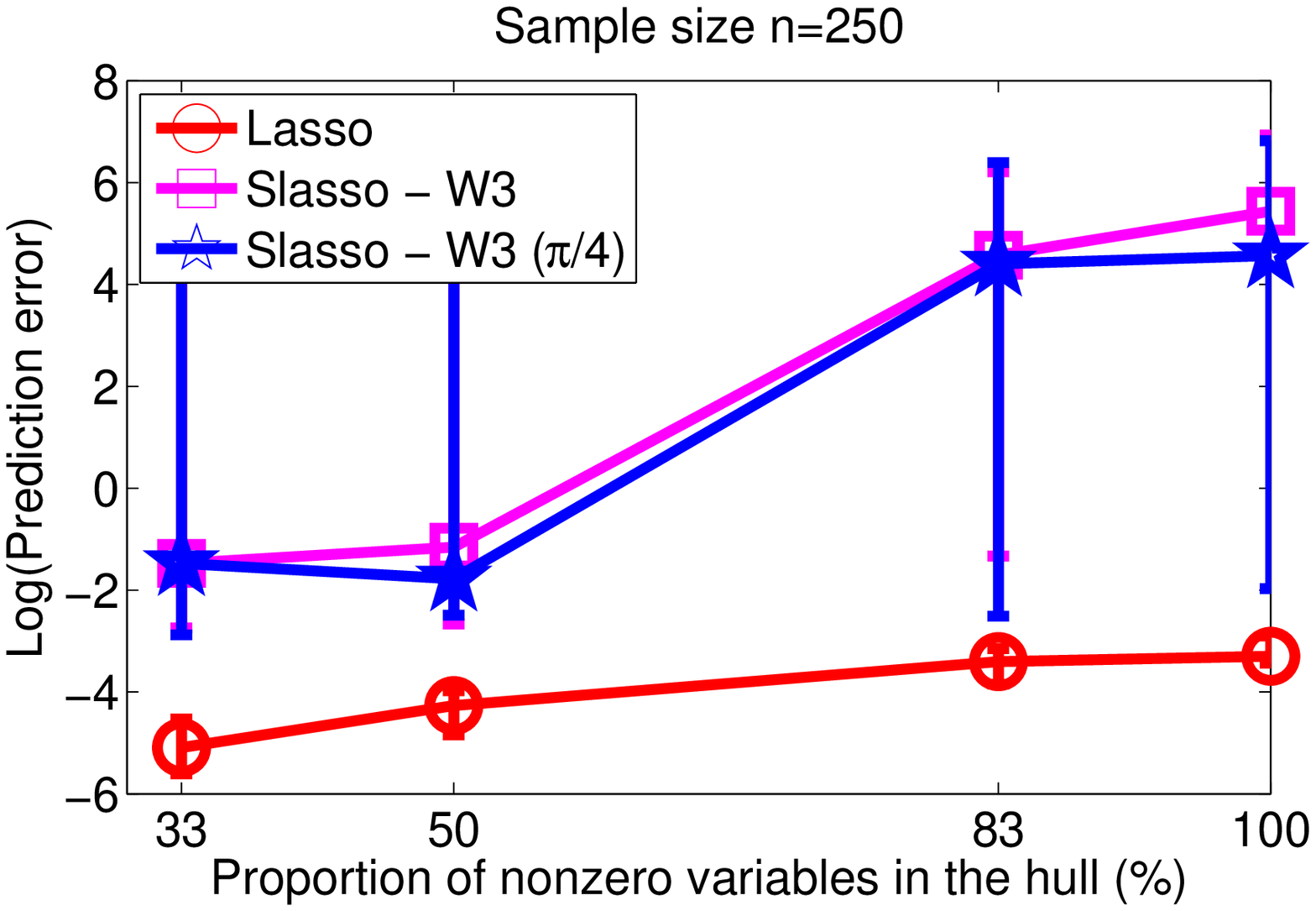} \\
	& \\
	\includegraphics[scale=.41]{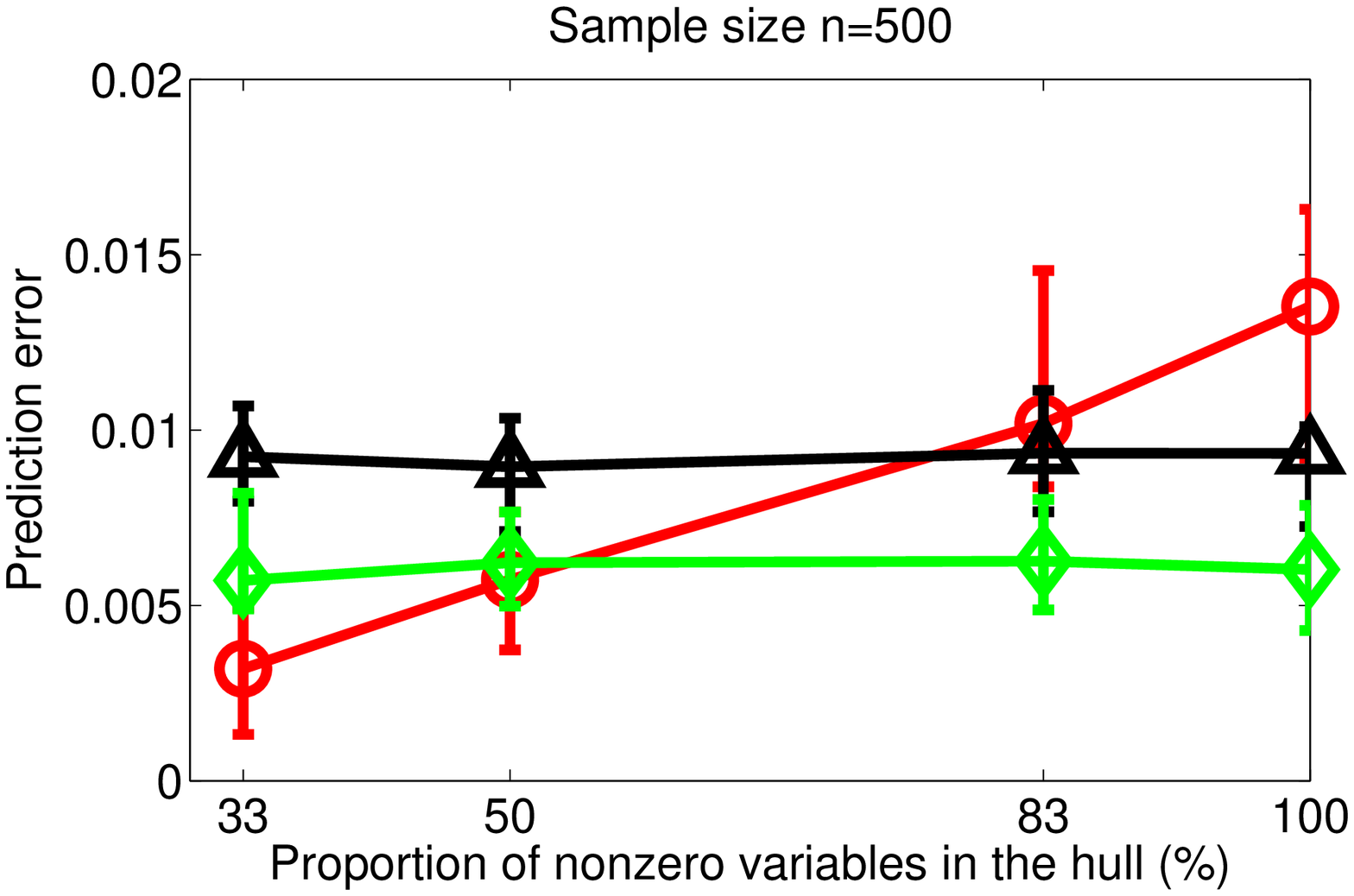} & \includegraphics[scale=.41]{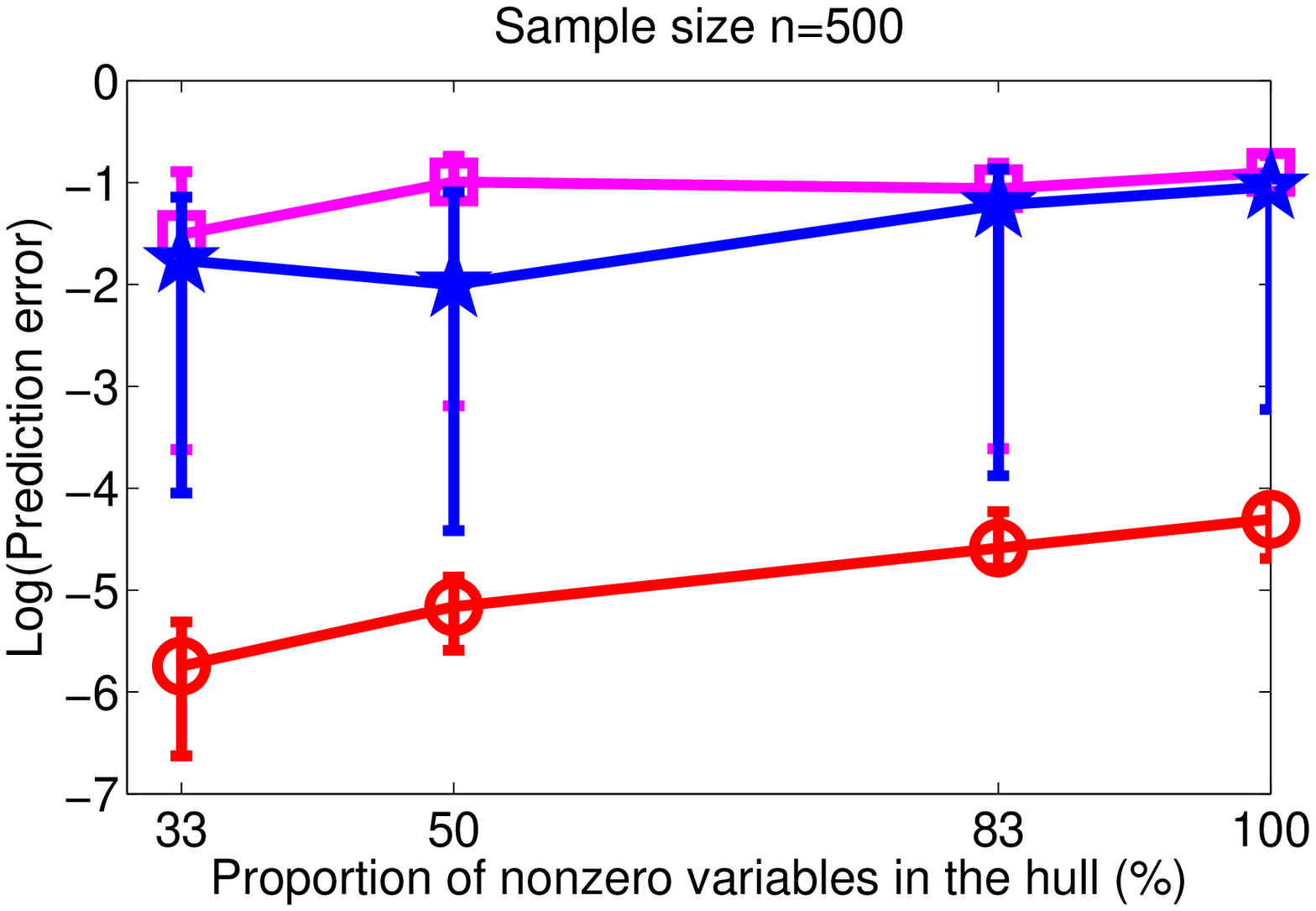} \\
	& \\
	\includegraphics[scale=.41]{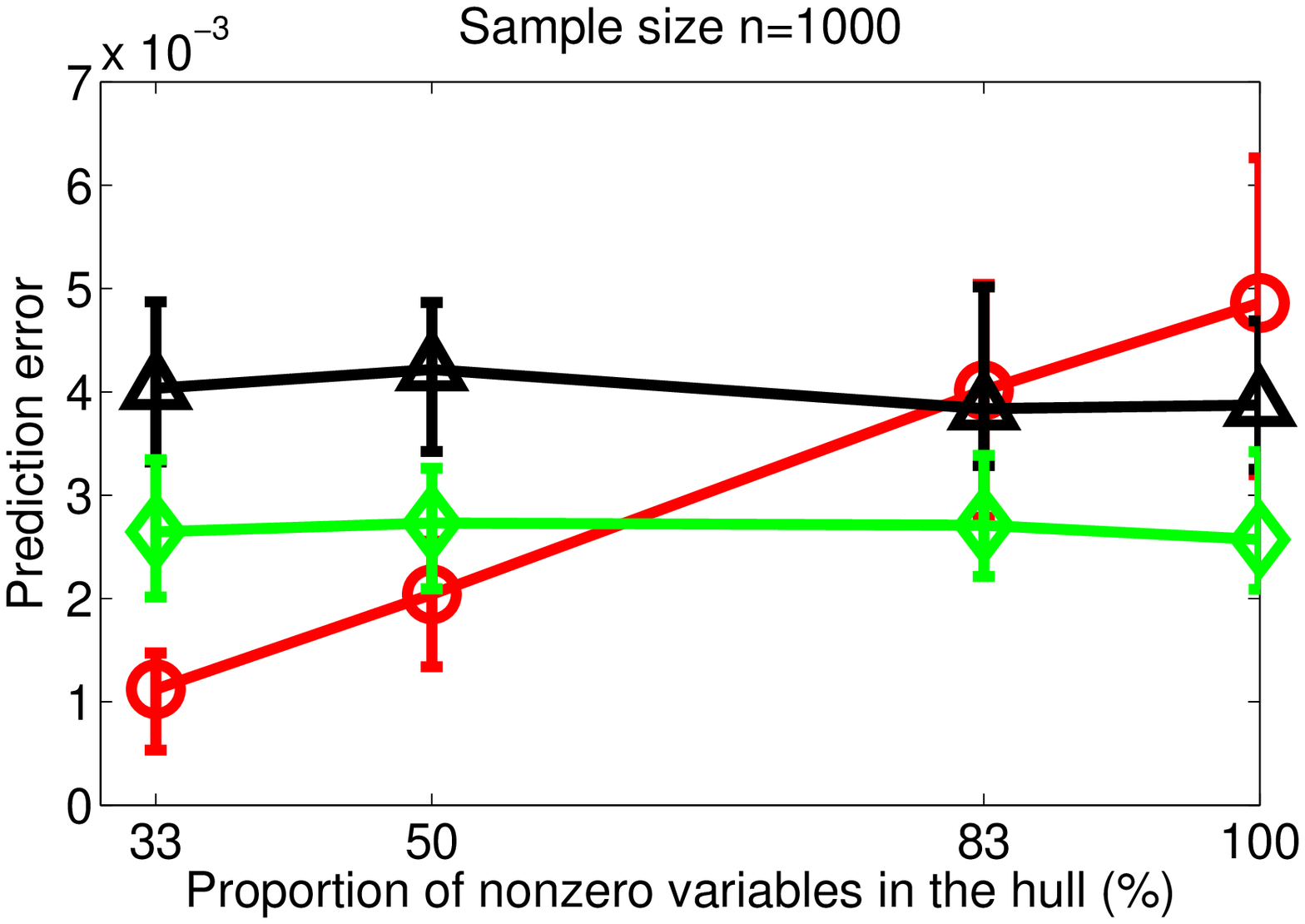} & \includegraphics[scale=.41]{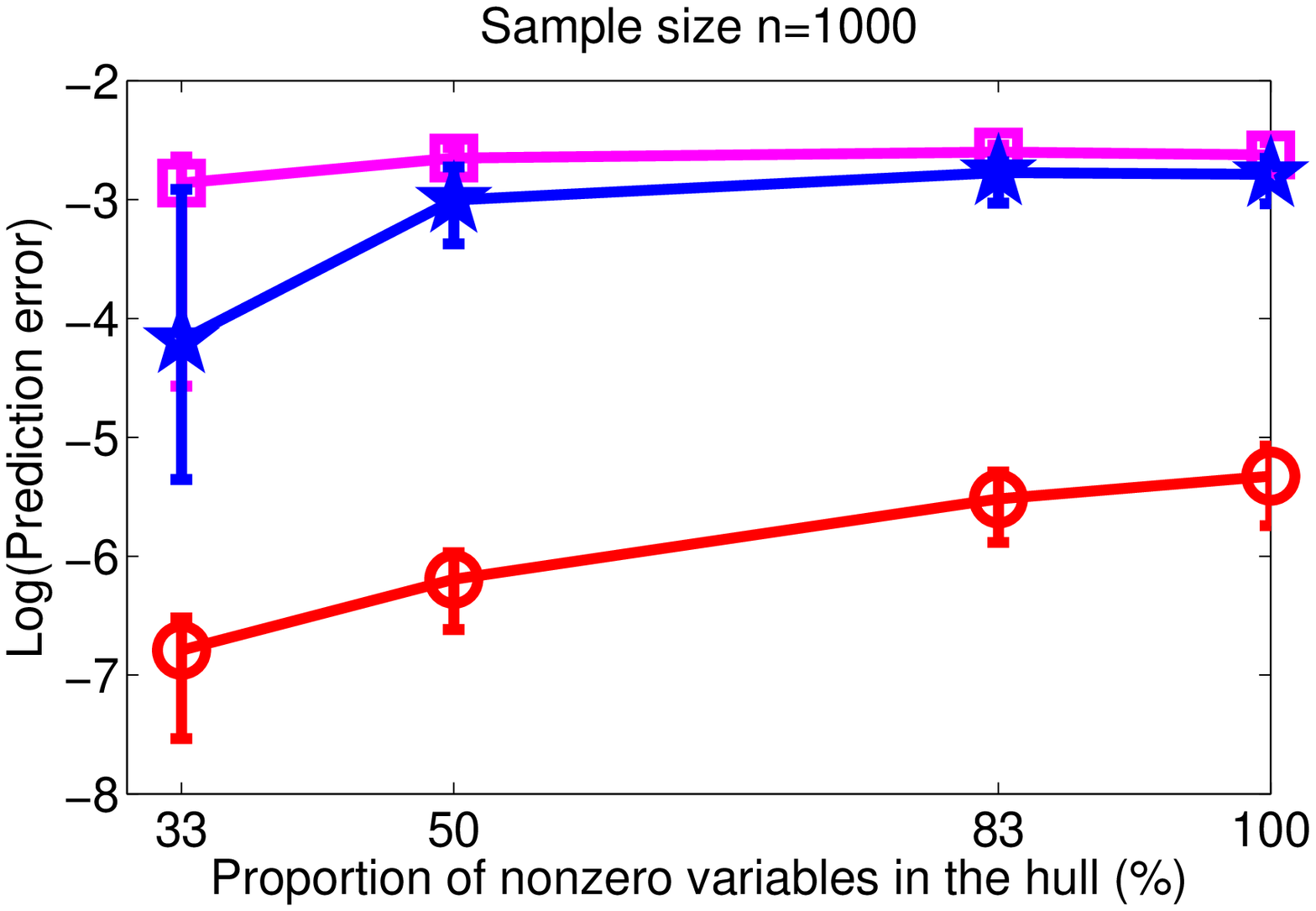}
	\end{tabular}

\end{center}

\DistanceToCaption

\caption{Experiments on the grid: for the sample sizes $n \in \{250, 500, 1000\}$,
 we plot the prediction error versus the proportion of nonzero variables in the hull of the generating pattern.
 We display the results on two different columns since the models obtain very heterogeneous performances (on the right column, the error is in $\log$ scale).
 The points, the lower and upper error bars on the curves respectively represent the median, the first and third quartile, based on 50 random settings $\{\Jb,\wb,X,\varepsilon\}$.
 Two sets of groups $\G$ are considered, the rectangular groups with or without the $\pm\pi/4$-groups (denoted by $(\pi/4)$ in the legend).
 In addition, we have dropped for clarity the models that performed already poorly on the sequence.}
\label{fig:grid_prediction_error}
\end{figure}

In what follows, we consider $p=400$ variables with generating patterns $\wb$ whose hulls have a constant size of $|\Jb|=24$ variables.
In order to evaluate the relevance of the contiguous (or convex) prior, we also vary the number of zero variables that are contained in the hull (see \myfig{generating_patterns}).
We then compute the prediction error for different sample sizes $n \in \{250,500,1000\}$.
The prediction error is understood here as 
$$
\frac{ \NormDeux{X^{\textrm{test}}(\wb - \wh)}^2 }{ \NormDeux{X^{\textrm{test}}\wb }^2 },
$$
where $\wh$ denotes the estimate of the OLS, performed on the nonzero pattern found by the model considered
(i.e., either Lasso, Slasso or ISlasso)\footnote{Even though we make comparisons based on prediction errors, the experiments illustrate the results from \mysec{consistency} since we first use our method as a variable selection step.}.
The regularization parameter is chosen by 5-fold cross-validation and the test set consists of 500 samples.
For each value of $n$, we display on \myfig{sequence_prediction_error} and \myfig{grid_prediction_error} the median errors over 50 random settings $\{\Jb,\wb,X,\varepsilon\}$, for respectively the sequence and the grid.

First and foremost, the simulations highlight how important the weights $(\dG)_{G\in\G}$ are. 
In particular, the uniform \textbf{(W1)} and size-dependent weights \textbf{(W2)} perform poorly since they do not take into account the overlapping groups. 
The models learned with such weights do not manage to recover the correct nonzero patterns 
(and even worse, they tend to select every variable---see the right column of \myfig{sequence_prediction_error}).

Although groups that moderately overlap do help (e.g., see the Slasso with the weights \textbf{(W3)} on the left column of \myfig{sequence_prediction_error}), 
it remains delicate to handle groups with many overlaps, even with an appropriate choice of $(\dG)_{G\in\G}$ 
(e.g., see the right column of \myfig{grid_prediction_error} where Slasso considers up to 8 overlaps on the grid). 
The ISlasso procedure does not suffer from this issue since it reduces the number of overlaps whilst keeping the desirable effects of overlapping groups.
Another way to yield a better level of sparsity, even with many overlaps, would be to consider non-convex alternatives to $\Omega$ \citep[see, e.g.,][]{SparseStructuredPCA}. 
Naturally, the benefit of ISlasso is more significant on the grid than on the sequence as the latter deals with fewer overlaps.
Moreover, adding the $\pm\pi/4$-groups to the rectangular groups enables to recover a nonzero pattern closer to the generating pattern. This is illustrated on the left column of 
\myfig{grid_prediction_error} where the error of ISlasso with only rectangular groups (in black) corresponds to the selection of the smallest rectangular box around the generating pattern.

On the other hand, and more importantly, the experiments show that if the prior about the generating pattern is relevant, then our structured approach performs better that the standard Lasso.
Indeed, as displayed on the left columns of \myfig{sequence_prediction_error} and \myfig{grid_prediction_error}, 
as soon as the hull of the generating pattern does not contain too many zero variables, Slasso/ISlasso outperform Lasso.
In fact, the sample complexity of the Lasso depends on the number of nonzero variables in $\wb$ as opposed to the size of the hull for Slasso/ISlasso. 
This also explains why the error for Slasso/ISlasso is almost constant with respect to the number of nonzero variables (since the hull has a constant size).
Note finally that, even though our structured approach does not always dramatically outperform the standard unstructured approach in terms of prediction, it has the advantage of being more interpretable. 

\begin{figure}[!ht]


\begin{center}
\includegraphics[scale=.7]{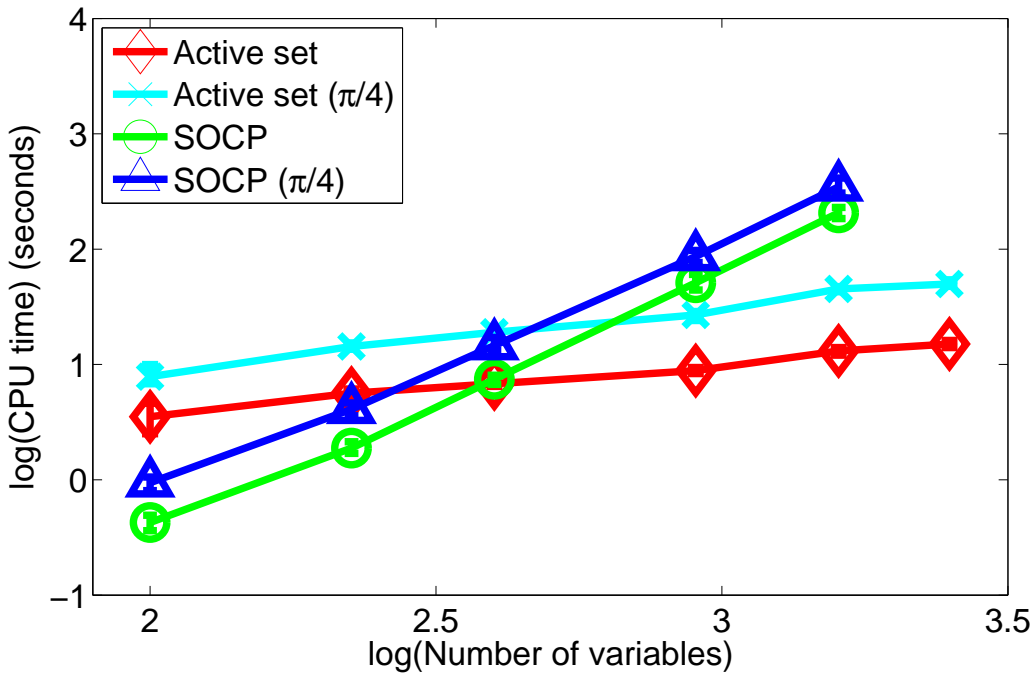}
\end{center}

\DistanceToCaption

\caption{Computational benefit of the active set algorithm: CPU time (in seconds) versus the number of variables $p$, displayed in $\log$-$\log$ scale.
The points, the lower and upper error bars on the curves respectively represent the median, the first and third quartile.
Two sets of groups $\G$ are considered, the rectangular groups with or without the $\pm\pi/4$-groups (denoted by $(\pi/4)$ in the legend).
Due to the computational burden, we could not obtain the SOCP's results for $p=2500$.}
\label{fig:cputime_vs_p}
\end{figure}
\vspace*{1cm}
\vspace*{1cm}

\paragraph{Active set algorithm.} 
We finally focus on the active set algorithm (see \mysec{optimization}) and compare its time complexity to the SOCP solver when we are looking for a sparse structured target.
More precisely, for a fixed level of sparsity $|\J|=24$ and a fixed number of observations $n=3500$, 
we analyze the complexity with respect to~the number of variables $p$ that varies in $\{100,225,400,900,1600,2500\}$.

We consider the same experimental protocol as above except that we display the median CPU time based only\footnote{Note that it already corresponds to several hundreds of runs for both the SOCP and the active set algorithms since we compute a 5-fold cross-validation for each regularization parameter of the (approximate) regularization path.} on 5 random settings $\{\Jb,\wb,X,\varepsilon\}$. 

We assume that we have a rough idea of the level of sparsity of the true vector and we set the stopping criterion $s=4|\Jb|$ (see Algorithm~\ref{alg:activeset}), 
which is a rather conservative choice. 
We show on \myfig{cputime_vs_p} that we considerably lower the computational cost for the same level of performance\footnote{We have not displayed this second figure that is just the superposition of the error curves for the SOCP and the active set algorithms.}.
As predicted by the complexity analysis of the active set algorithm (see the end of \mysec{optimization}), considering the set of rectangular groups with or without the $\pm\pi/4$-groups 
results in the same complexity (up to constant terms). 
We empirically obtain an average complexity of $\, \approx O(p^{2.13})$ for the SOCP solver and of $\, \approx O(p^{0.45})$ for the active set algorithm.

Not surprisingly, for small values of $p$, the SOCP solver is faster than the active set algorithm, 
since the latter has to check its optimality by computing necessary and sufficient conditions (see Algorithm~\ref{alg:activeset} and the discussion in the algorithmic complexity paragraph of \mysec{optimization}). 



\section{Conclusion}

We have shown how to incorporate prior knowledge on the form
of nonzero patterns for linear supervised learning.
Our solution relies on
a regularizing term which linearly combines $\ell_2$-norms of possibly
overlapping groups of variables.
Our framework brings into play intersection-closed families of nonzero patterns, such as all rectangles on a two-dimensional grid. 
We have studied the design of these groups,
efficient algorithms and theoretical guarantees of the structured sparsity-inducing method.
Our experiments have shown to which extent our model leads to better prediction, depending on the relevance of the prior information.

A natural extension to this work is to consider bootstrapping since this may improve
theoretical guarantees and result in better variable selection~\citep{bolasso, meinshausen2008stability}.
In order to deal with broader families of (non)zero patterns, it would be interesting to
combine our approach with recent work on structured sparsity: for instance, \citet{baraniuk2008model, LaurentGuillaumeGroupLasso} consider union-closed collections of nonzero patterns,
\citet{CaringWaveletBasedBayesian} exploit structure through a Bayesian prior
while
\citet{huang2009learning} handle non-convex penalties based on information-theoretic criteria.

More generally, 
our regularization scheme could also be used for various learning tasks,
as soon as prior knowledge on the structure of the sparse representation is available, e.g., for
multiple kernel learning~\citep{MicchelliP05}, 
multi-task learning~\citep{ArgyriouEP08, obozinski-joint, kim2009tree} 
and sparse matrix factorization problems~\citep{JulienJMLR, SparseStructuredPCA}.

Finally, although we have mostly explored in this paper the algorithmic
and theoretical issues related to these norms, this
type of prior knowledge is of clear interest for the spatially and temporally structured data typical in bioinformatics, computer vision and
neuroscience applications~\citep[see, e.g.,][]{SparseStructuredPCA}.

\section*{Acknowledgments}
We would like to thank the anonymous reviewers for their constructive comments
that improve the clarity and the overall quality of the manuscript. 
We also thank Julien Mairal and Guillaume Obozinski for insightful discussions.
This work was supported in part by a grant from the
Agence Nationale de la Recherche (MGA Project) and
a grant from the European Research Council (SIERRA Project).

\appendix

\section{Proof of Proposition \ref{thm:uniqueness} }\label{app:uniqueness_theorem}

We recall that $L(w) = \ERisk{y_i}{w ^\top x_i}$.
Since $w\mapsto \Omega(w)$ is convex and goes to infinite when $\NormDeux{w}$ goes to infinite,
and since $L$ is lower bounded, by Weierstrass' theorem, the problem in \myeq{minF} admits at least one global solution.\\
\textbullet\textit{First case: $Q$ invertible.}
The Hessian of $L$ is 
$$
\frac{1}{n} \sum_{i=1}^n x_i x_i^\top \frac{\partial^2 \ell}{\partial {y'}^2}(y_i,w ^\top x_i).
$$
It is positive definite since $Q$ is positive definite and $\min_{i\in\{1,\dots,n\}} \frac{\partial^2 \ell}{\partial {y'}^2}(y_i,w ^\top x_i)>0$.
So $L$ is strictly convex. Consequently the objective function $L+\mu\Omega$ is strictly convex, hence the uniqueness of its minimizer.\\
\textbullet\textit{Second case: $\{1,\dots,p\}$ belongs to $\G$.}
We prove the uniqueness by contradiction. 
Assume that the problem in \myeq{minF} admits two different solutions $w$ and $\tilde{w}$. 
Then one of the two solutions is different from $0$, say $w \neq 0$. 

By convexity, it means that any point of the segment 
$[w,\tilde{w}]=\big\{a w+(1-a)\tilde{w};\ a\in[0,1]\big\}$ 
also minimizes the objective function $L+\mu\Omega$.
Since both $L$ and $\mu\Omega$ are convex functions, it means that they are both linear when restricted to $[w,\tilde{w}]$.

Now, $\mu\Omega$ is only linear on segments of the form $[v,tv]$ with $v\in\R{p}$ and $t>0$. So
we necessarily have $\tilde{w}=t w$ for some positive $t$. We now show that $L$ is strictly convex on
$[w,t w]$, which will contradict that it is linear on $[w,\tilde{w}]$.
Let $E=\text{Span}(x_1,\dots,x_n)$ and $E^{\perp}$ be the orthogonal of $E$ in $\R{p}$.
The vector $w$ can be decomposed in $w=w'+w''$ with $w'\in E$ and $w''\in E^{\perp}$.
Note that we have $w'\neq 0$ (since if it was equal to $0$, $w''$ would be the minimizer of
$\mu\Omega$, which would imply $w''=0$ and contradict $w\neq 0$).
We thus have $(w^{\top}x_1,\dots, w^{\top}x_n)=({w'}^{\top}x_1,\dots, {w'}^{\top}x_n)\neq 0.$

This implies that the function $s\mapsto \ell(y_i,s w ^\top x_i)$ is a polynomial of degree $2$. So it is not linear.
This contradicts the existence of two different solutions, and concludes the proof of uniqueness.

\begin{remark}
Still by using that a sum of convex functions is constant on a segment if and only if the functions are linear on this segment,
the proof can be extended in order to replace the alternative assumption ``$\{1,\dots,p\}$ belongs to $\G$'' by the weaker but more involved assumption:
for any $(j,k)\in\{1,\dots,p\}^2$, there exists a group $G\in\G$ which contains both $j$ and $k$.
\end{remark}

\section{Proof of Theorem \ref{thm:stability} }\label{app:stability_theorem}

For $w \in \R{p}$, we denote by $Z(w)$ its zero pattern (i.e., the indices of zero-components of~$w$).
To prove the result, it suffices to prove that for any set $I \subset \{1,\dots,p\}$ with
$I^c\notin \ZPattern$ and $|I|\le k-1$, the probability of
$$
 \mathcal{E}_I=\big\{Y\in\R{n}\text{:
   there exists}\ w\ \text{solution of the problem in \myeq{minF} with}\ Z(w)=I^c \big\}
$$
is equal to $0$.
We will prove this by contradiction: assume that there exists a set $I \subset \{1,\dots,p\}$ with
$I^c\notin \ZPattern$, $|I|\le k-1$ and $\P(\mathcal{E}_I)>0$.
Since $I^c\notin \ZPattern$, there exists $\alpha\in\mathrm{Hull}(I)\setminus I$.
Let $J=I\cup\{\alpha\}$ and $\G_I=\{ G \in \G: G \cap I \neq \emptyset \}$ be the set of active groups.
Define $\R{J}=\{w\in\R{p}:w_{J^c}=0\}$.
The restrictions $L_J:\R{J}\rightarrow \R{}$ and $\Omega_J:\R{J}\rightarrow \R{}$ of $L$ and $\Omega$
are continuously differentiable functions on $\big\{w\in\R{J}:w_{I}\neq 0\big\}$ with respective gradients
 $$
 \nabla L_J(w) = \bigg(\frac{\partial L_J}{\partial w_j}(w)\bigg)^{\top}_{j\in J}
  \quad
  \text{and}
  \quad
 \nabla \Omega_J(w) = \Bigg(
    w_j \bigg( \sum_{  \substack{G \in \G_I,\\ G \ni j} } (\dG_j)^2 \NormDeux{\dG \circ w}^{-1}  \bigg)
   \Bigg)_{j\in J}^{\top}.
 $$
Let $f(w,Y)=\nabla L_J(w) + \mu \nabla \Omega_J(w),$ where the dependence in $Y$ of $f(w,Y)$ is hidden in
$\nabla L_J(w)=\frac1n\sum_{i=1}^n (x_i)_J\frac{\partial \ell}{\partial y'}(y_i,w ^\top x_i)$.

For $Y\in\mathcal{E}_I$, there exists $w\in \R{J}$ with $Z(w)=I^c$,
which minimizes the convex function $L_J+\mu\Omega_J$. The vector $w$
satisfies $f(w,Y)=0$. So we have proved
 $\mathcal{E}_I \subset \mathcal{E}_I',$
where
 $$\mathcal{E}_I'=\big\{Y\in \R{n}: \text{ there exists $w \in \R{J}$ with 
 $Z(w)=I^c$ and $f(w,Y)=0$}\big\}.$$

Let $\ty\in \mathcal{E}_I$. Consider the equation $f(w,\ty)=0$ on $\big\{w\in\R{J}:w_{j}\neq 0 \text{ for any }j\in I\big\}$.
By construction, we have $|J|\le k$, and thus, by assumption,
the matrix $X^J =\big((x_1)_J,...,(x_n)_J\big)^\top \in \RR{n}{|J|}$
has rank $|J|.$ As in the proof of Proposition~\ref{thm:uniqueness}, this implies that the function $L_J$ is strictly convex,
and then, the uniqueness of the minimizer of $L_J+\mu\Omega,$
and also the uniqueness of the point at which the gradient of this function vanishes.
So the equation $f(w,\ty)=0$ on $\big\{w\in\R{J}:w_{j}\neq 0 \text{ for any }j\in I\big\}$ has a unique solution,
which we will write $w^\ty$.

On a small enough ball around $(w^{\ty}_J,\ty)$, $f$ is continuously differentiable since none of the norms vanishes at $w^{\ty}_J$. Let $(f_j)_{j\in J}$ be the components of $f$ and
$H_{JJ}=\big(\frac{\partial f_j}{\partial w_k}\big)_{j\in J,k\in J }$. The matrix $H_{JJ}$ is actually the sum of:
\begin{enumerate}
 \item[a)] the Hessian of $L_J$, which is positive definite (still from the same argument as in the proof of Theorem \ref{thm:uniqueness}),
 \item[b)] the Hessian of the norm $\Omega_J$ around $(w^{\ty}_J,\ty)$ that is positive semidefinite on this small ball according to the Hessian characterization of convexity \citep[Theorem 3.1.11]{borwein2006caa}.
\end{enumerate}
Consequently, $H_{JJ}$ is invertible. We can now apply the implicit function theorem to obtain that for $Y$ in a neighborhood of $\ty$,
$$
w^Y = \psi(Y),
$$
with $\psi=(\psi_j)_{j\in J}$ a continuously differentiable function satisfying the matricial relation
$$
(\dots,\nabla \psi_j,\dots ) H_{JJ} + (\dots,\nabla_y f_j,\dots ) =0.
$$
Let $C_{\tj}$ denote the column vector of $H_{JJ}^{-1}$ corresponding to the index $\alpha$,
and let $D$ the diagonal matrix whose $(i,i)$-th element is
$\frac{\partial^2 \ell}{\partial y \partial {y'}}(y_i,w ^\top x_i)$.
Since $n (\dots,\nabla_y f_j,\dots ) = DX^J$, we have
$$
n \nabla \psi_{\tj} = - DX^J C_{\tj}.
$$
Now, since $X^J$ has full rank, $C_{\tj} \neq 0$ and none of the diagonal elements
of $D$ is null (by assumption on $\ell$), we have $\nabla \psi_{\tj}\neq 0$.
Without loss of generality, we may assume that $\partial \psi_{\tj}/\partial y_1\neq 0$ on a neighborhood of $\ty$.

We can apply again the implicit function theorem to show that on an open ball in $\R{n}$ centered at $\ty$,
say $\cB_{\ty}$, the solution to $\psi_{\tj}(Y)=0$ can be written
$y_1=\varphi(y_2,\dots,y_n)$ with $\varphi$ a continuously differentiable function.

By Fubini's theorem and by using the fact that the Lebesgue measure of a singleton in $\R{n}$ equals zero,
we get that the set $A(\ty)=\big\{Y\in \cB_{\ty}: \psi_{\tj}(Y)=0\big\}$
has thus zero probability.
Let $\cS\subset\mathcal{E}_I$ be a compact set. We thus have $\cS\subset\mathcal{E}_I'$.

By compacity, the set $\cS$ can be covered
by a finite number of ball $\cB_{\ty}$. So there exist $\ty_1,\dots,\ty_m$ such that we have
 $\cS \subset A(\ty_1) \cup \cdots \cup A(\ty_m).$
Consequently, we have $\P(\cS)=0$.

Since this holds for any compact set in $\mathcal{E}_I$ and since the Lebesgue measure is regular, we have
$\P(\mathcal{E}_I)=0$, which contradicts the definition of $I$, and concludes the proof.

\section{Proof of the minimality of the Backward procedure (see Algorithm~\ref{alg:backward})}\label{app:backward}

There are essentially two points to show:
 \begin{itemize}
  \item $\G$ spans $\ZPattern$.
  \item $\G$ is minimal.
 \end{itemize}
The first point can be shown by a proof by recurrence on the depth of the DAG. At step $t$, the base $\G^{(t)}$ verifies $\{\bigcup_{G\in\G'}G,\ \forall \G'\subseteq\G^{(t)}\}=\{G\in\ZPattern, |G| \leq t\}$ because an element $G\in\ZPattern$ is either the union of itself or the union of elements strictly smaller. The initialization $t=\min_{G\in\ZPattern}|G|$ is easily verified, the leafs of the DAG being necessarily in $\G$.

As for the second point, we proceed by contradiction. If there exists another base $\G^*$ that spans $\ZPattern$ such that $\G^* \subset \G$, then \[\exists\ e \in \G,\ e \notin \G^*.\]By definition of the set $\ZPattern$, there exists in turn $\G' \subseteq \G^*,\ \G'\neq\{e\}$ (otherwise, $e$ would belong to $\G^*$), verifying $e=\bigcup_{G\in\G'}G$, which is impossible by construction of $\G$ whose members cannot be the union of elements of $\ZPattern$.

\section{Proof of Proposition \ref{prop:dual_problem}}\label{app:dual_problem}

The proposition comes from a classic result of Fenchel Duality \citep[Theorem 3.3.5 and Exercise 3.3.9]{borwein2006caa} when we consider the convex function 
$$
h_J:w_J \mapsto \frac{\lambda}{2} \left[ \Omega_J(w_J) \right]^2,
$$
whose Fenchel conjugate $h^*_J$ is given by 
$\kappa_J \mapsto \frac{1}{2\lambda} \left[ \Omega_{J}^*(\kappa_{J}) \right]^2$ \citep[example 3.27]{boyd}. 
Since the set 
$$
\{w_J \in \R{|J|};\ h_J(w_J) < \infty \} \cap \{w_J \in \R{|J|};\ L_J(w_J) < \infty  
\text{ and } 
L_J \text{ is continuous at } w_J \}
$$
is not empty, we get the first part of the proposition. Moreover, the primal-dual variables $\{w_J,\kappa_J\}$ is optimal if and only if
$$
\begin{cases}
 -\kappa_J  & \in \quad \partial L_J(w_J), \\
 \kappa_J  & \in \quad \partial [\frac{\lambda}{2} \left[ \Omega_J(w_J) \right]^2] 
= \lambda \Omega_J(w_J) \partial \Omega_J(w_J),
\end{cases}
$$
where $\partial \Omega_J(w_J)$ denotes the subdifferential of $\Omega_J$ at $w_J$.
The differentiability of $L_J$ at $w_J$ then gives $\partial L_J(w_J)=\{ \nabla L_J(w_J) \}$.
It now remains to show that 
\begin{equation}\label{eq:CNS_duality_gap}
 \kappa_J  \in \lambda \Omega_J(w_J) \partial \Omega_J(w_J)
\end{equation}
is equivalent to
\begin{equation}\label{eq:CNS_duality_gap_bis}
w_J ^\top \kappa_J = \frac{1}{\lambda} \left[ \Omega_J^*(\kappa_J) \right]^2 = \lambda \left[ \Omega_J(w_J) \right]^2.
\end{equation}
As a starting point, the Fenchel-Young inequality \citep[Proposition 3.3.4]{borwein2006caa} gives the equivalence between \myeq{CNS_duality_gap} and 
\begin{equation}\label{eq:zero_duality_gap}
 \frac{\lambda}{2} \left[ \Omega_J(w_J) \right]^2  +\frac{1}{2\lambda} \left[ \Omega_{J}^*(\kappa_{J}) \right]^2 = w_J ^\top \kappa_J. 
\end{equation}
In addition, we have \citep{rockafellar1970ca}
\begin{equation}\label{eq:def_subdifferential}
\partial \Omega_J(w_J) = \{ u_J \in \R{|J|}; u_J^\top w_J = \Omega_J(w_J) \ \mbox{ and }\ \Omega_J^*(u_J) \leq 1 \}.
\end{equation}
Thus, if 
$
 \kappa_J  \in \lambda \Omega_J(w_J) \partial \Omega_J(w_J)
$
then 
$
w_J ^\top \kappa_J=\lambda \left[ \Omega_J(w_J) \right]^2.
$
Combined with \myeq{zero_duality_gap}, we obtain 
$
w_J ^\top \kappa_J=\frac{1}{\lambda} \left[ \Omega_{J}^*(\kappa_{J}) \right]^2.
$

Reciprocally, starting from \myeq{CNS_duality_gap_bis}, we notably have
$$
w_J ^\top \kappa_J = \lambda \left[ \Omega_J(w_J) \right]^2.
$$
In light of \myeq{def_subdifferential}, it suffices to check that 
$
\Omega_{J}^*( \kappa_{J} ) \leq \lambda\Omega_J(w_J)
$ 
in order to have \myeq{CNS_duality_gap}.
Combining \myeq{CNS_duality_gap_bis} with the definition of the dual norm, it comes
$$
\frac{1}{\lambda} \left[ \Omega_{J}^*(\kappa_{J}) \right]^2 = w_J ^\top \kappa_J \leq \Omega_{J}^*(\kappa_{J}) \Omega_J(w_J), 
$$
which concludes the proof of the equivalence between \myeq{CNS_duality_gap} and \myeq{CNS_duality_gap_bis}.

\section{Proofs of Propositions~\ref{prop:CN} and \ref{prop:CS}}\label{app:CN_CS}

In order to check that the reduced solution $w_J$ is optimal for the full problem in \myeq{minFc}, we complete with zeros on ${J^c}$ to define $w$, compute $\kappa = - \nabla L(w)$, which is such that $\kappa_J = - \nabla L_J(w_J)$, and get a duality gap for the full problem equal to
$$
\frac{1}{2\lambda} \left( \left[ \Omega^*(\kappa) \right]^2 - \lambda w_J^\top\kappa_J \right).
$$
By designing upper and lower bounds for $\Omega^*(\kappa)$, we get sufficient and necessary conditions.

\subsection{Proof of Proposition~\ref{prop:CN}}

Let us suppose that $w^* = \tbinom{w^*_J}{0_{J^c}}$ is optimal for the full problem in \myeq{minFc}.
Following the same derivation as in \mylemma{optimalitycond} (up to the squaring of the regularization $\Omega$), we have that $w^*$ is a solution of \myeq{minFc} if and only if for all $u \in \R{p}$,
\begin{equation*}
 u^\top \nabla L(w^*) + \lambda \Omega(w^*)( u_J^{\top} r_J +  (\Omega_J^c)[u_{J^c}] ) \geq 0,
\end{equation*}
with 
$$
 r= \sum_{  G \in \G_J } \frac{\dG \circ \dG \circ w^*}{\NormDeux{\dG \circ w^*}}.
$$
We project the optimality condition onto the variables that can possibly enter the active set, i.e., the variables in $\Pi_\Pattern(J)$.
Thus, for each $K \in \Pi_\Pattern(J)$, we have for all $u_{K \backslash J} \in \R{|K \backslash J|}$,
\begin{equation*}
 u_{K \backslash J}^\top \nabla L(w^*)_{K \backslash J} + \lambda \Omega(w^*) \!\!\!
 \sum_{G \in \G_{K \backslash J} \cap (\G_J)^c } \NormDeux{ \dG_{K \backslash J} \circ u_{G \cap K \backslash J} } \geq 0.
\end{equation*}
By combining \mylemma{GKminusGJ} and the fact that $\G_{K \backslash J} \cap (\G_J)^c = \G_K \backslash \G_J$,
we have for all 
$G \in \G_K \backslash \G_J$, 
$K \backslash J \subseteq G$
and therefore 
$u_{G \cap K \backslash J}=u_{K \backslash J}$.
Since we cannot compute the dual norm of 
$u_{K \backslash J} \mapsto \| \dG_{K \backslash J} \circ u_{K \backslash J} \|_2$ in closed-form, 
we instead use the following upperbound
$$
\NormDeux{ \dG_{K \backslash J} \circ u_{K \backslash J} } \leq \|\dG_{K \backslash J}\|_{\infty} \NormDeux{u_{K \backslash J}},
$$
so that we get for all $u_{K \backslash J} \in \R{|K \backslash J|}$,
\begin{equation*}
 u_{K \backslash J}^\top \nabla L(w^*)_{K \backslash J} + \lambda \Omega(w^*)
 \sum_{G \in \G_K \backslash \G_J }\!\! \|\dG_{K \backslash J}\|_{\infty} \NormDeux{ u_{K \backslash J} }
 \geq 0.
\end{equation*}
Finally, Proposition~\ref{prop:dual_problem} gives
$\lambda \Omega(w^*)=\big\{ \!\! -\lambda {w^*} ^\top \nabla L(w^*) \big\}^{\frac{1}{2}}$, which leads to the desired result.


\subsection{Proof of Proposition~\ref{prop:CS}}

The goal of the proof is to upper bound the dual norm $\Omega^*(\kappa)$ by taking advantage of the structure of $\G$; 
we first show how we can upper bound $\Omega^*(\kappa)$ by $(\Omega_J^c)^\ast[\kappa_{J^c}]$.
We indeed have:
\begin{eqnarray*}
\Omega^*(\kappa) & = & \max_{   \sum_{G \in \G_J} \NormDeux{\dG \circ v} 
                                 + \sum_{G \in (\G_J)^c} \NormDeux{\dG \circ v} \leq 1
                             }  v^\top \kappa \\
& \leq & \max_{  \sum_{G \in \G_J} \NormDeux{\dG_J \circ v_J} 
                    + \sum_{G \in (\G_J)^c} \NormDeux{\dG \circ v} \leq 1 
                    } v^\top \kappa \\ 
& = & \max_{ \Omega_J(v_{J}) + (\Omega_J^c)(v_{J^c}) \leq 1} v^\top \kappa \\
& = & \max \left\{ \Omega_J^\ast(\kappa_{J}) ,  (\Omega_J^c)^\ast[\kappa_{J^c}] \right\},
\end{eqnarray*} where in the last line, we use \mylemma{disjointnorms}.
Thus the duality gap is less than 
$$
\frac{1}{2\lambda} \left( \left[ \Omega^*(\kappa) \right]^2 - \left[ \Omega_{J}^*(\kappa_{J}) \right]^2 \right)
\leq 
\frac{1}{2\lambda} \max \{ 0 , \left[ (\Omega_J^c)^*[\kappa_{J^c}] \right]^2  - \left[ \Omega_{J}^*(\kappa_{J}) \right]^2 \},
$$
and a sufficient condition for the duality gap to be smaller than $\varepsilon$ is
$$
 (\Omega_J^c)^*[\kappa_{J^c}] \leq ( 2\lambda\varepsilon + \left[ \Omega_{J}^*(\kappa_{J}) \right]^2 )^{\frac{1}{2}}.
$$
Using Proposition~\ref{prop:dual_problem}, we have 
$ -\lambda w ^\top \nabla L(w) = \left[ \Omega_{J}^*(\kappa_{J}) \right]^2$
and we get the right-hand side of Proposition~\ref{prop:CS}.
It now remains to upper bound $(\Omega_J^c)^\ast[\kappa_{J^c}]$. 
To this end, we call upon \mylemma{dual_norm_lowerbound} to obtain:
$$
(\Omega_J^c)^\ast[\kappa_{J^c}]
\leq
\max_{G \in (\G_J)^c}
  \left\{
  \sum_{j\in G} \left\{ \dfrac{\kappa_j}{\sum_{H \in j, H \in (\G_J)^c} \! \dH_j} \right\}^2
\right\}^{\frac{1}{2}}.
$$
Among all groups  $G \in (\G_J)^c $, the ones with the maximum values are the largest ones, i.e., those in the fringe groups 
$\mathcal{F}_J = \{G\in (\G_J)^c\ ;\ \nexists G' \in (\G_J)^c, G \subseteq G' \}$.
This argument leads to the result of Proposition~\ref{prop:CS}.


\section{Proof of Theorem \ref{thm:lowdim_patternconsistency}}\label{app:lowdim_patternconsistency}

\textit{Necessary condition:} We mostly follow the proof of ~\citet{grouplasso, zou2006ala}.
Let $\hat{w} \in \R{p}$ be the unique solution of
\[
 \min_{w\in\R{p}}L(w)+\mu\, \Omega(w)=\min_{w\in\R{p}}F(w).
\]
The quantity $\hat{\Delta}=(\hat{w}-\wb)/\mu$ is the minimizer of $\tF$ defined as
\[
 \tF(\Delta)=\frac{1}{2}\Delta^{\top}Q\Delta-\mu^{-1}q^{\top}\Delta+\mu^{-1}\left[ \Omega(\wb+\mu \Delta) - \Omega(\wb) \right],
\]
where $q= \frac{1}{n} \sum_{i=1}^n \varepsilon_i x_i$.
The random variable $\mu^{-1} q^{\top}\Delta$ is a centered Gaussian with variance
$\sqrt{\Delta^\top Q \Delta} /(n\mu^2)$.
Since $Q \rightarrow \Qb$, we obtain that for all $\Delta \in \R{p}$,
$$
\mu^{-1} q^{\top}\Delta=o_p(1).
$$
Since $\mu \rightarrow 0$, we also have by taking the directional derivative of $\Omega$ at $\wb$ in the direction of $\Delta$
\[
\mu^{-1}\left[ \Omega(\wb+\mu \Delta) - \Omega(\wb) \right] = \mathbf{r}_{\Jb}^{\top}\Delta_{\Jb} + \Omega_\Jb^c (\Delta_{\Jb^c}) + o(1),
\]
so that for all $\Delta \in \R{p}$
\[
\tF(\Delta) = \Delta^\top \Qb \Delta + \mathbf{r}_{\Jb}^\top \Delta_\Jb + \Omega_\Jb^c (\Delta_{\Jb^c}) + o_p(1)
                  = \tFlim(\Delta) + o_p(1).
\]
The limiting function $\tFlim$ being stricly convex (because $\Qb \succ 0$) and $\tF$ being convex, we have that the minimizer $\hat{\Delta}$ of $\tF$ tends in probability to the unique minimizer of $\tFlim$ \citep{fu2000alt} referred to as $\Delta^\ast$.

By assumption, with probability tending to one, we have $\Jb=\{j\in\IntegerSet{p}, \hat{w}_j \neq 0 \}$, hence for any $j\in\Jb^c$
$\mu\hat{\Delta}_j=(\wh-\wb)_j=0$. This implies that the nonrandom vector $\Delta^\ast$ verifies $\Delta_{\Jb^c}^\ast=0$. 

As a consequence, $\Delta_{\Jb}^\ast$ minimizes $\Delta_{\Jb}^\top \Qb_{\Jb\Jb} \Delta_{\Jb} + \mathbf{r}_{\Jb}^\top \Delta_\Jb$, hence $\mathbf{r}_{\Jb} = -\Qb_{\Jb\Jb}\Delta_{\Jb}^\ast$. Besides, since $\Delta^{\ast}$ is the minimizer of $\tFlim$, by taking the directional derivatives as in the proof of
Lemma \ref{lem:optimalitycond}, we have 
$$
(\Omega_\Jb^c)^\ast [\Qb_{\JCb\Jb}\Delta_{\Jb}^\ast] \leq 1.
$$ 
This gives the necessary condition.
\vspace*{1cm}

\textit{Sufficient condition:} We turn to the sufficient condition. We first consider the problem reduced to the hull $\Jb$,
\[
 \min_{w \in \R{|\Jb|}}  L_\J(w_\J) + \mu\Omega_{\J}(w_\J).
\]
that is strongly convex since $Q_{\Jb\Jb}$ is positive definite and thus admits a unique solution $\hat{w}_{\Jb}$. With similar arguments as the ones used in the necessary condition, we can show that $\hat{w}_{\Jb}$ tends in probability to the true vector $\wb_{\Jb}$. We now consider the vector $\hat{w} \in \R{p}$ which is the vector $\hat{w}_{\Jb}$ padded with zeros on $\JCb$.
Since, from Theorem \ref{thm:stability}, we almost surely have
 $\mathrm{Hull}(\{j\in\IntegerSet{p}, \hat{w}_j \neq 0 \})=\{j\in\IntegerSet{p}, \hat{w}_j \neq 0 \}$, we have already that the vector $\hat{w}$ consistently estimates the hull of $\wb$ and we have that $\hat{w}$ tends in probability to $\wb$. From now on, we consider that $\hat{w}$ is sufficiently close to $\wb$, so that for any $G\in\G_{\Jb}$, $\NormDeux{\dG \circ \wh} \neq 0$. We may thus introduce
 $$
 \hat{r}= \sum_{  G \in \G_{\Jb} } \frac{\dG \circ \dG \circ \hat{w}}{\NormDeux{\dG \circ \hat{w}}}.
 $$
It remains to show that $\hat{w}$ is indeed optimal for the full problem (that admits a unique solution due to the positiveness of $Q$).
By construction, the optimality condition (see Lemma \ref{lem:optimalitycond}) relative to the active variables $\Jb$ is already verified. More precisely, we have
\[
       \nabla L(\hat{w})_{\Jb} + \mu\, \hat{r}_{\Jb} = Q_{\Jb\Jb}(\hat{w}_{\Jb}-\wb_{\Jb})-q_{\Jb} + \mu\, \hat{r}_{\Jb}  =  0.
\]
Moreover, for all $u_{\JCb}\in\R{|\JCb|}$, by using the previous expression and the invertibily of $Q$, we have
\[
 u_{\JCb}^{\top} \nabla L(\hat{w})_{\JCb} = u_{\JCb}^{\top}
       \left\lbrace
               -\mu\, Q_{\JCb\Jb}Q_{\Jb\Jb}^{-1}\hat{r}_{\Jb} + Q_{\JCb\Jb}Q_{\Jb\Jb}^{-1}q_{\Jb} - q_{\JCb}
       \right\rbrace.
\]
The terms related to the noise vanish, having actually $q = o_p(1)$. Since $Q \rightarrow \Qb$ and $\hat{r}_{\Jb}\rightarrow \mathbf{r}_{\Jb}$, we get for all $u_{\JCb}\in\R{|\JCb|}$
\[
 u_{\JCb}^{\top} \nabla L(\hat{w})_{\JCb} = -\mu\, u_{\JCb}^{\top} \left\lbrace \Qb_{\JCb\Jb}\Qb_{\Jb\Jb}^{-1}\mathbf{r}_{\Jb} \right\rbrace + o_p(\mu).
\]
Since we assume $(\Omega_\Jb^c)^\ast[ \Qb_{\JCb\Jb} \Qb_{\Jb\Jb}^{-1} \mathbf{r}_\Jb] < 1$, we obtain
\[
 - u_{\JCb}^{\top} \nabla L(\hat{w})_{\JCb} < \mu (\Omega_{\Jb}^c)[u_{\JCb}] + o_p(\mu),
\]
which proves the optimality condition of Lemma \ref{lem:optimalitycond} relative to the inactive variables: $\hat{w}$ is therefore optimal for the full problem.

\section{Proof of Theorem \ref{thm:highdim_patternconsistency}}
\label{app:highdim}
Since our analysis takes place in a finite-dimensional space, all the norms defined on this space are equivalent. Therefore, we introduce the equivalence parameters $a(\Jb),A(\Jb) >0$ such that
$$
 \forall u \in \R{|\Jb|},\  a(\Jb) \NormUn{u} \leq  \Omega_{\Jb}[u] \leq A(\Jb) \NormUn{u}.
$$
We similarly define $a(\JCb),A(\JCb) >0$ for the norm $(\Omega_{\Jb}^c)$ on $\R{|\JCb|}$. In addition, we immediatly get by order-reversing:
$$
 \forall u \in \R{|\Jb|},\ A(\Jb)^{-1} \NormInf{u} \leq (\Omega_{\Jb})^\ast[u] \leq a(\Jb)^{-1} \NormInf{u}.
$$
For any matrix $\Gamma$, we also introduce the operator norm $\| \Gamma \|_{m,s}$ defined  as
$$
	\| \Gamma \|_{m,s} = \sup_{ \|u\|_s \leq 1} \| \Gamma u \|_m .
$$
Moreover, our proof will rely on the control of the \emph{expected dual norm for isonormal vectors}: $\E \left[ (\Omega_\Jb^c)^\ast(W) \right]$ with $W$ a centered Gaussian random variable with unit covariance matrix. In the case of the Lasso, it is of order $(\log p)^{1/2}$.

\renewcommand{\le}{\leq}
\renewcommand{\ge}{\geq}

Following~\citet{grouplasso} and~\citet{nardy}, we consider the reduced problem on~$\Jb$,
$$
\min_{w \in \R{p}}  L_\J(w_\J) + \mu\Omega_{\J}(w_\J)
$$
with solution $\hat{w}_\J$, which can be extended to $\J^c$ with zeros.
From optimality conditions (see Lemma \ref{lem:optimalitycond}), we know that
\begin{equation} \label{eq:521}
\Omega_\J^\ast[ Q_{\J\J}(\wh_\J - \w_\J) - q_\J ] \leq \mu,
\end{equation}
where the vector $q \in \R{p}$ is defined as $q= \frac{1}{n} \sum_{i=1}^n \varepsilon_i x_i$.
We denote by $\nu = \min \{ |\wb_j|;\ \wb_j\neq 0 \}$ the smallest nonzero components of $\wb$.
We first prove that we must have with high probability $\NormInf{\hat{w}_G} > 0$ for all $G \in \G_\Jb$, proving that the hull of the active set of $\hat{w}_\J$ is exactly $\J$ 
(i.e., no active group is missing).

We have
\begin{align*}
 \NormInf{\wh_\J - \w_\J } & \leq \|Q_{\J \J}^{-1}\|_{\infty,\infty} \NormInf{Q_{\J \J}(\wh_\J - \w_\J)} \\
& \le |\Jb|^{1/2} \kappa^{-1}  \left( \NormInf{Q_{\J \J} ( \wh_\J - \w_\J )  - q_\Jb }
+ \NormInf{q_\Jb}
\right),
\end{align*}
hence from \eqref{eq:521} and the definition of $A(\Jb)$,
 \begin{equation}\label{eq:delta_w}
 \NormInf{\wh_\J - \w_\J } \leq |\Jb|^{1/2} \kappa^{-1}
   \left( \mu A(\Jb) + \NormInf{q_\Jb} \right).
 \end{equation}
Thus, if we assume
 $\mu \leq \frac{ \kappa \nu}{3 |\Jb|^{1/2} A(\Jb) }$
and
 \begin{equation}\label{eq:qj}
 \|q_\J\|_\infty \leq \frac{  \kappa \nu}{3 |\Jb|^{1/2}},
 \end{equation}
we get
 \begin{equation}\label{eq:diffw}
 \NormInf{\wh_\J - \w_\J } \le 2 \nu/3,
 \end{equation}
so that for all $G \in \G_\J$, $\NormInf{\wh_G} \ge \frac{\nu}{3}$, hence the hull is indeed selected.

This also ensures that $\hat{w}_\J$ satisfies the equation (see Lemma \ref{lem:optimalitycond})
\begin{equation}\label{eq:optimilaty_cond_on_J}
Q_{\J\J} \left( \hat{w}_\J - \wb_\Jb \right) - q_\J + \mu \hat{r}_\J = 0,
\end{equation}
where
 $$
 \hat{r}= \sum_{  G \in \G_{\Jb} } \frac{\dG \circ \dG \circ \hat{w}}{\NormDeux{\dG \circ \hat{w}}}.
 $$

We now prove that the $\hat{w}$ padded with zeros on $\JCb$ is indeed optimal for the full problem with high probability. According to Lemma \ref{lem:optimalitycond}, since we have already proved \eqref{eq:optimilaty_cond_on_J},
it suffices to show that
\[
(\Omega_\J^c)^\ast[ \nabla L(\hat{w})_{\J^c} ] \leq \mu.
\]
Defining $q_{\J^c|\J} = q_{\J^c} - Q_{\J^c \J} Q_{\J \J}^{-1} q_\J$, we can write the gradient of $L$ on $\JCb$ as
\[
 \nabla L(\hat{w})_{\J^c} = -q_{\J^c|\J} - \mu Q_{\JCb \Jb} Q_{\Jb \Jb}^{-1} \hat{r}_\Jb
 =
 -q_{\J^c|\J} - \mu Q_{\JCb \Jb} Q_{\Jb \Jb}^{-1} (\hat{r}_\Jb -  \mathbf{r}_\Jb) - \mu Q_{\JCb \Jb} Q_{\Jb \Jb}^{-1} \mathbf{r}_\Jb,
\]
which leads us to control the difference $\hat{r}_\Jb - \mathbf{r}_\Jb$.
Using \mylemma{mean_value_thm_on_r}, we get
\begin{align*}
\NormUn{\hat{r}_\Jb - \mathbf{r}_\Jb} & \le \NormInf{\wh_\Jb-\wb_\Jb}
 \left( \sum_{G \in \G_{\Jb}} \frac{\NormDeux{\dG_\Jb}^2}{\NormDeux{\dG \circ w}}
 + \sum_{G \in \G_{\Jb}} \frac{\NormUn{\dG\circ\dG\circ w}^2}{\NormDeux{\dG \circ w}^{3}} \right),
\end{align*}
where $w = t_0 \wh + (1-t_0) \wb\,$ for some $\,t_0 \in (0,1)$.

Let $\Jo = \{k\in\Jb: \wb_k\neq 0\}$ and let $\varphi$ be defined as
$$
 \varphi= \sup_{\substack{u\in\R{p}:\Jo\subset\{k\in\Jb: u_k\neq 0\}\subset\Jb\\G\in\G_\Jb}}
   \frac{ \NormUn{\dG\circ\dG\circ u} }{ \NormUnc{\dG_{\Jo}\circ\dG_{\Jo}\circ u_{\Jo}} } \ge 1.
$$
The term $\varphi$ basically measures how close $\Jb$ and $\Jo$  are, i.e., how relevant the prior encoded by $\G$ about the hull $\Jb$ is.
By using \eqref{eq:diffw}, we have
$$
 \NormDeux{\dG\circ w}^2 \ge \NormDeux{\dG_\Jo\circ w_\Jo}^2 \ge
   \NormUnc{\dG_\Jo\circ \dG_\Jo\circ w_\Jo} \frac{\nu}{3} \ge
   \NormUnc{\dG\circ \dG\circ w} \frac{\nu}{3\varphi},
$$
$$
 \NormDeux{\dG\circ w} \ge \NormDeuxc{\dG_\Jo\circ w_\Jo} \ge
   \NormDeuxc{\dG_\Jo} \frac{\nu}{3} \ge
   \NormDeux{\dG_\Jb} \frac{\nu}{3\sqrt{\varphi}}
$$
and
$$
 \NormInf{w}\le \frac53 \NormInf{\wb}.
$$
Therefore we have
 \begin{align*}
\NormUn{\hat{r}_\Jb - \mathbf{r}_\Jb} & \le \NormInf{\wh_\Jb-\wb_\Jb}
 \sum_{G \in \G_{\Jb}} \left( \frac{\NormDeux{\dG_\Jb}^2}{\NormDeux{\dG \circ w}}
 + \frac{5\varphi}{\nu}\frac{\|\wb\|_\infty \NormUn{\dG_\Jb\circ\dG_\Jb}}{\NormDeux{\dG \circ w}} \right)\\
 & \le \frac{3\sqrt{\varphi} \NormInf{\wh_\Jb-\wb_\Jb}}{\nu}
   \left( 1+\frac{5\varphi \|\wb\|_\infty}{\nu}\right) \sum_{G \in \G_{\Jb}} \NormDeux{\dG_\Jb}.
\end{align*}
Introducing
 $\constant=\frac{18\varphi^{3/2}\|\wb\|_\infty}{\nu^2}\sum_{G \in \G_{\Jb}} \NormDeux{\dG_\Jb},$
we thus have proved
\begin{equation}\label{eq:delta_r_delta_w}
 \NormUn{\hat{r}_\Jb - \mathbf{r}_\Jb} \leq \constant \NormInf{\wh_\Jb - \wb_\Jb}.
\end{equation}

By writing the Schur complement of $Q$ on the block matrices $Q_{\JCb \JCb} $ and $Q_{\J \J}$,
the positiveness of $Q$ implies that the diagonal terms ${\rm diag}(Q_{\J^c \J} Q_{\J \J}^{-1} Q_{\J \J^c})$ are less than one, which results in 
$\|Q_{\J^c \J} Q_{\J \J}^{-1/2}\|_{\infty,2} \le 1$. We then have
\begin{eqnarray}
\NormInf{Q_{\J^c \J} Q_{\J \J}^{-1} ( \hat{r}_\J - \mathbf{r}_\J  )}
& = & \NormInf{Q_{\J^c \J} Q_{\J \J}^{-1/2}   Q_{\J \J}^{-1/2}( \hat{r}_\J - \mathbf{r}_\J )}
\\
& \leq & \|Q_{\J^c \J} Q_{\J \J}^{-1/2}\|_{\infty,2} \|Q_{\J \J}^{-1/2}\|_2 
\NormDeux{\hat{r}_\J - \mathbf{r}_\J }
\\
& \leq &  \kappa^{-1/2} \NormUn{\hat{r}_\J - \mathbf{r}_\J } \label{eq:zzzz} \\
& \leq &  \kappa^{-3/2}  \constant  |\Jb|^{1/2}  \left( \mu A(\Jb) + \NormInf{q_\Jb} \right)
,
\end{eqnarray}
where the last line comes from Eq. (\ref{eq:delta_w}) and (\ref{eq:delta_r_delta_w}).
We get
$$ (\Omega_\J^c)^\ast[ Q_{\J^c \J} Q_{\J \J}^{-1} ( \hat{r}_\J - \mathbf{r}_\J  ) ]
\leq  
\frac{\constant |\Jb|^{1/2}}{ \kappa^{3/2}  a(\JCb) } \left( \mu A(\Jb) + \NormInf{q_\Jb} \right).
$$

Thus, if the following inequalities are verified
\begin{align}
 \frac{\constant |\Jb|^{1/2} A(\Jb)}{\kappa^{3/2} a(\JCb)}
   \mu 
 & \leq \frac{\tau}{4} , \label{eq:1}\\
 \frac{\constant |\Jb|^{1/2}}{\kappa^{3/2} a(\JCb) }       \NormInf{q_\Jb}
 & \leq \frac{\tau}{4} , \label{eq:2}\\
 (\Omega_\J^c)^\ast[q_{\J^c| \J} ] & \leq \frac{\mu\tau}{2}, \label{eq:3}
\end{align}
we obtain
\begin{eqnarray*}
(\Omega_\J^c)^\ast[ \nabla L(\hat{w})_{\J^c} ]
&\!\!\! \leq \!\!\!& (\Omega_\J^c)^\ast[ - q_{\J^c|\J} - \mu Q_{\J^c \J} Q_{\J \J}^{-1} \mathbf{r}_\J ] \\
&\!\!\! \leq\!\!\! &  (\Omega_\J^c)^\ast[ - q_{\J^c|\J} ] + \mu (1 - \tau) + \mu \tau/2 \leq \mu,
\end{eqnarray*}
i.e.,
 $\J$ is exactly selected.

 Combined with earlier constraints, this leads to the first part of the desired proposition.

We now need to make sure that the conditions \eqref{eq:qj}, \eqref{eq:2} and \eqref{eq:3} hold with high probability. To this end, we upperbound, using Gaussian concentration inequalities, two tail-probabilities.
First, $q_{\J^c| \J}$ is a centered Gaussian random vector with covariance matrix
 \begin{align*}
 \E \big[ q_{\J^c| \J} q_{\J^c| \J}^\top \big] & = \E \Big[ q_{\J^c}q_{\J^c}^\top - q_{\J^c} q_{\J}^\top Q_{\J \J}^{-1} Q_{\J \J^c}
   - Q_{\J^c \J} Q_{\J \J}^{-1} q_{\J} q_{\J^c}^\top
   + Q_{\J^c \J} Q_{\J \J}^{-1} q_{\J} q_{\J}^\top Q_{\J \J}^{-1} Q_{\J \J^c} \Big] \\
 & = \frac{\sigma^2}{n} Q_{\J^c \J^c |\J},
 \end{align*}
where $Q_{\J^c \J^c |\J}=Q_{\J^c \J^c} - Q_{\J^c \J} Q_{\J \J}^{-1} Q_{\J \J^c}$.
In particular, $(\Omega_\J^c)^\ast[q_{\J^c| \J} ]$ has the same distribution as
$\psi(W)$, with $\psi: u \mapsto (\Omega_\J^c)^\ast( \sigma n^{-1/2} Q_{\J^c \J^c| \J}^{1/2} u)$ and
$W$ a centered Gaussian random variable with unit covariance matrix.

Since for any $u$ we have $u^\top Q_{\J^c \J^c| \J} u \le u^\top Q_{\J^c \J^c} u \le \NormDeux{Q^{1/2}}^2\NormDeux{u}^2$,
by using Sudakov-Fernique inequality~\citep[Theorem 2.9]{Adler90}, we get:
\begin{align*}
\E [ (\Omega_\J^c)^\ast[q_{\J^c| \J} ] = \E \sup_{ \Omega_\J^c(u) \le 1 } u^\top q_{\J^c| \J} &
\leq \sigma n^{-1/2} \| Q \|_2^{1/2} \E \sup_{ \Omega_\J^c(u) \le 1 } u^\top W\\
& \leq \sigma n^{-1/2} \| Q \|_2^{1/2} \E [ (\Omega_\J^c)^\ast(W) ].
\end{align*}

In addition, we have
$$
|\psi(u)-\psi(v)| \leq \psi( u-v ) \leq \sigma n^{-1/2} a(\JCb)^{-1} \NormInf{Q_{\J^c \J^c| \J}^{1/2} (u-v)}.
$$
On the other hand, since $Q$ has unit diagonal and $Q_{\J^c \J} Q_{\J \J}^{-1} Q_{\J \J^c}$ has diagonal terms less than one, 
$Q_{\J^c \J^c |\J}$ also has diagonal terms less than one, which implies that $\|Q_{\J^c \J^c |\J}^{1/2}\|_{\infty,2} \leq 1$.
Hence $\psi$~is a Lipschitz function with Lipschitz constant upper bounded by
$\sigma n^{-1/2} a(\JCb)^{-1}$.
Thus by concentration of Lipschitz functions of multivariate standard random variables~\citep[Theorem 3.4]{massart-concentration}, we have for $t > 0$:
 $$
\P\! \left[ (\Omega_\J^c)^\ast[q_{\J^c| \J} ] \!\geq\!  t \! + \!
 \sigma n^{-1/2} \| Q \|_2^{1/2} \E \left[ (\Omega_\Jb^c)^\ast(W) \right] \right] \leq
\exp\left( \! - \frac{n t^2 a(\JCb)^2 }{2 \sigma^2 } \right).
$$
Applied for $t=\mu \tau/2 \ge 2\sigma n^{-1/2}\| Q \|_2^{1/2}\E \left[ (\Omega_\Jb^c)^\ast(W) \right] $, we get (because $(u-1)^2 \geq u^2 /4$ for $u\geq 2$):
 $$
\P\! \left[ (\Omega_\J^c)^\ast[q_{\J^c| \J} ] \!\geq\!  t  \right]  
\leq
\exp\left( \! - \frac{ n \mu^2 \tau^2 a(\JCb)^{2}}{32 \sigma^2} \right).
$$

It finally remains to control the term $\P(  \NormInf{ q_{\Jb} } \geq \xi )$, with
$$
   \xi = \frac{ \kappa \nu}{3 } \min\left\{
     1 , \frac{3 \tau  \kappa^{1/2} a(\JCb)}{ 4 \constant \nu}
   \right\}.
$$
We can apply classical inequalities for standard random variables~\citep[Theorem 3.4]{massart-concentration} that directly lead to
\[
\P(  \NormInf{ q_{\Jb} } \geq \xi )
       \leq 2 |\J| \exp\left( - \frac{n \xi^2 }{ 2\sigma^2 } \right).
\]
To conclude, Theorem \ref{thm:highdim_patternconsistency} holds with
\begin{eqnarray}
 C_1(\G,\Jb) &=& \frac{a(\JCb)^{2}}{16}, \\
 C_2(\G,\Jb) &=& \left( \frac{ \kappa \nu}{3} \min\left\{ 1 , \frac{\tau  \kappa^{1/2} a(\JCb) \nu}
     {24 \varphi^{3/2}\|\wb\|_\infty \sum_{G \in \G_{\Jb}} \NormDeux{\dG_\Jb} } \right\}
     \right)^2, \\
 C_3(\G,\Jb) &=& 4 \| Q \|_2^{1/2}\E \left[ (\Omega_\Jb^c)^\ast(W) \right],
\end{eqnarray}
and
 \[
 C_4(\G,\Jb) = \frac{\kappa \nu}{3 A(\Jb)}
   \min\left\{ 1 , \frac{\tau  \kappa^{1/2} a(\JCb) \nu }
    {24 \varphi^{3/2} \NormInf{\wb} \sum_{G \in \G_{\Jb}} \NormDeux{\dG_\Jb}}\right\},
 \]
where we recall the definitions: $W$ a centered Gaussian random variable with unit covariance matrix, $\Jo = \{j\in\Jb: \wb_j\neq 0\}$, $\nu = \min \{ |\wb_j|;\ j \in \Jo \}$,
 $$
 \varphi= \sup_{\substack{u\in\R{p}:\Jo\subset\{k\in\Jb: u_k\neq 0\}\subset\Jb\\G\in\G_\Jb}}
   \frac{ \NormUnc{\dG\circ\dG\circ u} }{ \NormUnc{\dG_{\Jo}\circ\dG_{\Jo}\circ u_{\Jo}} },
 $$
$\kappa = \lambda_{\min}(Q_{\Jb \Jb})>0$ and $\tau>0$ such that
$(\Omega_\Jb^c)^\ast[ Q_{\JCb\Jb} Q_{\Jb\Jb}^{-1} \mathbf{r}] < 1  - \tau $.

\section{A first order approach to solve \myeq{minF} and \myeq{minFc}}\label{app:eta_trick}

Both regularized minimization problems \myeq{minF} and \myeq{minFc} (that just differ in the squaring of $\Omega$) can be solved by using generic toolboxes for second-order cone programming (SOCP) \citep{boyd}.
We propose here a first order approach that takes up ideas from \cite{grouplasso, MicchelliP05} and that is based on the following variational equalities:
for $x \in \R{p}$, we have 
$$
 \NormUn{x}^2 = \min_{ \substack{ z \in \R{p}_{+}, \\ \sum_{j=1}^p \! z_j \leq 1 } } \sum_{j=1}^p \frac{x_j^2}{z_j},
$$
whose minimum is uniquely attained for $z_j = |x_j| / \NormUn{x}$.
Similarly, we have
$$
 2\NormUn{x} = \min_{  z \in \R{p}_{+} } \sum_{j=1}^p \frac{x_j^2}{z_j} + \NormUn{z},
$$
whose minimun is uniquely obtained for $z_j = |x_j|$.
Thus, we can equivalently rewrite \myeq{minF} as 
\begin{equation}\label{eq:eta_trick_one}
\min_{ \substack{ w \in \R{p}, \\ (\etaG)_{G\in\G} \in \R{|\G|}_+ } }  \ERisk{y_i}{w ^\top x_i} + 
\frac{\mu}{2} \sum_{j=1}^p w_j^2 \zeta_j^{-1} + \frac{\mu}{2}\NormUn{ (\etaG)_{G\in\G} }, 
\end{equation}
with $\zeta_j = ( \sum_{G \ni j} (\dG_j)^2 (\etaG)^{-1} )^{-1}$.
In the same vein, \myeq{minFc} is equivalent to 
\begin{equation}\label{eq:eta_trick_two}
\min_{ \substack{ w \in \R{p}, \\ (\etaG)_{G\in\G} \in \R{|\G|}_+, \\ \sum_{G\in\G} \etaG \leq 1 } }  \ERisk{y_i}{w ^\top x_i} + 
\frac{\lambda}{2} \sum_{j=1}^p w_j^2 \zeta_j^{-1}, 
\end{equation}
where $\zeta_j$ is defined as above.
The reformulations \myeq{eta_trick_one} and \myeq{eta_trick_two} lend themselves well to a simple alternating optimization scheme between 
$w$ (for instance, $w$ can be computed in closed-form when the square loss is used) 
and 
$(\etaG)_{G\in\G}$ (whose optimal value is always a closed-form solution).

This first order approach is computationally appealing since it allows \emph{warm-restart}, which can dramatically speed up the computation over regularization paths.

\section{Technical lemmas}
In this last section of the appendix, we give several technical lemmas.
We consider $I \subseteq \IntegerSet{p}$ and $\G_I=\{G\in\G;\ G \cap I \neq \varnothing \} \subseteq \G$, i.e., the set of active groups when the variables $I$ are selected.

We begin with a dual formulation of $\Omega^*$ obtained by conic duality \citep{boyd}:
\begin{lemma}\label{lem:dual_norm_dual_formulation} Let $u_I \in \R{|I|}$. We have
\begin{eqnarray*}
  (\Omega_I)^*[u_I] &=& \min_{ ( \xi_I^{\scriptscriptstyle G} )_{G\in\G_I} } \max_{ G\in\G_I } \NormDeux{\xi_I^{\scriptscriptstyle G}} \\
  &\mbox{ s.t. }& u_j + \!\! \sum_{ G \in \G_I, G \ni j } \dG_j \xi^{\scriptscriptstyle G}_j = 0  \, \mbox{ and } \, \xi^{\scriptscriptstyle G}_j = 0 \, \mbox{ if } \, j \notin G.
\end{eqnarray*}
\end{lemma}

\begin{proof} 
By definiton of $(\Omega_I)^*[u_I]$, we have 
$$
 (\Omega_I)^*[u_I] = \max_{\Omega_I(v_I) \leq 1} u_I^\top v_I.
$$
By introducing the primal variables $(\alpha_G)_{G\in\G_I} \in \R{|\G_I|}$, we can rewrite the previous maximization problem as
$$
 (\Omega_I)^*[u_I] = \max_{ \sum_{G\in\G_I} \!\! \alpha_G \leq 1 } u_I^\top v_I,\quad \mbox{ s.t. } \quad \forall\ G\in\G_I,\ \NormDeux{ \dG_I \circ u_{G\cap I} } \leq \alpha_G,
$$
which is a second-order cone program (SOCP) with $|\G_I|$ second-order cone constraints. 
This primal problem is convex and satisfies Slater's conditions for generalized conic inequalities, which implies that strong duality holds \citep{boyd}.
We now consider the Lagrangian $\mathcal{L}$ defined as
$$
\mathcal{L}(v_I,\alpha_G,\gamma, \tau_G,\xi_I^{\scriptscriptstyle G}) 
= 
u_I^\top v_I + \gamma (1 \! - \!\! \sum_{G\in\G_I} \!\! \alpha_G ) + \sum_{G\in\G_I} \binom{\alpha_G}{\dG_I \circ u_{G\cap I}}^\top \! \binom{\tau_G}{\xi_I^{\scriptscriptstyle G}}, 
$$ 
with the dual variables 
$\{\gamma, (\tau_G)_{G\in\G_I},(\xi_I^{\scriptscriptstyle G})_{G\in\G_I}\} \in \R{}_+ \! \times \! \R{|\G_I|} \! \times \!\RR{|I|}{|\G_I|}$
such that for all $G\in\G_I$, 
$\xi^{\scriptscriptstyle G}_j = 0 \, \mbox{ if } \, j \notin G$
and 
$ \NormDeux{\xi^{\scriptscriptstyle G}_I} \leq \tau_G $.
The dual function is obtained by taking the derivatives of $\mathcal{L}$ with respect to the primal variables $v_I$ and $(\alpha_G)_{G\in\G_I}$ and equating them to zero,
which leads to 
\begin{eqnarray*}
	\forall j\in I,&      u_j + \!\! \sum_{ G \in \G_I, G \ni j } \dG_j \xi^{\scriptscriptstyle G}_j & =  0 \\
	\forall G \in \G_I,& \gamma -  \tau_G & = 0.
\end{eqnarray*}
After simplifying the Lagrangian, the dual problem then reduces to
$$
	\min_{ \gamma, (\xi_I^{\scriptscriptstyle G})_{G\in\G_I} }  \gamma  \quad \mbox{ s.t. }
	\begin{cases}
	 \forall j \in I, u_j + \!\! \sum_{ G \in \G_I, G \ni j } \dG_j \xi^{\scriptscriptstyle G}_j = 0\,\mbox{ and }\, \xi^{\scriptscriptstyle G}_j = 0 \, \mbox{ if } \, j \notin G,\\
	 \forall G \in \G_I, \NormDeux{\xi^{\scriptscriptstyle G}_I} \leq \gamma,
	\end{cases}
$$
which is equivalent to the displayed result.
\end{proof}
Since we cannot compute in closed-form the solution of the previous optimization problem, 
we focus on a different \textit{but closely related} problem, i.e., when we replace the objective 
$\max_{ G\in\G_I } \NormDeux{\xi_I^{\scriptscriptstyle G}}$
by
$\max_{ G\in\G_I } \NormInf{\xi_I^{\scriptscriptstyle G}}$,
to obtain a \textit{meaningful} feasible point:

\begin{lemma}\label{lem:dual_norm_aux_dual_formulation} Let $u_I \in \R{|I|}$. The following problem
\begin{eqnarray*}
  &\min_{ ( \xi_I^{\scriptscriptstyle G} )_{G\in\G_I} } & \max_{ G\in\G_I } \NormInf{\xi_I^{\scriptscriptstyle G}} \\
  &\mbox{ s.t. }& u_j + \!\! \sum_{ G \in \G_I, G \ni j } \dG_j \xi^{\scriptscriptstyle G}_j = 0  \, \mbox{ and } \, \xi^{\scriptscriptstyle G}_j = 0 \, \mbox{ if } \, j \notin G,
\end{eqnarray*}
is minimized for 
$
 (\xi_j^{\scriptscriptstyle G})^* = -\dfrac{ u_j }{ \sum_{H \in j, H \in \G_I} \!\! \dH_j }.
$
\end{lemma}

\begin{proof} We proceed by contradiction. Let us assume there exists $( \xi_I^{\scriptscriptstyle G} )_{G\in\G_I}$ such that
\begin{eqnarray*}
 \max_{ G\in\G_I } \NormInf{\xi_I^{\scriptscriptstyle G}} &<& \max_{ G\in\G_I } \NormInf{ (\xi_I^{\scriptscriptstyle G})^*} \\
                                                          &=& \max_{ G\in\G_I } \max_{j \in G} \dfrac{ |u_j| }{ \sum_{H \in j, H \in \G_I} \!\! \dH_j }\\
                                                          &=& \dfrac{ |u_{j_0}| }{ \sum_{H \in j_0, H \in \G_I} \!\! \dH_{j_0} },
\end{eqnarray*}
where we denote by $j_0$ an argmax of the latter maximization. We notably have for all $G \ni j_0$:
$$
	|\xi_{j_0}^{\scriptscriptstyle G}| < \frac{ |u_{j_0}| }{ \sum_{H \in j_0, H \in \G_I} \!\! \dH_{j_0} }. 
$$
By multiplying both sides by $\dG_{j_0}$ and by summing over $G \ni j_0$, we get
$$	
	|u_{j_0}|=| \!\! \sum_{ G \in \G_I, G \ni {j_0} } \!\! \dG_{j_0} \xi^{\scriptscriptstyle G}_{j_0}| \leq \sum_{G \ni j_0} \dG_{j_0} |\xi_{j_0}^{\scriptscriptstyle G}| < |u_{j_0}|, 
$$
which leads to a contradiction.
\end{proof}

We now give an upperbound on $\Omega^*$ based on \mylemma{dual_norm_dual_formulation} and \mylemma{dual_norm_aux_dual_formulation}:

\begin{lemma}\label{lem:dual_norm_lowerbound} Let $u_I \in \R{|I|}$. We have
$$
  (\Omega_I)^*[u_I] \leq 
  \max_{G \in \G_I}
  \left\{
  \sum_{j\in G} \left\{ \dfrac{u_j}{\sum_{H \in j, H \in \G_I} \!\! \dH_j} \right\}^2
  \right\}^{\frac{1}{2}}.
$$
\end{lemma}

\begin{proof} 
We simply plug the minimizer obtained in \mylemma{dual_norm_aux_dual_formulation} into the problem of \mylemma{dual_norm_dual_formulation}.
\end{proof}

We now derive a lemma to control the difference of the gradient of $\Omega_J$ evaluated in two points:

\begin{lemma}\label{lem:mean_value_thm_on_r} Let $u_J,v_J$ be two nonzero vectors in $\R{|J|}$. Let us consider the mapping 
$w_J \mapsto r(w_J)=
\sum_{  G \in \G_J } \frac{\dG_J \circ \dG_J \circ w_J}{\NormDeux{\dG_J \circ w_J}} \in \R{|J|}.
$
There exists
$z_J = t_0 u_J + (1-t_0) v_J\,$ for some $\,t_0\in(0,1)$ 
such that 
\begin{align*}
\NormUn{r(u_J) - r(v_J)} & \le \NormInf{u_J-v_J}
 \left( \sum_{G \in \G_J} \frac{\NormDeux{\dG_J}^2}{\NormDeux{\dG_J \circ z_J}}
 + \sum_{G \in \G_J} \frac{\NormUn{\dG_J \circ \dG_J \circ z_J}^2}{\NormDeux{\dG_J \circ z_J}^{3}} \right).
\end{align*}
\end{lemma}
\begin{proof}For $j,k \in J$, we have 
$$
\frac{\partial r_j}{\partial w_k}(w_J)
 = \sum_{G \in \G_J} \frac{(\dG_j)^2}{\NormDeux{\dG_J \circ w_J}} \mathbb{I}_{j=k}
 - \sum_{G \in \G_J} \frac{(\dG_j)^2 w_j}{\NormDeux{\dG_J \circ w_J}^{3}} (\dG_k)^2 w_k,
$$
with $\mathbb{I}_{j=k}=1$ if $j=k$ and $0$ otherwise.
We then consider 
$t \in [0,1] \mapsto h_j(t) = r_j( t u_J + (1-t) v_J ).$
The mapping $h_j$ being continuously differentiable, we can apply the mean-value theorem: there exists $t_0 \in (0,1)$ such that
$$
h_j(1)-h_j(0) = \frac{\partial h_j(t)}{\partial t}(t_0).
$$
We then have
\begin{align*}
| r_j(u_J) - r_j(v_J) | & \le \sum_{k\in J} \bigg|\frac{\partial r_j}{\partial w_k}(z)\bigg| |u_k-v_k|\\
 & \le \NormInf{u_J-v_J} \left( \sum_{G \in \G_J} \frac{(\dG_j)^2}{\NormDeux{\dG_J \circ z_J}}
 + \sum_{k\in J} \sum_{G \in \G_J} \frac{(\dG_j)^2 |z_j|}{\NormDeux{\dG_J \circ z_J}^{3}} (\dG_k)^2 |z_k| \right),
\end{align*}
which leads to
\begin{align*}
\NormUn{ r(u_J) - r(v_J) } & \le \NormInf{u_J-v_J}
 \left( \sum_{G \in \G_J} \frac{\NormDeux{\dG_J}^2}{\NormDeux{\dG_J \circ z_J}}
 + \sum_{G \in \G_J} \frac{\NormUn{\dG_J\circ\dG_J\circ z_J}^2}{\NormDeux{\dG_J \circ z_J}^{3}} \right).
\end{align*}
\end{proof}


Given an active set $J \subseteq \IntegerSet{p}$ and a direct parent $K \in \Pi_\Pattern(J)$ of $J$ in the DAG of nonzero patterns, we have the following result:

\begin{lemma}\label{lem:GKminusGJ} For all $G \in \G_K \backslash \G_J$, we have
\[ K\backslash J \subseteq G \]
\end{lemma}

\begin{proof} We proceed by contradiction. We assume there exists $G_0 \in \G_K \backslash \G_J$ such that $K\backslash J \nsubseteq G_0$. Given that $K \in \Pattern$, there exists $\G' \subseteq \G$ verifying $K = \bigcap_{G \in \G'}G^c$. Note that $G_0 \notin \G'$ since by definition $G_0 \cap K \neq \varnothing$.

We can now build the pattern 
$\tilde{K}=\bigcap_{G \in \G'\cup\{G_0\}}G^c=K \cap G_0^c$
that belongs to $\Pattern$. 
Moreover, $\tilde{K} = K \cap G_0^c \subset K$ since we assumed $G_0^c \cap K \neq \varnothing$. 
In addition, we have that $J \subset K$  and $J \subset G_0^c$ because $K \in \Pi_\Pattern(J)$ and $G_0 \in \G_K \backslash \G_J$. 
This results in
$$
J \subset \tilde{K} \subset K,
$$
which is impossible by definition of $K$.
\end{proof}

We give below an important Lemma to characterize the solutions of (\ref{eq:minF}).

\begin{lemma}\label{lem:optimalitycond} The vector $\hat{w} \in \R{p}$ is a solution of 
\[
\min_{w\in\R{p}}L(w)+\mu\, \Omega(w)
\]
if and only if
\[
\begin{cases}
 \nabla L(\hat{w})_{\Jh} + \mu\, \hat{r}_{\Jh}   =  0 \\
 (\Omega_{\Jh}^c)^\ast [\nabla L(\hat{w})_{\JCh}] \leq  \mu,
\end{cases}
\]
with $\Jh$ the hull of $\{j\in\IntegerSet{p}, \hat{w}_j \neq 0 \}$ and the vector $\hat{r} \in \R{p}$ defined as 
$$
 \hat{r}= \sum_{  G \in \G_{\Jh} } \frac{\dG \circ \dG \circ \hat{w}}{\NormDeux{\dG \circ \hat{w}}}.
$$
In addition, the solution $\hat{w}$ satisfies
\[
\Omega^\ast[\nabla L(\hat{w})] \leq \mu.
\]
\end{lemma}
\begin{proof}
The problem
$$
\min_{w\in\R{p}}L(w)+\mu\, \Omega(w)=\min_{w\in\R{p}}F(w)
$$
being convex, the directional derivative optimality condition are necessary and sufficient \citep[Propositions 2.1.1-2.1.2]{borwein2006caa}.
Therefore, the vector $\hat{w}$ is a solution of the previous problem if and only if for all directions $u\in\R{p}$, we have
$$
\lim_{ \substack{ \varepsilon\to 0 \\ \varepsilon> 0} } 
	 \frac{F(\hat{w}+\varepsilon u)-F(\hat{w})}{\varepsilon}  \geq 0.
$$
Some algebra leads to the following equivalent formulation
\begin{equation}\label{eq:optimalitycond_eq}
\forall u\in\R{p},\ u^{\top} \nabla L(\hat{w}) + \mu\, u_{\Jh}^{\top} \hat{r}_{\Jh} + \mu\, (\Omega_{\Jh}^c)[u_{\JCh}] \geq 0.
\end{equation}
The first part of the lemma then comes from the projections on $\Jh$ and $\JCh$.

An application of the Cauchy-Schwartz inequality on $u_{\Jh}^{\top} \hat{r}_{\Jh}$ gives for all $u \in \R{p}$
$$
 u_{\Jh}^{\top} \hat{r}_{\Jh} \leq (\Omega_{\Jh})[u_{\Jh}].
$$
Combined with the equation (\ref{eq:optimalitycond_eq}), we get
$$
\forall u\in\R{p},\ u^{\top} \nabla L(\hat{w}) + \mu\, \Omega(u) \geq 0, 
$$
hence the second part of the lemma.
\end{proof}


We end up with a lemma regarding the dual norm of the sum of two \textit{disjoint} norms \citep[see][]{rockafellar1970ca}:

\begin{lemma}\label{lem:disjointnorms} Let $A$ and $B$ be a partition of $\IntegerSet{p}$, i.e.,  $A \cap B = \varnothing$ and $A \cup B = \IntegerSet{p}$. We consider two norms $u_{A} \in \R{|A|} \mapsto \|u_{A}\|_{A}$ and $u_{B} \in \R{|B|} \mapsto \|u_{B}\|_{B}$, with dual norms $\|v_{A}\|_{A}^*$ and $\|v_{B}\|_{B}^*$. We have
\[\max_{\|u_{A}\|_{A}+\|u_{B}\|_{B} \leq 1} u^{\top}v = \max \left\lbrace \|v_{A}\|_{A}^*,\ \|v_{B}\|_{B}^* \right\rbrace . \]
\end{lemma}



\bibliography{StructuredSparsityInducingNorms_JMLR}

\end{document}